Master's Thesis

Czech Technical University in Prague

F3

Faculty of Electrical Engineering
Department of Computer Science

# Estimating Bicycle Route Attractivity from Image Data

**Vít Růžička**
Master Programme: Open Informatics
Branch of Study: Computer Graphics and Interaction
ruzicvi3@fel.cvut.cz

Supervisor: Ing. Jan Drchal, Ph.D.
Department of Computer Science

# Acknowledgement

I would like to thank my supervisor Ing. Jan Drchal, Ph.D. for his help and leadership of this thesis and my alma mater Czech Technical University in Prague for giving me education. I would further like to thank the Indian Institute of Technology Madras भारतीय प्रौद्योगिकी संस्थान मद्रास and professor N. S. Narayanaswamy for his help with the research part of my thesis when I was on my study abroad in India. I would also like to thank Hosei University 法政大学 and professor Masami Iwatsuki for providing me with facilities to work on this thesis during my study abroad in Japan.

Access to computing and storage facilities owned by parties and projects contributing to the National Grid Infrastructure MetaCentrum provided under the programme “Projects of Large Research, Development, and Innovations Infrastructures” (CESNET LM2015042), is greatly appreciated.

# Declaration

I declare that I worked out the presented thesis independently and I quoted all used sources of information in accord with Methodical instructions about ethical principles for writing academic thesis.

........................................

Vít Růžička

Prague, 22 July 2017

Prohlašuji, že jsem předloženou práci vypracoval samostatně, a že jsem uvedl veškeré použité informační zdroje v souladu s Metodickým pokynem o dodržování etických principů při přípravě vysokoškolských závěrečných prací.

........................................

Vít Růžička

V Praze, 22. července 2017

# Abstrakt / Abstract

Tato diplomová práce se zaměřuje na praktické použití konvolučních neuronových sítí na úloze ohodnocování jednotlivých úseků zvolené cesty pro cyklistu. Používáme lokality reálného světa abstrahované do struktury bodů a ohodnocených hran v částečně anotovaném datasetu. Tato původní data obohatíme o fotografickou informaci z lokace pomocí služby Google Street View a o vektorovou informaci objektů v blízkém sousedství z Open Street Maps databáze. Trénujeme model inspirovaný pokrokem v oblasti Počítačového vidění a moderními technikami použitými v soutěži ImageNet Large Scale Visual Recognition Competition. Experimentujeme s různými metodami rozšiřování datasetu a s různými architekturami modelů ve snaze se co nejpřesněji přiblížit původnímu skórování. Používáme též metody přenosu příznaků z úlohy s dostatečně bohatým datasetem ImageNet na úlohu s menším množstvím obrázků, abychom předešli přeučování modelu.

**Překlad titulu:** Odhadování atraktivity cyklistických tras z obrazových dat

This master thesis focuses on practical application of Convolutional Neural Network models on the task of road labeling with bike attractivity score. We start with an abstraction of real world locations into nodes and scored edges in partially annotated dataset. We enhance information available about each edge with photographic data from Google Street View service and with additional neighborhood information from Open Street Map database. We teach a model on this enhanced dataset and experiment with ImageNet Large Scale Visual Recognition Competition. We try different dataset enhancing techniques as well as various model architectures to improve road scoring. We also make use of transfer learning to use features from a task with rich dataset of ImageNet into our task with smaller number of images, to prevent model overfitting.

# Contents

# Codes / Figures

# Chapter 1
# Introduction

This thesis is concerned with the task of evaluating road segments on abstracted map composed of edges and nodes while making use of the real world data available via Google Street View service and Open Street Map. We enrich our initial dataset with imagery and neighborhood information.

We are using Convolutional Neural Network models trained on a small annotated dataset for evaluation of edges with unknown score values. We explore both our task definition as well as the available methodology. Finally we describe our implementation in detail and explain various experiments testing different settings and their efficiency. The resulting evaluated dataset is then visualized as a map overlay.

## 1.1 Structure of the thesis

Section 2 Research contains information about similar tasks as well as usage history of Convolutional Neural Networks.

Section 3 Task describes what we are trying to achieve as well as the dataset available for this task. We also comment on the possible ways of enhancing the initial dataset either with imagery data from Google Street View, or with vector representation of neighborhood collected from Open Street Map database.

Section 4 Method goes deeper into the methodology of Convolutional Neural Networks and discusses the architecture decisions we have made in designing our models. We also cover the topics of model training and evaluation schemes.

Section 5 Implementation explores the actual code solutions and tries to illustrate chosen approaches with pseudocode. We also explore the syntax necessary for model building in framework Keras.

Section 6 Results presents measurements and evaluation of each individual approach we employed to enhance the performance of our models. It also shows the final assessment of our model capabilities and the results visualization.

Section 7 Conclusion closes the topic with final words.

# Chapter 2
# Research

In this chapter we are looking into various approaches taken for our task as well as a brief history overview of the methods we believe are useful for solving our task. Finally we also look at the related works.

## 2.1 Planning

There are many applications for the task to search for the shortest route on graph representation of map. We can abstract many real world scenario problems into this representation and then use many already existing algorithms commonly used for this class of tasks.

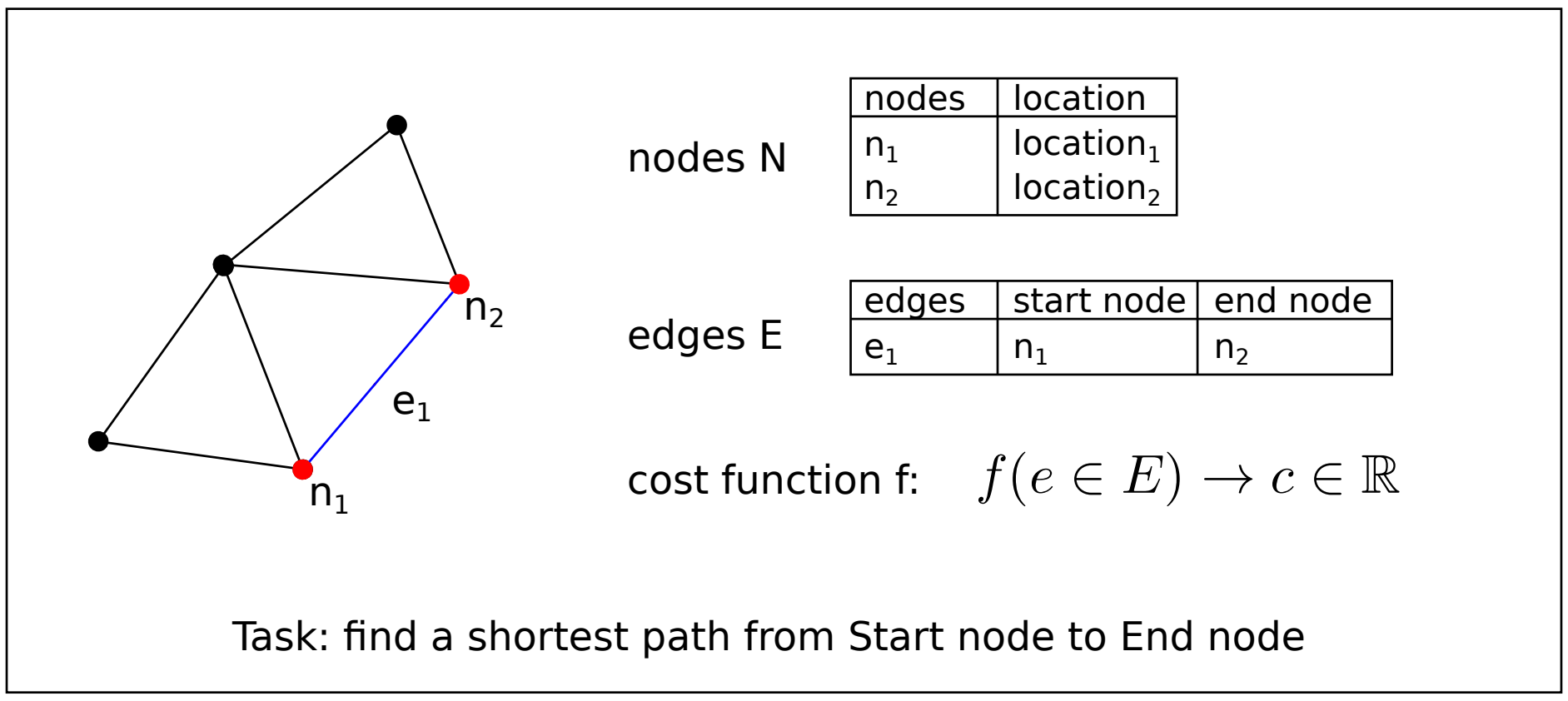

**Figure 2.1.** Abstracted representation of map, description of real world location via set of nodes and edges. This representation allows us to run generic algorithms over it.

We have a set of nodes N, which can be understood as places on map and edges E, which are the possible paths from one node to another. In order to have a measurement of quality of traveling between two nodes, we also need a cost function. This cost function will assign a positive value $c \in \mathbb{R}+$ to each edge. Typically we are facing the task of finding the shortest path, in which we are minimizing the aggregated cost. See illustration of Figure 2.1.

In real life scenario, road segments exhibit many different parameters which influence how fast we can traverse them. There are more or less objective criteria such as surface material, size of the road, time of the day, and criteria which depends solely on the preference of driver. What is the surrounding environment, what is the comfort level of the road, etc.

This tasks gets more interesting, when we look at more complicated examples, where the cost function is multi-criterial. In such case we need more data at our disposal and we can also expect the model to be more computationally difficult. For practical use, we need to use effective algorithm with speed up heuristics (see [1]). Great part of research is also in the area of hardware efficient algorithms, which would work on maps containing continent-sized datasets of nodes and edges, and yet coming up with solution in real time. We can imagine such necessity on hand-held devices of car gps.

### 2.1.1 Data collection

Cost function can be very simple, but in order that it works on real life scenarios, we usually need more complicated one with lots of data recorded. Estimation of how much "cost" we associate with one street (represented by edge $e \in E$ connecting two crossroads represented by nodes) should reflect how much time we spend in crossing it.

In case of planning for cars we generally just want to get across as fast as possible, or to cover minimal distance. When the user is driving a bicycle, more factors become relevant. We need to know the quality of terrain, steepness of the road, amount of traffic in the area and overall pleasantness of the road. In many cases the bikers will not follow the strictly shortest path, choosing their own criteria, such as for example stopping for a rest in a park.

Some of these these criteria are measurable and objectively visible in the real world. For these we need to have highly detailed data available with parameters such as the quality of road and others. Other criteria are based on subjective, personal preference of some routes over other and for these we might need a long period of traces recording which routes have users selected in past. For example the work of [2] makes heavy use of user recorded traces.

See [1] and Figure 2.2 for examples of types of measurements we would likely need to estimate cost of each edge considering the slowdown effect of these features.

| | | |
|---|---|---|
| surface | ∈ | [cobblestone, compacted, dirt, glass, gravel, ground, mud, paving stones, sand, unpaved, wood] |
| obstacle | ∈ | [elevator, steps, bump] |
| crossing | ∈ | [traffic signals, stop, uncontrolled, crossing] |
| cycleway | ∈ | [lane, shared busway, shared lane] |
| highway | ∈ | [living street, primary, secondary, tertiary] |

**Figure 2.2.** Example of the categories of highly detailed data we would require for multi-criteria cost function formulation as presented by [1]. List of features contributing to a slowdown effect on route segment.

In any case highly qualitative, detailed and annotated dataset is required alongside with a carefully fitted cost function which would take all these parameters into account. Large companies are usually protective of their proprietary formulas of evaluating costs for route planners. For example [2] makes use of the road network data of Bing Maps with many parameters related to categories such as speed, delay on segment and turning to another street, however the exact representation remains unpublished.

As we will touch upon this topic in later chapters, its useful to realize that this highly qualitative dataset is not always available. We would like to carry information we can infer from small annotated dataset into different areas, where we lack detailed measurements. We are instead using visual information of Google Street View images, which is more readily available in certain areas than the highly qualitative dataset.

## 2.2 History of Convolutional Neural Networks

Initial idea to use Convolutional Neural Networks (CNNs) as a model was introduced in [3] by LeCun with his LeNet network design trained on the task of handwritten digits recognition.

In this section we trace the important steps in the field of Computer Vision which lead to the widespread use of CNNs in current state of the art research. For more detailed overview we recommend [4].

### 2.2.1 ImageNet dataset

Computer Vision research has experienced a great boost in the work of [5] in the form of image database ImageNet. ImageNet contains full resolution images built into the hierarchical structure of WordNet, database of synsets, "synonym sets" linked into a hierarchy.

WordNet is often used as a resource in the tasks of natural language processing, such as word sense disambiguation, spellcheck and other. ImageNet is a project which tries to populate the entries of WordNet with imagery representation of given synset with accurate and diverse enough images illustrating the object in various poses, viewing points and with changing occlusion.

As the work suggests, with internet and social media, the available data is plenty, but qualitative annotation is not, which is why such hierarchical dataset like ImageNet is needed. The argument for choosing WordNet is that the resulting structure of ImageNet is more diverse than any other related database. The goal is to populate the whole structure of WordNet with 500-1000 high quality images per synset, which would roughly total to 50 million images. Even till today, the ImageNet project is not yet finished, however many following articles already take advantage of this database with great benefits.

Effectively ImageNet became the huge qualitative dataset which was needed to properly teach large CNN models.

### 2.2.2 ImageNet Large Scale Visual Recognition Competition

The ImageNet Large-Scale Visual Recognition Challenge (ILSVRC) [6] fueled the path of progress in the field of Image Recognition. While the tasks for each year slightly differs to encourage new novel approaches, it is considered the most prestigious competition in this field. The victorious strategies, methods and trends of each years of this competition offer a reliable looking glass into the state of art techniques. The success of these works and fast rate of progress has lead to popularization of CNNs into more practical implementations as is the case of Japanese farmer automatizing the sorting procedure on his cucumber farm [7].

### 2.2.3 AlexNet, CNN using huge datasets

The task of object recognition has been waiting for larger databases with high quality of annotation to move from easier tasks done in relatively controlled environment, such as the MNIST [8] handwritten digit recognition task. In the work of [9] the ImageNet database was used to teach a Deep Convolutional Neural Network (CNN) in a model later referred to as AlexNet. At the time this method has achieved more than 10% improvement over its competitors in an ILSVRC 2012 competition.

Given hardware limitations and limited time available for learning, this work made use of only subsets of the ImageNet database used in ILSVRC-2010 and ILSVRC-2012 competition. Choice of CNN as a machine learning model has been made with the reasoning that it behaves almost as well as fully connected neural networks, but the amount of parameters of connections is much smaller and the learning is therefore more efficient.

For the competition an architecture of five convolutional and three fully-connected layers composed of Rectified Linear Units (ReLUs) as neuron models was used. Other alterations on the CNN architecture were also employed to combat overfitting, to fine-tune and increase score and reflect the limitations of hardware. Output of last fully-connected layer feeds to a softmax layer which produces a probabilistic distribution over 1000 classes as a result of CNN.

The stochastic gradient descent was used for training the model for roughly 90 cycles through training set composed of 1.2 million of images, which took five to six days to train on two NVIDIA GTX 580 3GB GPUs.

### 2.2.4 VGG16, VGG19, going deeper

Following the successful use of CNNs in ILSVRC2012, the submissions of following years tried to tweak the parameters of CNN architecture. Approach chosen by [10] stands out because of its success. It focused on increasing the depth of CNN while altering the structure of network. In their convolutional layers they chose to use very small convolution filter (with 3x3 receptive field), which leads to large decrease of amount of parameters generated by each layer.

This allowed them to build much deeper architectures and acquire second place in the classification task and first place in localization task of ILSVRC 2014. Similar approach of going deeper with their CNN design was chosen by the works of [11] and [12].

### 2.2.5 ResNet, recurrent connections and residual learning

The work of [13] introduced a new framework of deep residual learning which allowed them to go even deeper with their CNN models. They encountered the problem of degradation, where accuracy was in fact decreasing with deeper networks. This issue is not caused by overfitting as the error was increased both in the training and validation dataset.

Alternative architecture of model, where a identity shortcut connection was introduced between building blocks of the model, allowing it to combat this degradation issue and in fact gain better results with increasing CNN depth.

Their model ResNet 152 using 152 layers achieved a first place in the classification task of ILSVRC 2015.

### 2.2.6 Ensemble models

The state of the art models as of ILSVRC 2016 made use of the ensemble approach. Multiple models are used for the task and final ensemble model weights their contribution into an aggregated score.

The widespread of using the ensemble technique reflects the democratization and emergence of more platforms and public cloud computing solutions giving more processing power to the competing teams of ILSVRC.

### 2.2.7 Feature transfer

The success of large CNNs with many parameters trained on large datasets like ImageNet has not only been positive, it also poses a question - will we always need huge datasets like ImageNet to properly teach CNN models? ImageNet has millions of annotated images and it has been gradually growing over time.

Article [14] talks about this issue and proposes a strategy called feature transfer or model finetuning, which consists of using CNN models trained at one task to be retrained to a different task.

They offer a solution, where similar architecture of CNNs can be taught upon one task and then several of its layers can be reused, effectively transferring the mid-level feature representations to a different task.

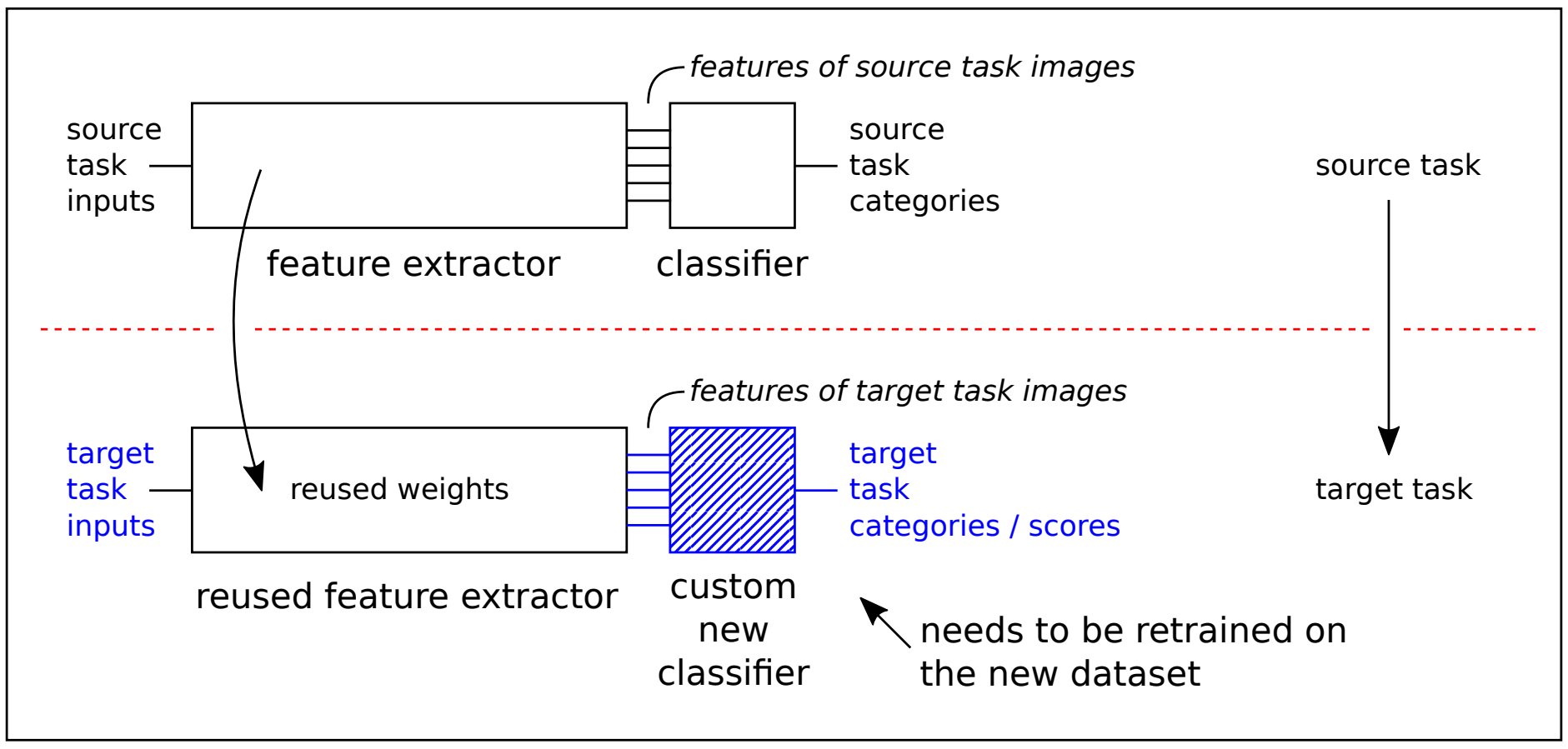

**Figure 2.3.** Illustration of the basic idea of feature transfer between source and target task. Imagine the source task being a classification task on ImageNet and the target task as a new problem without large dataset at its disposal.

This can be used, when the source task has a rich annotated dataset available (for example ImageNet), whereas the target one doesn't. When talking about two tasks, the source and target classes might differ, when for example the labeling is different. Furthermore the whole domain of class labels can be also different - for example two datasets of images, where first mostly exhibits single objects and the second one rather contains more objects composed into scenes. This issue is referred to as a "dataset capture bias".

The issue of different class labels is combated by adding new adaptation layers which are retrained on the new set of classes. The problems of different positions and distributions of objects in image is addressed by employing a strategy of sliding window decomposition of the original image, continuing with sensible subsamples and finally having the result classifying all objects in the source image separately.

The article also works with a special target dataset Pascal VOC 2012 containing difficult class categories of activities described like "taking photos" or "playing instrument". In this case the source dataset of ImageNet doesn't contain labels which could overlap with these activities, yet the result of this "action recognition" task achieves best average precision on this dataset.

This article gives us hope, that we can similarly transfer layers of Deep CNN trained on the ImageNet visual recognition source task to a different target task of Google Street View imagery analysis for cost estimation over each edge segment.

Study of [15] also explores the feature transfer technique, essentially using an already taught CNN as a black box feature extractor for target task.

Articles being submitted to the ILSVRC competition often include a section dedicated on using their designed models on different tasks than what they were trained upon. Effectively they test the models suitability for feature transfering.

Object localization is an example of typical target task, where a large qualitative dataset is not available, yet where the feature transfer technique brings in good results in works of [16] and [17].

### 2.2.8 Common structures

When designing the architecture of custom CNN model, we are using certain layers and building block schemes established as common practice in model building and in the ILSVRC competition. For a new unresearched task it has been suggested (by lecturers and online sources such as [18]) to stick to an established way of designing the overall architecture.

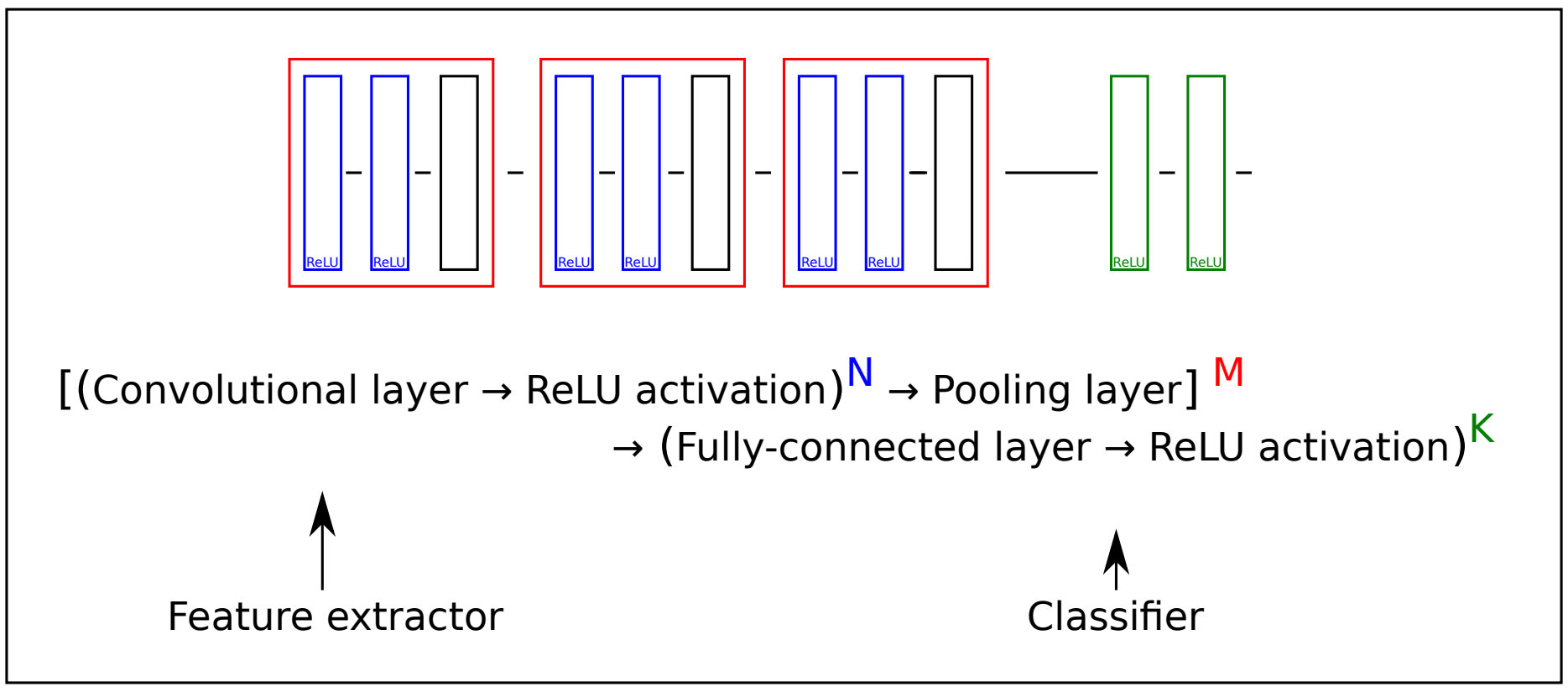

**Figure 2.4.** Formula describing a generic CNN architecture used with image data. Layers are used as building block for more complicated models. While designing a model, we need to keep in mind it's number of parameters.

Refer to Figure 2.4 for illustration of this recommended architecture. Naturally for custom tasks this architecture is later adapted and tweaked to serve well in its specific situation. We will return to this suggested architecture scheme when building our own custom CNN models in 4.1.

## 2.3 Related works

### 2.3.1 Practical application of CNNs

We can find practically applied CNNs on the task of object recognition in several highly specialized datasets, such as plant identification in [19] and bird species categorization in [20]. Both of these works make use of the feature transfer and later CNN fine-tuning. They note the problems of over-fitting on dataset dramatically smaller than the source dataset, which need to be addressed.

Work of [21] uses R-CNN to automate the process of object categorization at specified geological location for the purpose of cataloging. They experiment on small scale area with plans to go US-wide and then cover the whole planet. Objects like trees, lamp posts, mailboxes, traffic lights etc. can be targeted. Main focus is on the automation of processes which otherwise require a lot of manual labor.

### 2.3.2 Geolocation

Large research field is the so called geolocation, assigning a location to images. Instead of answering the "what" of object identification task, we are asking "where" and in the work of [22] eventually also "when". Article [23] names this task "proximate sensing" using an extensive geolocation referrenced dataset collected from social networks, blogs and other not-reviewed data collections. For comparison they also use Geograph British Isles photograph dataset which was taken with the intention to objectively represent the area of Great Britain and Ireland. They note the influence of photographers intent on the usability of images for datasets for machine learning models and mention the necessity of filtration. We may also note the work of [24] which also deals with the geolocation tasks by the means of CNN model with custom top "NetVLAD" layer.

On the topic of processing publicly available photographs we have the study [25] which focuses on analysis of where the images are taken by producing heatmaps and lists of landmarks ordered by the quantity of photographs and their viability as image datasets. Similarly the study of [26] focuses on mining the community created geotagged images. They cluster sets of publicly available images of the same subject.

Paper [27] makes use of aerial imagery with the street level photography in cross-view manner. They explore the intermediate relational attributes such as the quality of neighborhoods to circumvent the lack of ground level images in certain scenarios. Their algorithm produces a probability distribution of localization over limited map segment. Work of [28] also discusses the problems associated with matching the two datasets of aerial and ground level images.

Note that these datasets could be used as alternative data sources even for our task - instead of just using downloaded images from Street View Image, we could download a set of representative images for nearby area, perhaps getting the feel for the neighborhood. Neighborhoods of those edges with high scores could be a guide of what we are looking for in our bicycle route planning.

As for the problem of temporal localization on top of the task of geolocation, the work of [22] explores the scene chronology and task of time-stamping photos. It focuses on presence of temporary objects, such as poster, signs, street art, etc. The presence of temporal information is usually ignored, which produces what is described as *time chimeras* of objects from different times placed and used together. This is an issue in the case of 3D scene reconstruction from images rather than in our case, however it should be noted that Google Street View images do carry a time-stamp. For most precise information we may prefer the most recent imagery data, however certain locations carry older photographs. A source of possible problems would also be a sudden appearance of images from different time of the year, such as images from Summer while using a dataset from images taken during Winter.

### 2.3.3 Google Street View

We have already mentioned using the Google Street View service [29] to obtain images applicable for our task. There has been extensive research over the application of Neural Networks on tasks regarding the Google Street View itself, such as in [30] where multiple classifiers are used to identify and blur out human faces and license plates on the recorded images for the sake of privacy protection.

Data collected from Google Street view service has also been used in [31] for the task of digit recognition. Using Street View House Numbers (SVHN) dataset introduced by [32] as well as an internal dataset generated from Google Street View imagery, they achieve almost human operator grade performance on digit recognition as well as on reCAPTCHA reverse turing test challenge.

The tool of Google Street View is also being used as a cheap method of in-field visits for social observations in [33] or in [34]. This study measures the applicability of using Street View service for cheap location auditing. While this study needed human intervention in both its generated datasets, an in-person field audit and Street View human operated audit, it is of interest to us. While some labels cannot be conclusively obtained just from the imagery data, other environmental characteristics are easily recognized. Measures of pedestrian safety, such as speed bumps, infrastructure object of public transportation systems and parking spaces could be discovered on the images. This study gives us an idea of what kind of information can be concluded from Street View images by human evaluators.

Another work of [35] makes use of large datasets of Google Street View like images for the task of missing masked out section of photograph recreation, by estimating inter-image similarity and patch of image transplantation suitability.

# Chapter 3
# Task

In this chapter we describe what we want to achieve as well as the dataset we have available. We explore the options to enhance this dataset with additional imagery and neighborhood information. We present the details and explanations of important choices we took in the process of acquiring data.

## 3.1 Route planning for bicycles

The task we are faced with consists of planning a route for bicycle on a map of nodes and edges. We are designing an evaluation method, which will give each edge segment appropriate cost. In such a way we are building one part of route planner, which will use our model for cost evaluation and fit into a larger scheme mentioned in 2.1. We are building a machine learning model, such as the one jokingly described in xkcd comic strip [36].

As has been stated in 2.1.1 a cost function can be an explicitly defined formula depending on many measured variables. Similar formula has been used by the ATG research group (such as in [1]), which produced a partially annotated section of map with scores of bike attractivity. We want to enrich this dataset with additional visual information from Google Street View and with vector data from Open Street Map.

We want to train a model on the small annotated map segment and later use it in areas where such detailed information is not available. We argue that Google Street View and Open Street Map data are more readily obtainable, than supply of highly qualitative measurements.

## 3.2 Available imagery data

### 3.2.1 Initial dataset

We are given a dataset from the ATG research group of nodes and edges with score ranking ranging from 0 to 100. Score of 0 denotes, that in simulation this route segment was not used and value 100 means that it was a highly attractive road to take. We rescale these to the range of 0 to 1.

Each node in supplied with longitude, latitude location, which gives us the option to enrich them with additional real world information. Figure 3.1. shows the structure of initial data source.

### 3.2.2 Google Street View

As each edge segment connecting two nodes is representing a real world street connecting two crossroads, we can get additional information from the location. We can download one or more images alongside the road and associate it with the edge and it's score from the initial dataset.

We are using a Google Street View API which allows us to generate links of images at specific locations and facing specific ways.

Edges.geojson

```
{
  "type":"FeatureCollection",
  "features":[
  {
    "type":"Feature",
    "geometry":{
      "type":"LineString",
      "coordinates":[
        [14.434785, 50.07245],
        [14.434735, 50.07255]
      ]                ↖location
    },
    "properties":{
      "length":13,
      "roadtype":"RESIDENTIAL",
      "attractivity":24   ←score
    }
  },
  ... more edges ...
}
```

Nodes.geojson

```
{
  "type":"FeatureCollection",
  "features":[
  {
    "type":"Feature",
    "geometry":{
      "type":"Point",
      "coordinates":
        [14.434785,50.07245]
    },
    "properties":{
      "id":1109,
      "ele":250365
    }
  },
  ... more nodes ...
}
```

**Figure 3.1.** Sample of the structure of initial dataset. We are presented with a GeoJSON file with multiple properties stored for each Feature. Note that we are only interested in the location and score value in Edges file.

### 3.2.3 Downloading Street View images

Google Street View API uses the parameters of location which is the latitude and longitude and heading, which is the deviation angle from the North Pole in degrees. See Figure 3.3. In calculation of heading we are making a simplification of Earth being spherical using formula for initial bearing in Figure 3.2.

$$
\begin{aligned}
y &= \sin(lon_2 - lon_1) * \cos(lat_2) \\
x &= \cos(lat_1) * \sin(lat_2) - \sin(lat_1) * \cos(lat_2) * \cos(lon_2 - lon_1) \\
\Theta &= \operatorname{atan2}(y, x)
\end{aligned}
$$

**Figure 3.2.** Formula for calculation of initial bearing when looking from one location ($\mathrm{lon}_1$, $\mathrm{lat}_1$) to another ($\mathrm{lon}_2$, $\mathrm{lat}_2$).

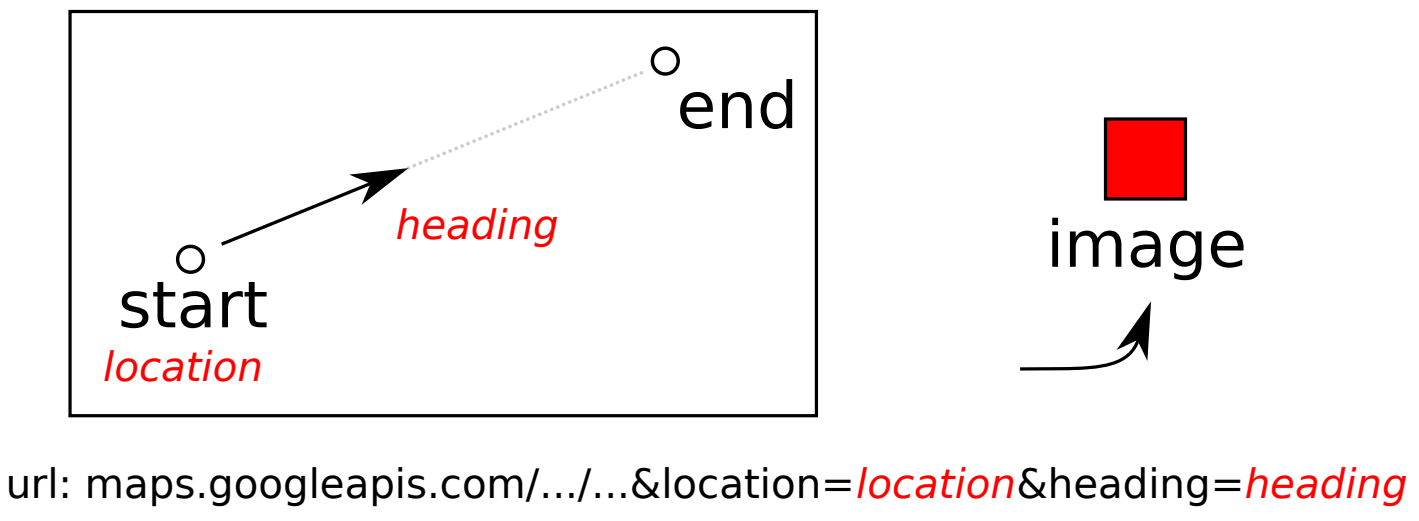

**Figure 3.3.** Illustration of Google Street View API url generation. We are standing in starting location and look in the direction calculated from the relative position of ending point. Note that Google Street View API also requires a private API key to allow for frequently repeated queries.

In order to make good use of the location and collect enough data, we decided to break down longer edges into smaller segments maintaining the minimal edge size fixed. We also select both of the starting and ending locations of each segment. In each position we also rotate around the spot.

We collect total of 6 images from each segment, 3 in each of its corners while rotating 120 ° degrees around the spot. This allows us to get enough distinct images from each location which don't overlap with neighboring edges. See the illustration in Figure 3.4. Note that all of these images will correspond to one edge and thus to one shared score value.

Also note that we are limited to downloading images of maximal size 640x640 pixels as per the limitation of free use of Google Street View API.

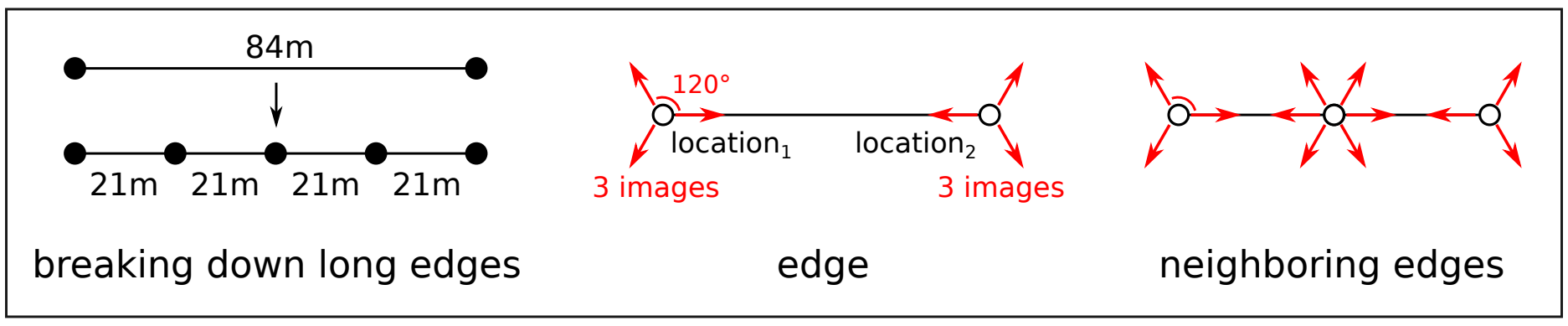

**Figure 3.4.** Splitting of initial possibly large edge segments into sections not smaller than the minimal length limit for edge segment. Each of these generates six images, which are usually not overlapping even with neighboring edges.

In CNNs we often use the method of data augmentation to extend datasets by simple image transformations to overcome limitations of small datasets. We can generate crops of the original images, flip and rotate them or alter their colors.

We will return to the issue of dataset augmentation in 4.5.

## 3.3 Neighborhood data from Open Street Map

We are looking for another source of information about an edge segment in the neighborhood surrounding its location in Open Street Map data.

Open Street Map [37] data is structured as vector objects placed with location parameters and large array of attributes and values. We can encounter point objects, line objects and polygon objects, which represent points of interest, streets and roads, buildings, parks and other landmarks. For more detailed description see [38].

From implementation standpoint, we have downloaded the OSM data covering map of our location and loaded it into a PostgreSQL database. In this way we can send queries for lists of objects in the vicinity of queried location. We will get into more detail about implementation in 5.3.

### 3.3.1 OSM neighborhood vector

The structure of OSM data consists of geometrical objects with attributes describing their properties. In the PostgreSQL database each row represents object and attributes are kept as table columns.

Depending on the object type, different attributes will have sensible values while the rest will be empty. For better understanding consult Table 3.6 with examples of attributes and their values and Table 3.7. for examples of objects in OSM dataset.

For example attribute “highway” will be empty for most objects, unless they are representing roads, in which case it reflects its importance and size. We will be interested in counting attribute-value pairs, for example the number of residential roads in the area, which will have “highway=residential”.

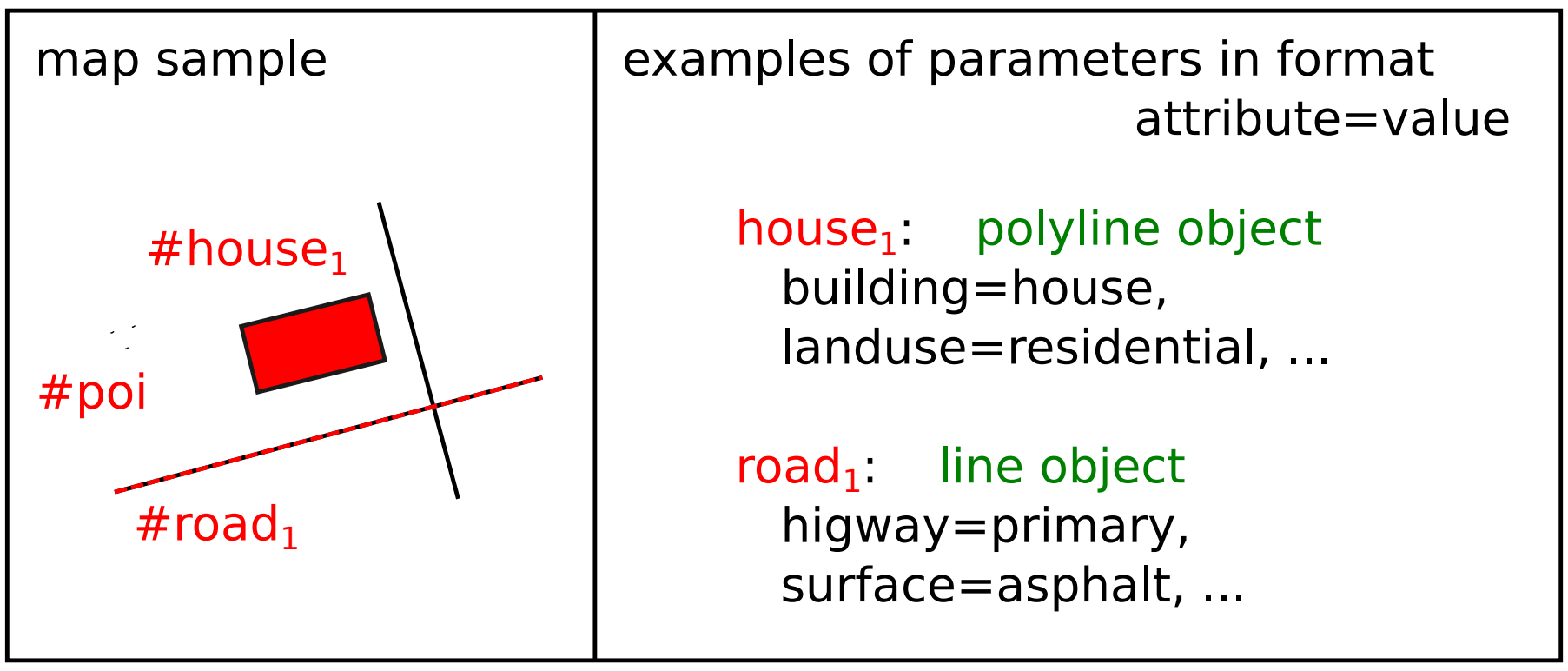

**Figure 3.5.** Example of structure of OSM data with parameters showing its properties. Each object also has a location information and most cases also a non-trivial shape. We will be interested in getting the attribute-value pairs.

| attribute | possible values |
|---|---|
| highway | residential, service, track, primary, secondary, tertiary, ... |
| building | house, residential, garage, apartments, hut, industrial, ... |
| natural | tree, water, wood, scrub, wetland, coastline, tree_row, ... |
| surface | asphalt, unpaved, paved, ground, gravel, concrete, dirt, ... |
| landuse | residential, farmland, forest, grass, meadow, farmyard, ... |

**Table 3.6.** Sample of interesting attributes and their possible values in OSM dataset. Note that there is a large number of attribute and value pairs that can be generated.

| objects id | surface | bicycle | highway | ... |
|---|---|---|---|---|
| *356236228* | paved | ” | footway | ... |
| *356236245* | stone | ” | footway | ... |
| *115927367* | asphalt | yes | cycleway | ... |

**Table 3.7.** Example of values given to a selection of objects from OSM dataset. Note that for some objects we can encounter empty values.

Out of these pairs, we can build a vector of their occurrences. The only remaining issue is to determine which attribute-value pairs will we select into our vector. If we used every pair possible, the vector would be rather large and more importantly mostly filled with zeros.

In order to select which pairs are important, we chose to look at OSM data statistics available at webpage *taginfo.openstreetmap.org* [39]. From an ordered list of most commonly used attribute-value pairs, we have selected relevant pairs and generated our own list of pairs which we consider important.

In Table C.8 we show a more detailed selection of used attribute-value pairs. This table constitutes what the index position of a value in the OSM vector actually signifies.

Then for each distinct location of edge segment, we look into its neighborhood and count the number of occurrences of each pair from the list. See Figures 3.8 and 3.9.

Each distinct location of each edge will end up with same sized OSM vector marking their neighborhood. Note that due to the method of downloading multiple images per one edge, for example by using the same location and rotating around the spot, some of these will have the same OSM vectors.

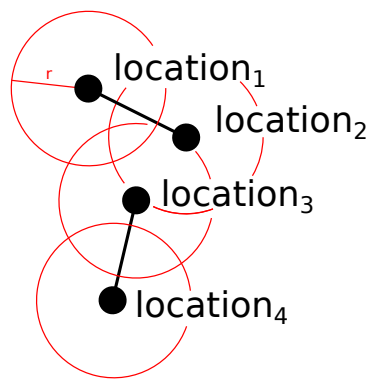

| | 0 | 1 | 2 | 3 | ... |
|---|---|---|---|---|---|
| | higway=primary | building=residential | natural=tree | surface=asphalt | ... |
| location$_1$ | 1 | 3 | 0 | 2 | ... |
| location$_2$ | 2 | 0 | 0 | 1 | ... |
| location$_3$ | 2 | 4 | 1 | 1 | ... |
| location$_4$ | 0 | 1 | 3 | 0 | ... |

**Figure 3.8.** Four unique locations with their corresponding neighborhood vectors. Note that these can be very similar for close enough locations.

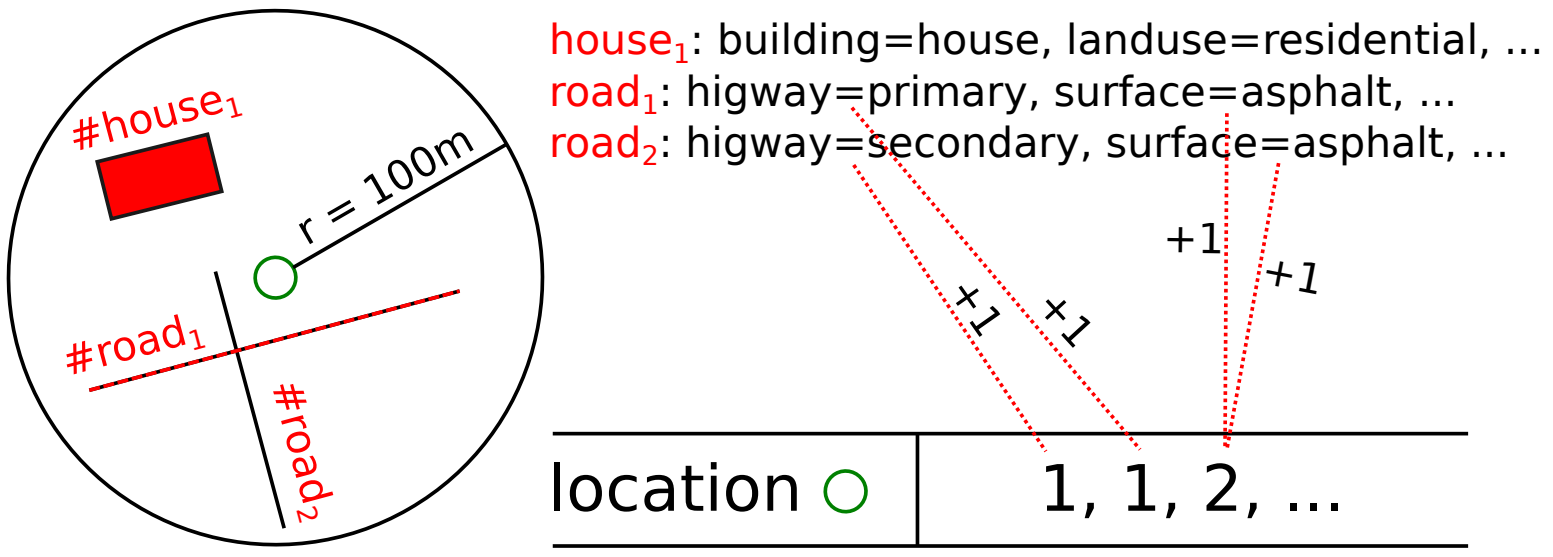

**Figure 3.9.** The construction of neighborhood vector from collection of nearby objects. We need a list of objects in proximity of desired location and then count in their attribute-value pairs. Note that each pair has its own index reserved in the final vector. Only some of the pairs were selected as important, to limit the length of the OSM vector.

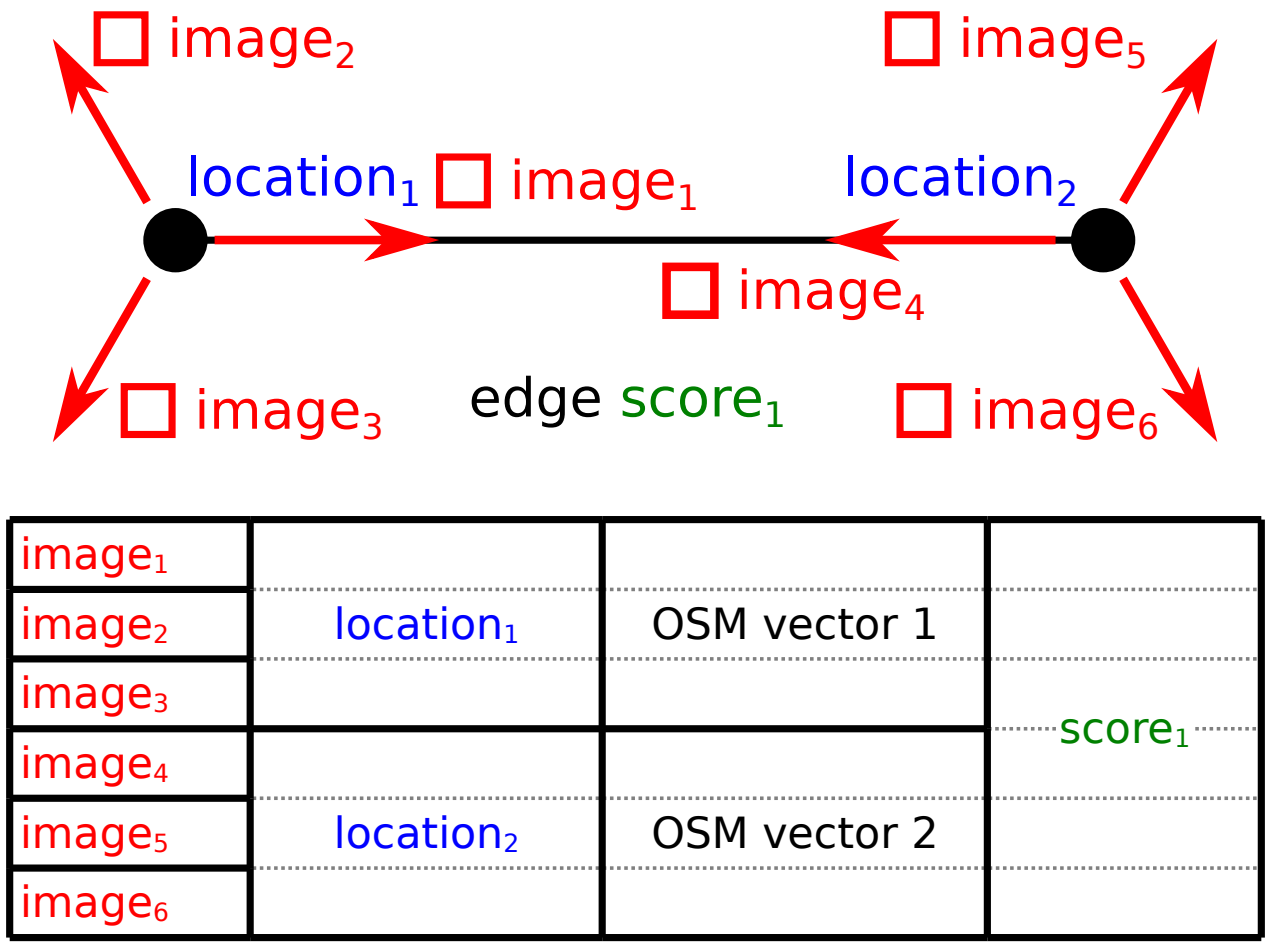

**Figure 3.10.** Example of further not divided edge segment, which generates data entries, where certain values overlap. The score is shared among all images produced from one segment.

One further undivided edge segment contains 6 images which share the same score, and some of which will share location and therefore also their OSM vectors. For illustration see Figure 3.10.

### 3.3.2 Radius choice

Depending on the radius choice, different area will be considered as neighborhood. If we were to choose too small radius, the occurrences would mostly result in zero OSM vector. On the other hand selecting too high radius would lead many OSM vectors to be indistinguishable from each other as they would share the exact same values.

Experimentally we have found radius of 100 meters to be effective.

### 3.3.3 Data transformation

Similar to the spirit of data augmentation for images, we can try editing the OSM vectors in order that they will be more easily used by CNN models.

Instead of raw data of occurrences, we can convert this information into one-hot categorical representations or reduce them into Boolean values. Multiple readings of varying radius size can also be used for better insight of the neighborhood area.

See more about data augmentation in 4.5.

# Chapter 4
# Method

Our method will rely upon using Convolutional Neural Networks (CNNs) mentioned in 2.2 on an annotated dataset described in 3. As we have enriched our original dataset with multiple types of data, particularly imagery Street View data and the neighborhood vectors, we have an option to build more or less sophisticated models, depending on which data will they be using. We can build a model which uses only relatively simple OSM data, or big dataset of images, or finally the combination of both.

Depending on which data we choose to use, different model architecture will be selected. Furthermore we can slightly modify each of these models to tweak its performance.

## 4.1 Building blocks

Regardless of the model type or purpose, there are certain construction blocks, which are repeated in the architecture used by most CNN models.

### 4.1.1 Model abstraction

When building a CNN model, we can observe an abstracted view of such model in terms of its design. Whereas at the input side of the model we want to extract general features from the dataset, at the output side we strive for a clear classification of the image. See Figure 4.1.

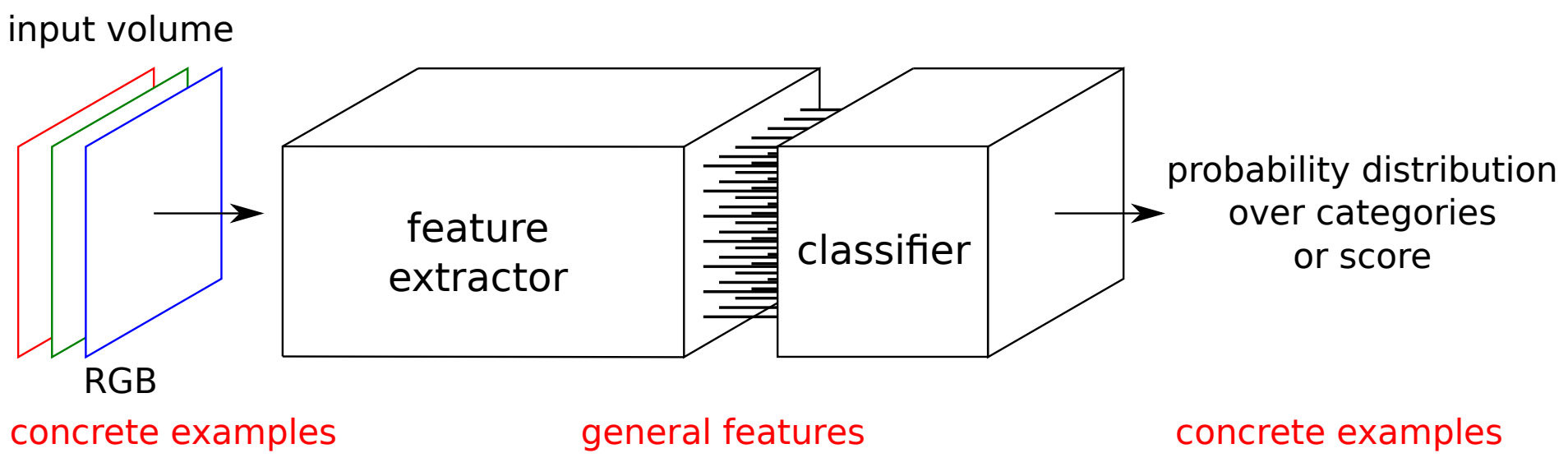

**Figure 4.1.** An abstraction of CNN model into feature extractor section and classificator section. Feature extractor tries to abstract features from the concrete data and classificator projects the features back into concrete category or score data.

Each of these segments will require different sets of building blocks and will prioritize different behavior. Good model design will lead to generalization of concrete task-specific data into general features, which will then be again converted into concrete categories or scores. The deeper the generic abstraction is, the better the model behavior in terms of overfitting will be.

Classification segment transforms the internal feature representation back into the realm of concrete data related to our task. In our tasks we are interested in score in range from 0 to 1 as illustrated by Figure 4.2. We can consider the score to be the probability distribution between two extreme classes, or the whole task as a regression problem.

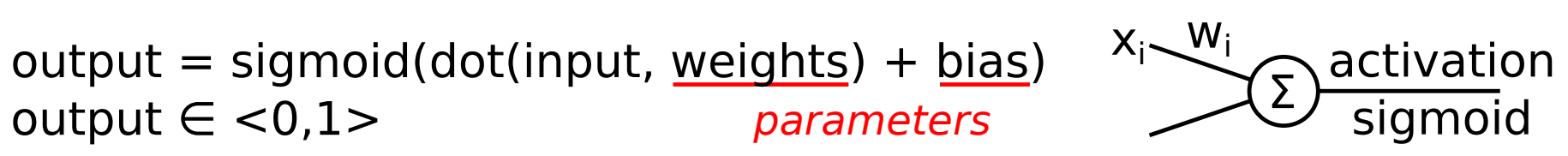

**Figure 4.2.** Classificator section formula describing how we generate one value at the end of CNN model ranging from 0 to 1 as a score estimate.

### 4.1.2 Fully-connected layers

The fully-connected layer stands for a structure of neurons connected with every input and output by weighed connections. In Neural Networks these are named as hidden layers.

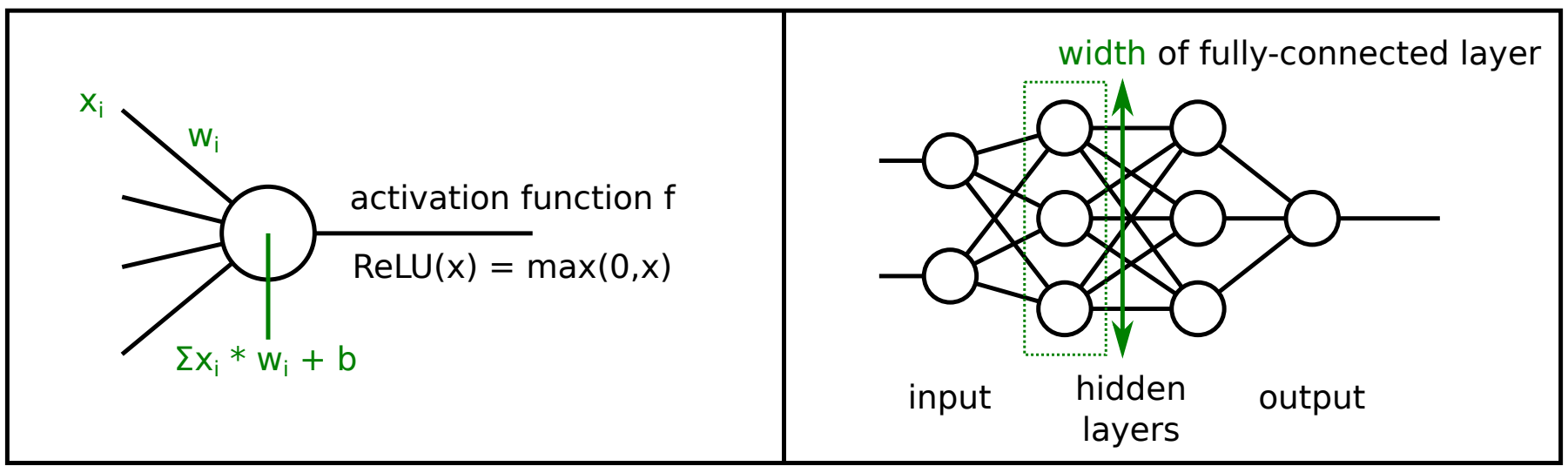

**Figure 4.3.** Shows the model of connections of neurons in fully connected layer. Each neuron has weight for each input and its own bias value. The activation function can be customized, but the ReLU activation function is commonly used. Fully connected layer is composed of a certain number of these neurons, which is defined by the width of the layer.

The fully-connected layer suffers from a large amount of parameters it generates: weights in each connection between neurons and biases in individual neuron units. Fully-connected layers are usually present in the classification section of the model.

### 4.1.3 Convolutional layers

Convolutional layers are trying to circumvent the large amount of parameters of fully-connected layers by localized connectivity. Each neuron looks only at certain area of the previous layer.

For their property to use considerably less parameters while at the same time to focus on features present in particular sections of image, they are often used as the main workforce in the feature extractor section of CNN models.

### 4.1.4 Pooling layers

Pooling layers are put in between convolutional layers in order to decrease the size of data effectively by downsampling the volume. This forces the model to reduce its number of parameters and to generalize better over the data. Pooling layers can apply different functions while they are downsamplig the data – max, average, or some other type of normalization function.

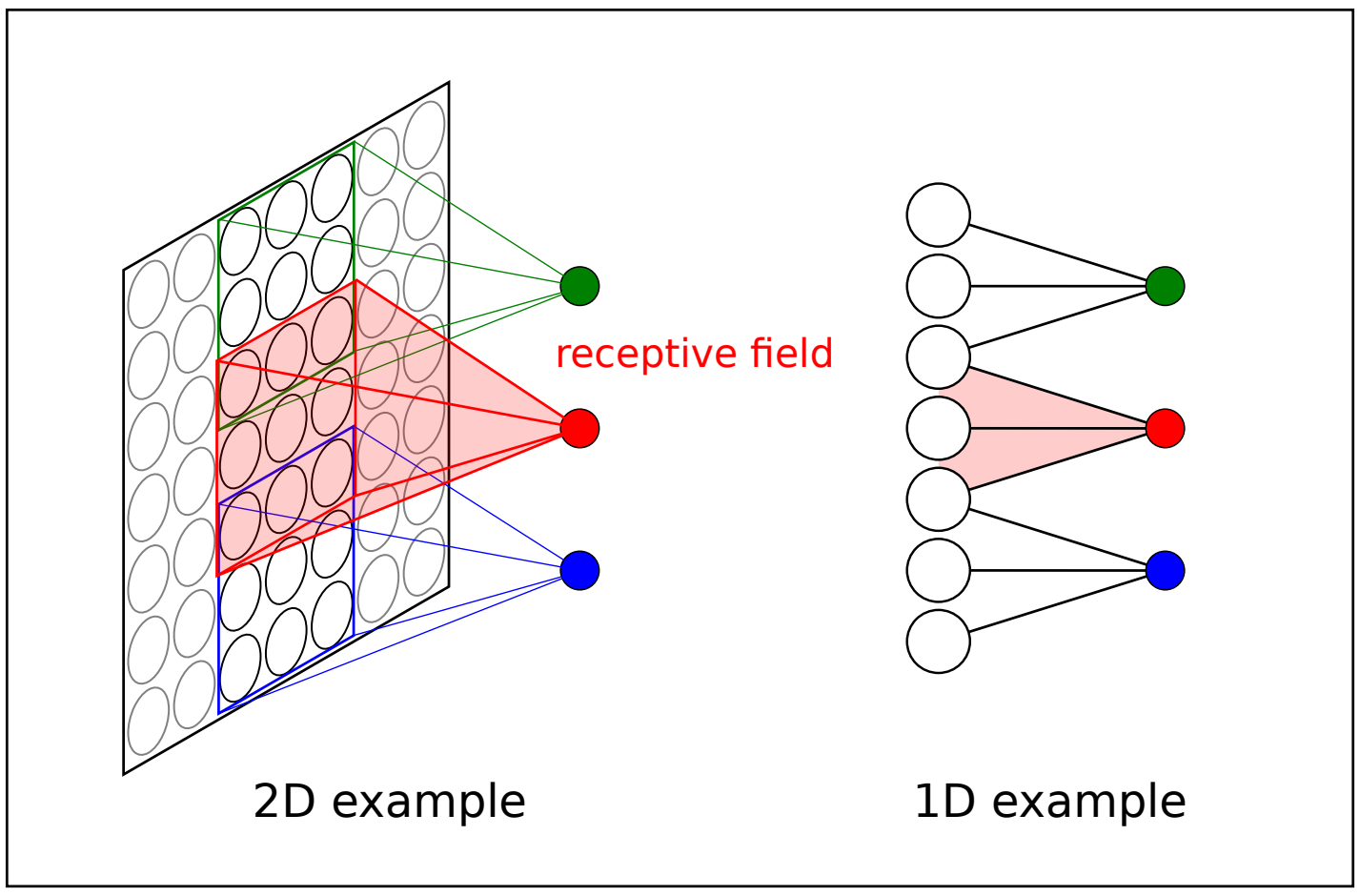

**Figure 4.4.** Schematic illustration of connectivity of convolutional layer. Connections are limited to the scope of receptive field.

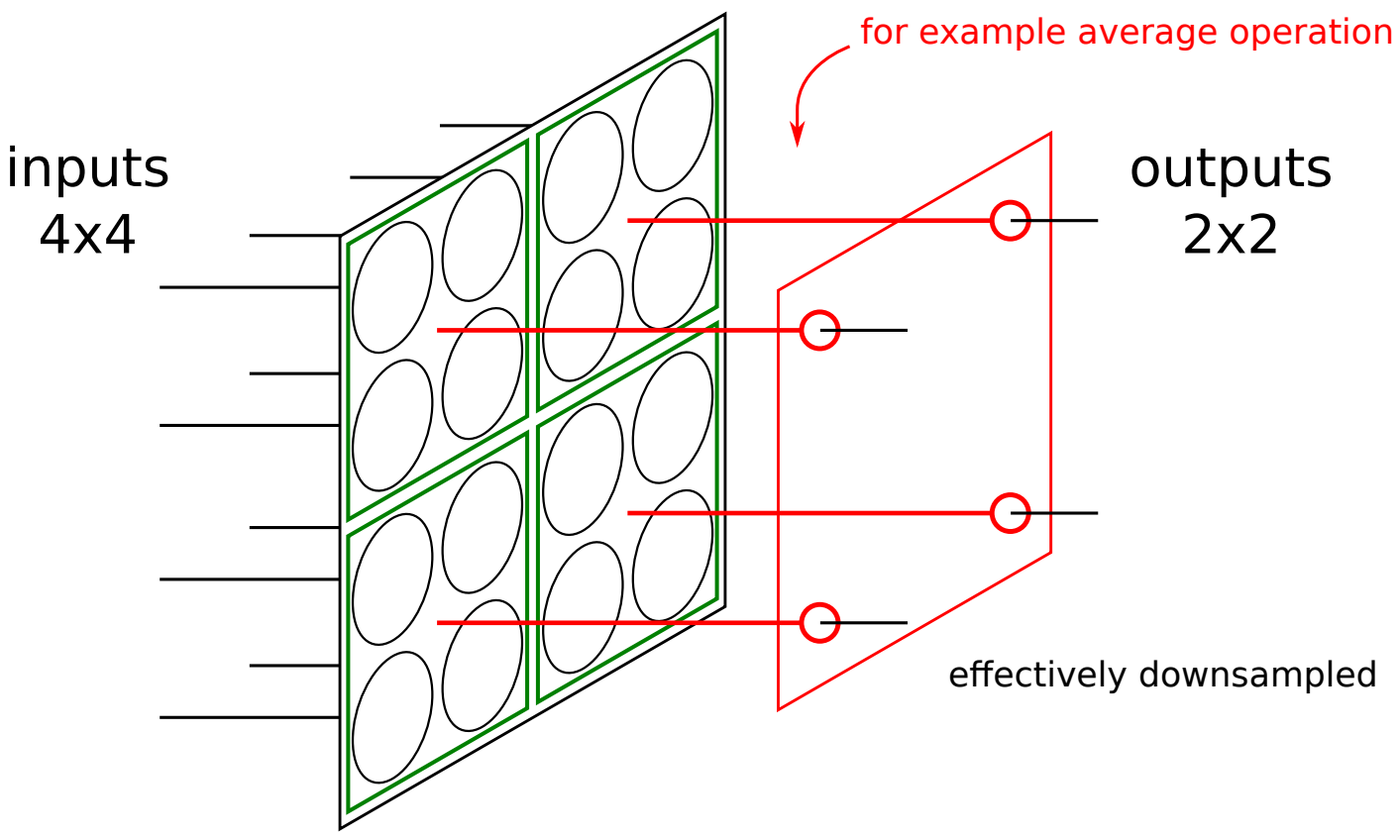

**Figure 4.5.** Pooling layer structure is effectively downsampling the volume of input data. The operation applied when downsampling the data can be customized such as the maximal function in case of Max Pooling layer or average function in case of Average Pooling.

### 4.1.5 Dropout layers

Dropout layer is special layer suggested by [40]. It has become widely used feature in the design of CNN architectures as a tool to prevent model overfitting.

The dropout layer placed between two fully-connected layers functions randomly drops connections between neurons with certain probability during the training period. Instead of fully-connected network of connections we are left with a thinned network with only some of the connections remaining. This thinned networks is used during training and prevents neurons to rely too much on co-adaptation. They are instead forced to develop more ways to fit the data as there is the effect of connection dropping. During test evaluation, the full model is used with its weights multiplied by the dropout probability.

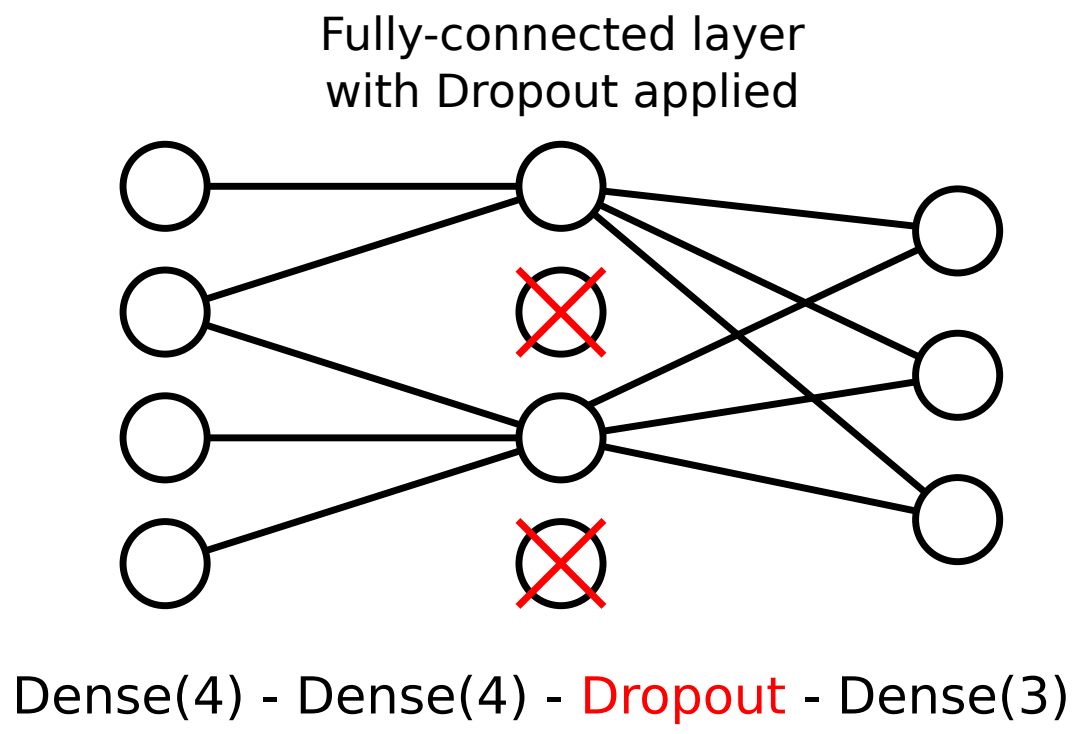

**Figure 4.6.** Illustration of dropout layers effect during the training period, which renders certain connection invalid with set probability. This prevents the network to depend too much on overly complicated formations of neurons.

## 4.2 Open Street Map neighborhood vector model

In this version of model, we broke down edge segments formerly representing streets in real world into regularly sized sections each containing two locations of its beginning and ending location. The unique locations were enriched with OSM neighborhood vectors in 3.3.

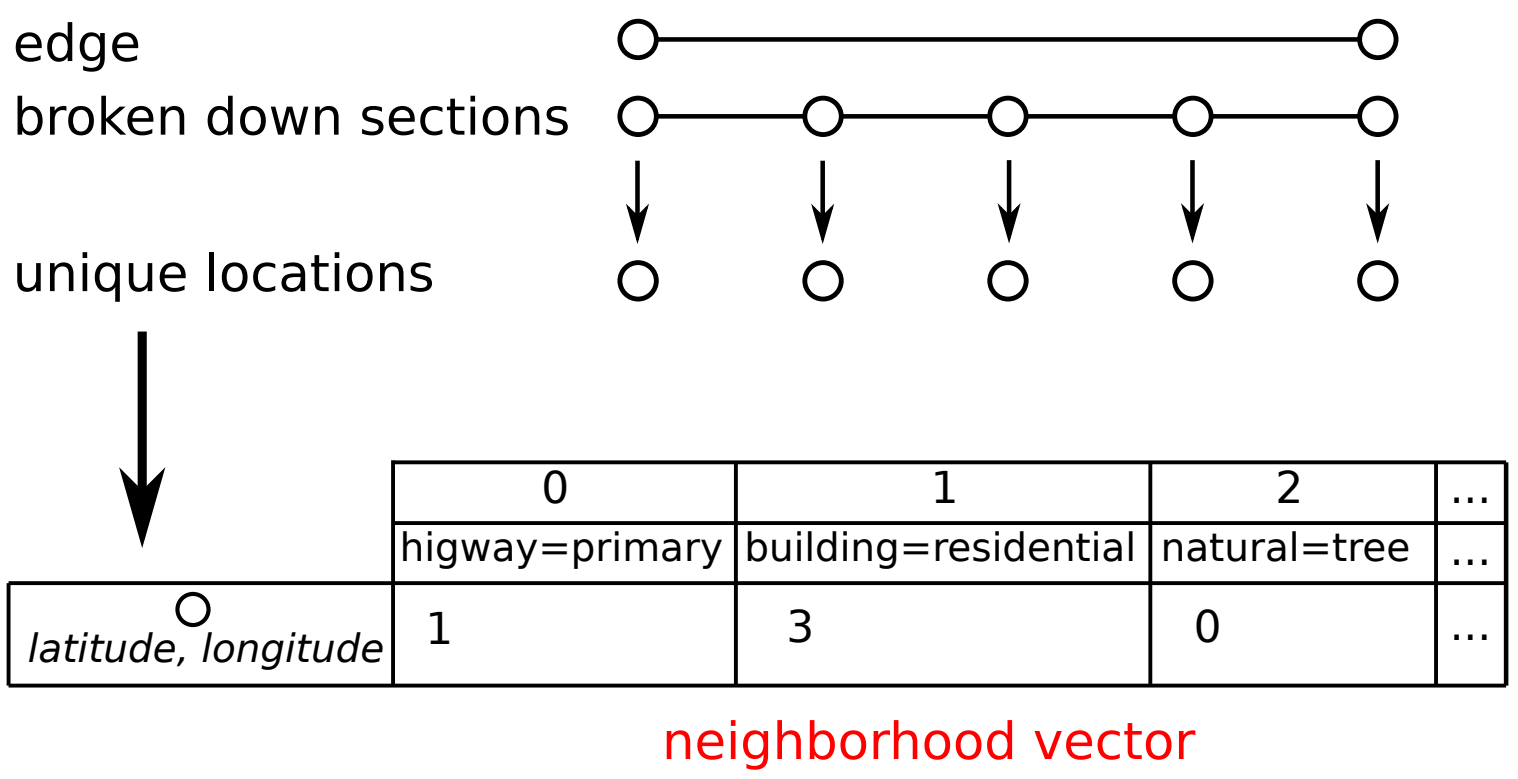

**Figure 4.7.** Edge segment broken down into set of distinct locations, where each of these locations is assigned its own neighborhood vector.

Single unit of data is therefore a neighborhood vector linked to each distinct location of the original dataset. We have designed a model which takes these vectors as inputs and scores as outputs.

The *OSM model* is built from repeated building blocks of fully-connected layer followed by dropout layer. Fully connected layer of width 1 with sigmoid activation function is used as the final classification segment. Figure 4.8 shows the model alongside its dimensions.

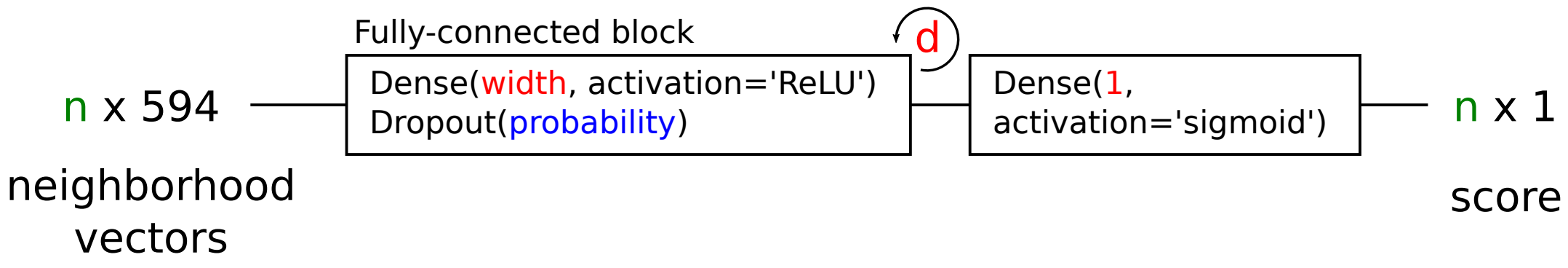

**Figure 4.8.** CNN model making use of only the neighborhood OSM vector data. Size of 594 corresponds to the size of the vectors. As stated in 3.3.1 we have made a selection of frequent attribute-value pairs.

## 4.3 Street View images model

Each edge segment is represented by multiple images captured via the Street View API. Images generated from the same edge segment will share the same score label, however the individual images will differ. We can understand one image-score pair as a single unit of data.

The image data can be augmented in order to achieve richer dataset, see 4.5.

As discussed in 2.2 we are using a CNN model which has been trained on ImageNet dataset. We reuse parts of the original model keeping its weights and attach a new custom classification segment architecture at the top of the model.

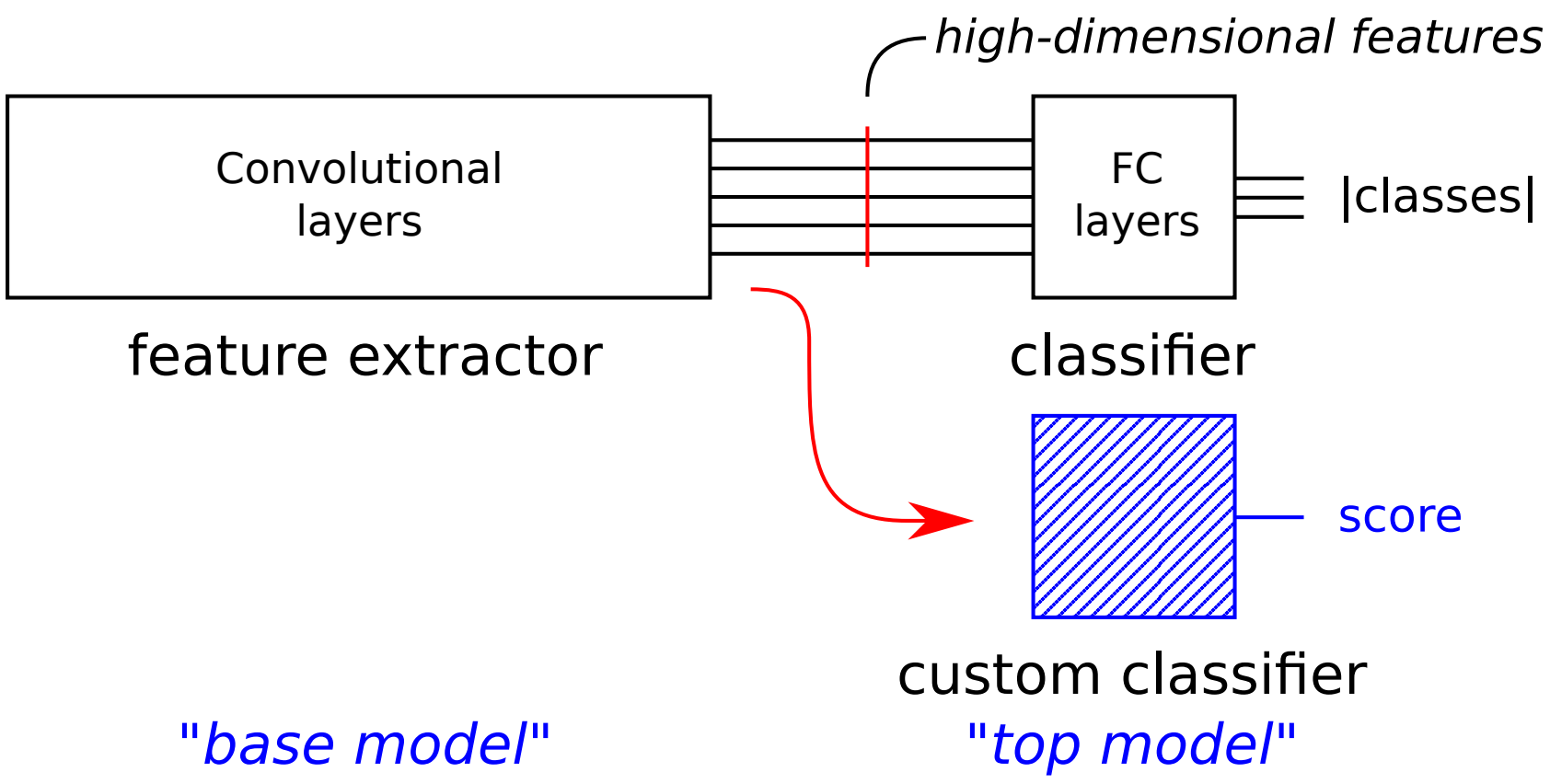

**Figure 4.9.** Reusing part of already trained CNN model as a base model and adding our custom top model. Note that some of the features can be reused in between several runs of the code. We can change the character of the task from classification to regression problem by changing the custom top model.

We can generally divide even the more complicated CNN models into two abstract segments as mentioned in 4.1.1. The beginning of the model, which usually composes of repeated structure of convolutional layers is the base model, followed by a classification section usually made of fully-connected layers.

The former works in extracting high dimensional features of incoming imagery data, whereas the classification section translates those features into a probability distribution over categories or score. In our case we are considering a regression problem model, which works with score instead of categories.

As was mentioned in 2.2.7 we reuse the model trained on large dataset of ImageNet, separate it from its classifier and instead provide our own custom made top model. See Figure 4.9.

We prevent the layers of base model from changing their weights and train only the newly attached top model for the new task. There are certain specifics connected with this approach which we will explore in 4.6.4 section.

### 4.3.1 Model architecture

The final model architecture is determined by two major choices: which CNN to choose as its base model and how to design the custom top model so it's able to transfer the base model's features to our task.

### 4.3.2 Base model

The framework we are working with, Keras, allows us to simply load many of the successful CNN models and by empirical experiments asses, which one is best suited for our task. More about Keras in the appropriate section 5.6.

The output of the base CNN model is data in feature space. The dimension will vary depending on the type of the model we choose, the size of images we feed the base model and also the depth in which we chose to cut the base CNN model.

The remains of base CNN model are followed by a Flatten layer which converts the possibly multidimensional feature data into a one dimensional vector, which we can feed into the custom top model.

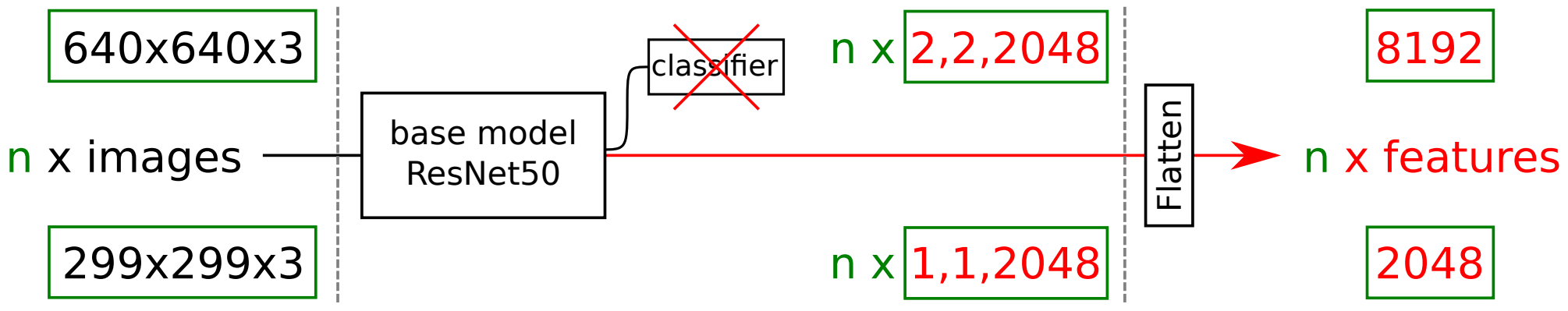

**Figure 4.10.** Example of differently sized images on the input which results in different feature vector size. While the architecture of the base model will adapt itself to arbitrarily sized input, we have to keep in mind the consequences this will have for the following top model.

Base CNN models in Keras are loaded while specifying their `input_shape`. The description files for these models are adjustable, they are defined by a sequence of interconnected layers. Without going deeper into the syntax of Keras (like we later do in 5.6), it's worth noting that the size of input of each layer will influence the size of its output. Pooling and Convolutional layers are simply moving over the input volume data and as such they can adapt to any dimensions, however their output volume will be influenced. Pooling layers function effectively in downsampling their input. Some other layers can have fixed output sizes. Where this matters is in the moment of joining base CNN model with later custom top model. We need to be prepared for the dimension of features captured at the output of base model to be influenced by the size of its inputs. As we can see on example of Figure 4.10, this can lead to vastly different sizes of feature vectors.

### 4.3.3 Custom top model

We feed the feature vector into a custom model built from repeated blocks of fully-connected neuron layers interlaced with dropout layers. The number of neurons used in each of the layers influences the so called model “width” and the number of used layers influences the model “depth”. Both of these attributes influence the amount of parameters of our model. We can try various combinations of these parameters to explore the models optimal shape.

The final layer of the classification section consists of fully-connected layer of width 1 with sigmoid activation function which weighs in all neurons of the previous layer. See Figure 4.11. Note that in the final model we chose to interlace individual fully-connected layers with dropout layers.

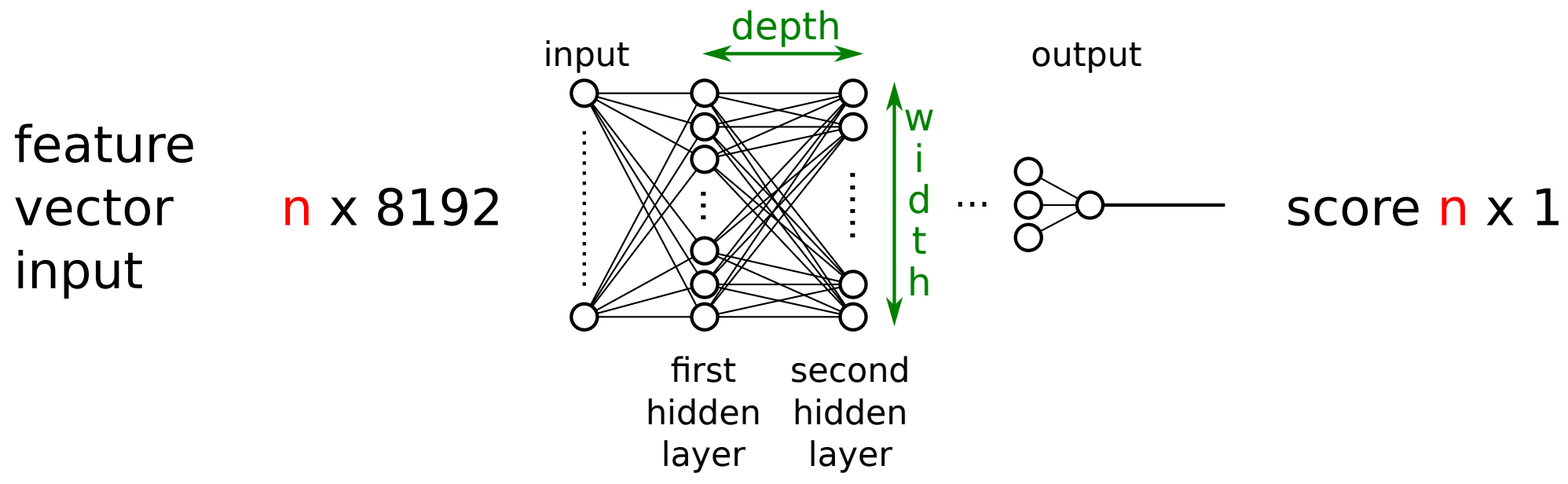

**Figure 4.11.** Top model structure which takes in the feature vector and follows with fully connected layers which are retrained on our own task. Note that in our case we also place dropout layers in between the fully connected layers.

### 4.3.4 The final architecture

The final architecture in Figure 4.12 consists of base CNN model with its weights trained on the ImageNet dataset and of custom classification top model trained to fit the base model for our task.

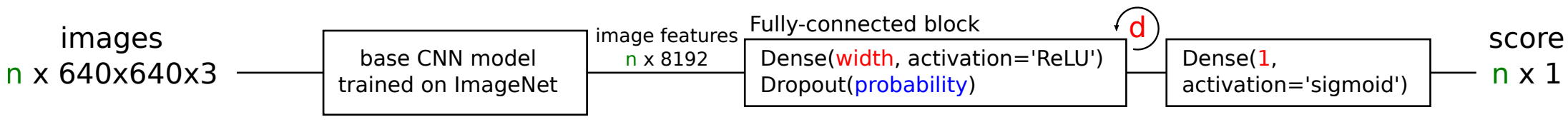

**Figure 4.12.** Final schematic representation of CNN model using Google Street View images. Note that parameters of width and depth can change its performance.

## 4.4 Mixed model

After discussing the architecture of two models making only a partial use of the data collected in our dataset, we would like to propose a model combining the two previous ones.

In this case one segment again generates multiple images, which share the same score and depending on how they are created they could also share the same neighborhood

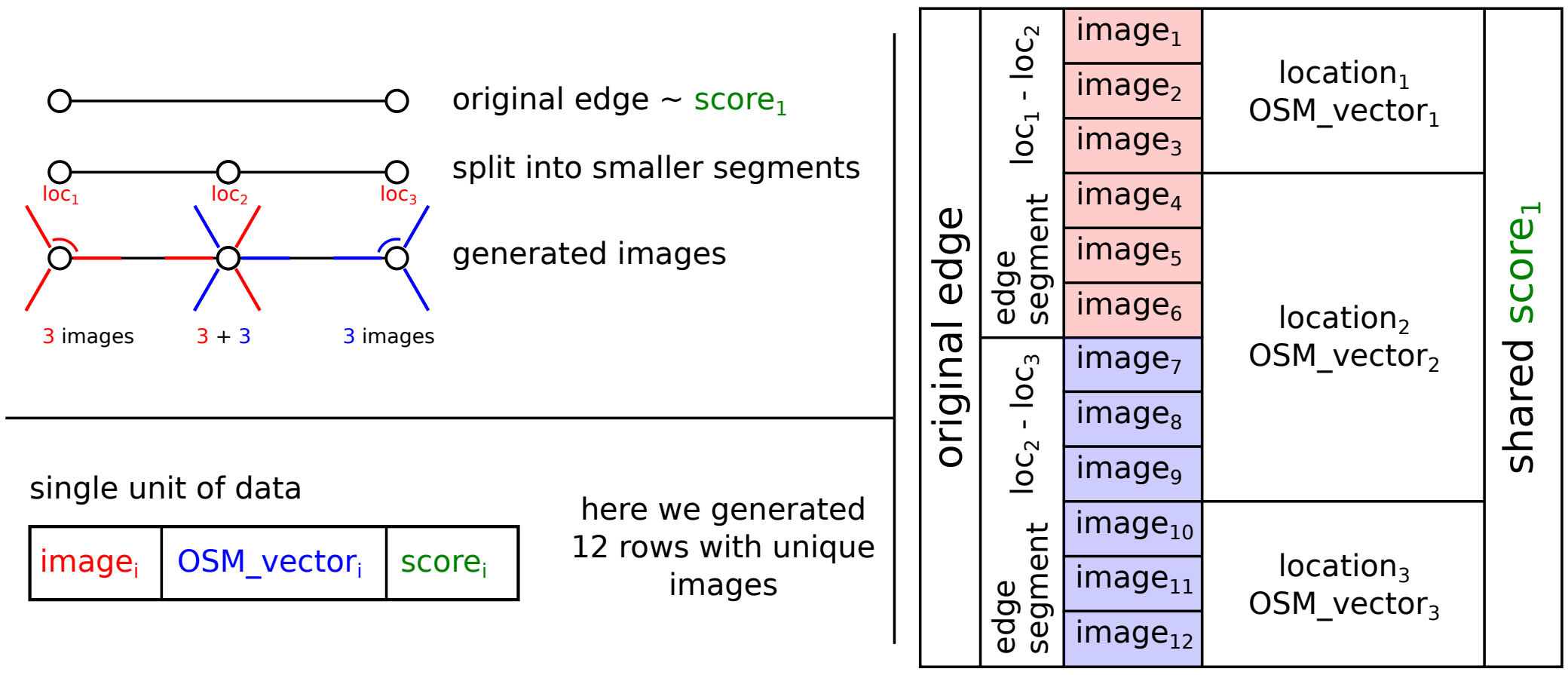

**Figure 4.13.** Example of possible decomposition of original edge into rows of data accepted by the Mixed model. We use both the image data and the OSM neighborhood vectors.

OSM vector representing the occurrences of interesting structures in its proximity. Different edges will generate not only different images, but also different scores and OSM vectors.

As a single unit of data we can consider the triplet of image, OSM vector and score. It's useful to note that later in designing the evaluation method of models, we should take into account, that the neighborhood vector and score can be repeated across data. When splitting the dataset into training and validation sets, we should be careful and place images from one edge into only one of these sets. Otherwise data with distinct images, but possibly the same neighborhood vector and score could end up in both of these sets. Figure 4.13 illustrates how single data units are generated from one edge.

We can join the architectures designed in previous steps, or we can design a new model. We chose to join the models in their classification segment. We propose a basic idea for a simple model architecture, which concatenates feature vectors obtained in previous models and follows with structure of repeated fully-connected layers with dropout layers in between. Concatenation joins the two differently sized one dimensional vectors into one. As is observable on Figure 4.14 we use several parameters to describe the models width and depth.

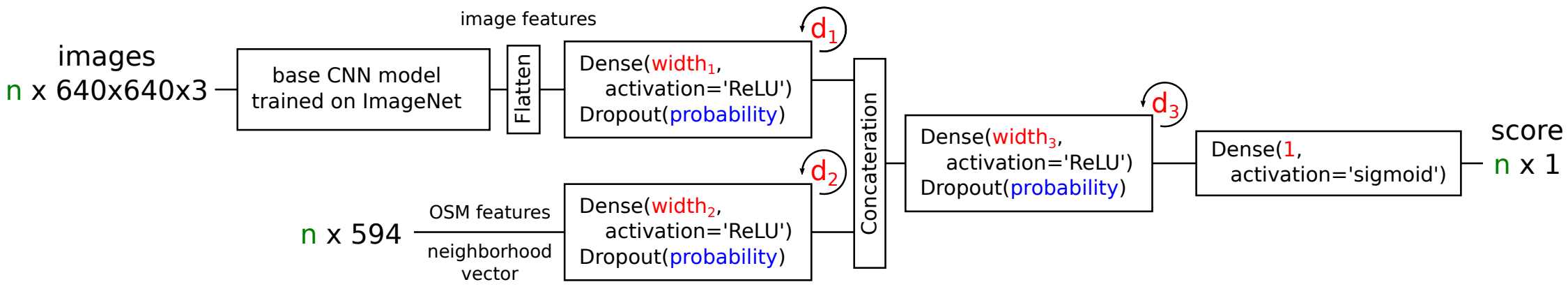

**Figure 4.14.** Final architecture of the Mixed model, which uses both OSM vector and the image data. There are several variables such as the depths and widths which control the shape of the final model as well as the number of it's parameters.

## 4.5 Data Augmentation

When using complicated models with high number of parameters on relatively small datasets, the danger of overfitting is always present. We would like to combat this by expanding our dataset with the help of data augmentation.

Overfitting occurs when the model is basically able to remember all the samples of the training dataset perfectly and incorporate them into its structure. It achieves very low error on the training data, but looses its ability to generalize and results in comparably worse results on the validation set.

The idea of data augmentation is to transform the data we have in order to get more samples and a model which in turn behaves better on more generalized cases.

This cannot be done just blindly, as some of these transformations could mislead our model (for example left to right vertical flip makes sense in our case, but a up side down horizontal flip wouldn't).

There are multiple ways we can approach the problem of generating as many images from our initial dataset as we can. Before getting to the data augmentation aspect, please note, that this is also the reason, why we are generating multiple images per segment. We stand in two corners of each edge segment and rotate $120°$ degrees to get three images on each side. We have also employed a technique, which splits long edges into as many small segments as possible, while not hitting the self imposed minimal edge length. The implementation can be seen in Code 5.3.

Upon inspection of our initial dataset in section C.4, we can see that the edge lengths as presented in Figure C.13 give us plenty of room for splitting too long segments.

It could be debated, that we could rotate for smaller angle or split edges to even smaller segments in order to take advantage of the initial dataset fully. However we came across an issue, that with too small minimal edge length or with different rotation scheme, we obtain very similar images, which actually do not improve the overall performance. This occurs when we don't generate differing enough images during downloading. For actual performance change see chapter 6.2.2.

We face similar issue when choosing a radius for obtaining OSM neighborhood vector as specified in the section 3.3. Instead of selecting one particular radius setting, we can make use of results of multiple queries. When building the OSM vector we would effectively multiply its length by concatenating it with other versions of OSM vectors. We could concatenate the vectors acquired with one fixed radius setting with another version with different radius. See Figure 4.15 for illustration.

Finally we also come across the method of data augmentation by transformation of the original image dataset. Certain operation, such as vertically flipping the image make sense for our dataset.

We show an example of images undergoing such transformation on Figure 4.16. Note that in this particular example we chose vertical flipping alongside with shifting the 90% of the image while making up for the lost 10%. These operations are random and the resulting images are added to create a larger dataset. For the sake of repetition of experiments with the same data, we save these generated images into an expanded dataset.

## 4.6 Model Training

### 4.6.1 Data Split

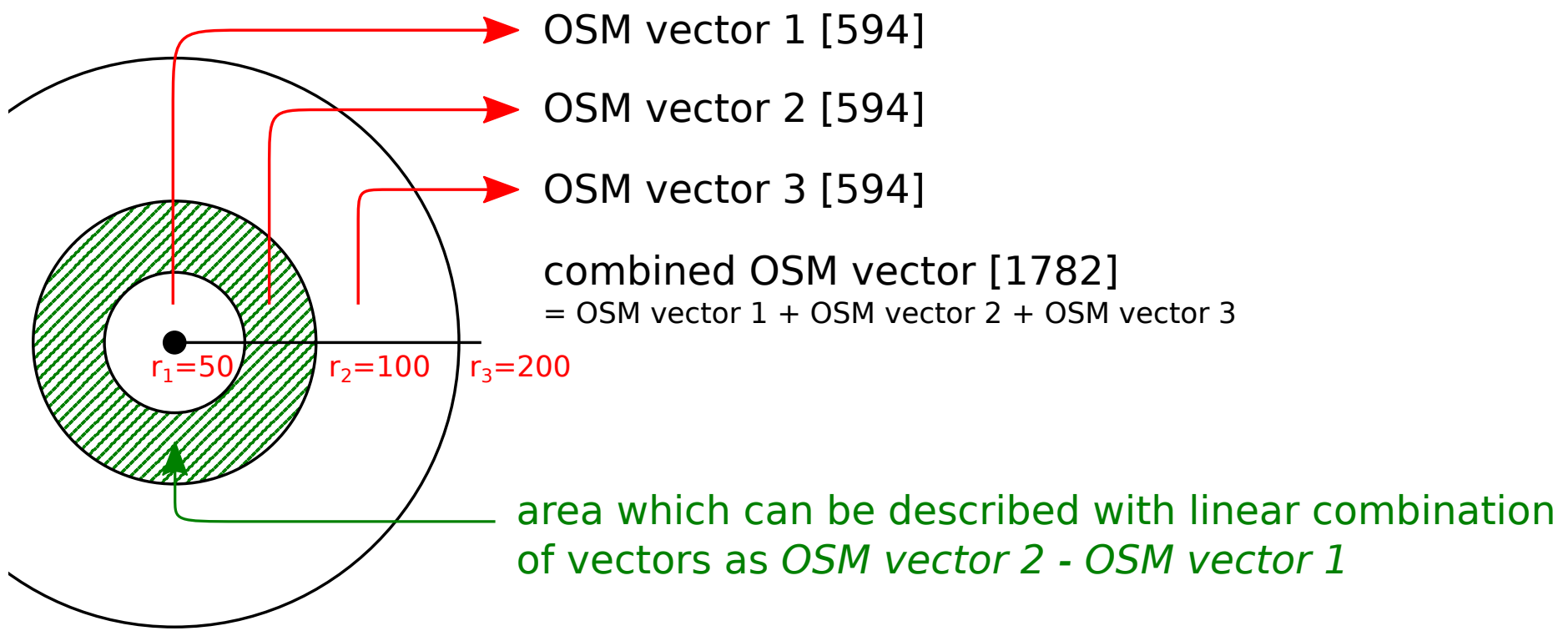

**Figure 4.15.** Illustration of the construction of combined OSM vector and the expected result of more specific area targetting by the model.

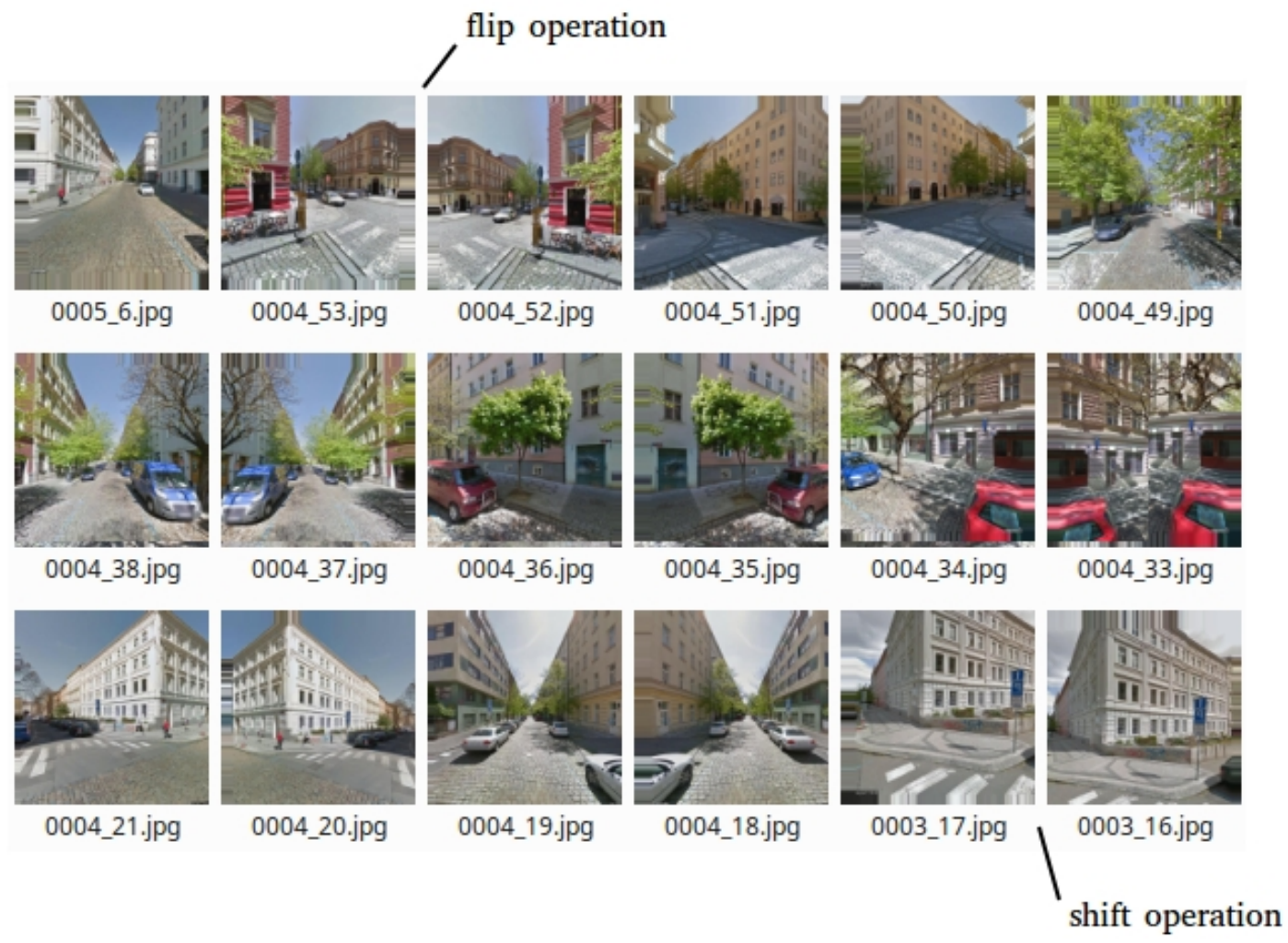

**Figure 4.16.** Data augmentation example. Images can be mirrored or slightly slided in one or both of the axes. We should keep in mind the number of images we are generating in total to limit the complexity of resulting augmented dataset.

Traditionally we split our dataset into two sets – training set, which we use for training of model and so called validation set, which is used only for models performance evaluation. As we discuss in 4.7, we employ more complex strategy of k-fold cross-validation test to obtain more precise results.

The difference between the error achieved on training data and on validation data can be used as a measure of our model overfitting.

### 4.6.2 Settings

We are using backpropagation algorithm to train our models as is supported by the selected Keras framework. We can choose from a selection of optimizers, which control the learning process. We made use of the more automatic optimizers supported by Keras, such as `rmsprop` [41] and `adam` (see [42] where Adam has been tested as an effective CNN learning optimizer). For greater parametric control we can also select the SGD optimizer.

Given the nature of our task, we are solving a regression problem, trying to minimize deviation from scored data. In most models mentioned in 2.2.2, the task instead revolves around selecting the correct category to classify objects.

Accordingly we have to choose appropriate loss function. We have selected the mean squared error metric [43] with the formula given by Figure 4.17.

$$MSE = \frac{1}{n} \sum_{i=0}^{n} (score_i - estimate_i)^2$$

**Figure 4.17.** Mean squared error metric used as loss function when training models.

### 4.6.3 Training stages

As has been discussed in 4.3.1, with image data we are building models that reuse base of other already trained CNN models. In our case we make use of weights loaded from model trained on the ImageNet dataset.

We attach a custom classifier section to a base model and try to train it on a new task. In order to preserve the information stored in connections of the base model, we lock its weights and prevent it from retraining. The only weight values which are changing are in the custom top model. This can be understood as a first stage of training the model as illustrated on Figure 4.18.

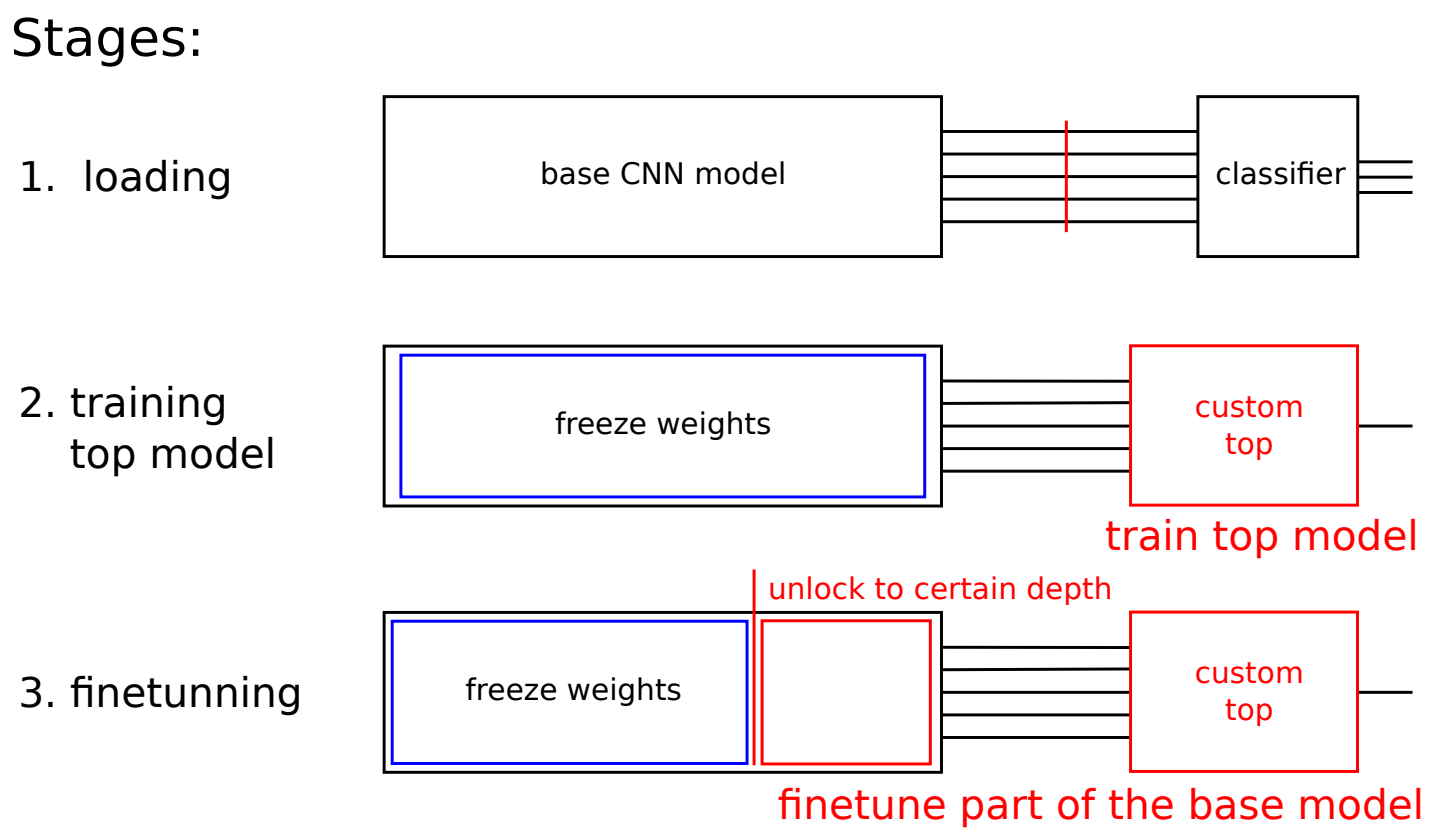

**Figure 4.18.** Training stages illustrated for feature transfer approach. We are reusing values of already trained base model, while adjusting the weights of our custom model. Another stage of finetunning can be used to adjust even some weights of the original base model, however note that it usually is computationally expensive.

This is sometimes followed with a finetunning period, where we unlock certain levels of the base CNN model for training. However this is commonly done with customized optimizer setting such as lowering the learning rate, so that the changes to the whole model weights are not too drastic. We are also usually not retraining the whole model, because of the computational load this would take.

### 4.6.4 Feature cooking

This method is specific to situation when we are training a model with parts, which are frozen, as is in the case of 4.3 image and 4.4 mixed model design. We can take an advantage of the fact that certain section of the model will never change and precompute the image features for a fixed dataset.

In this way we can save computational costs associated with training the model. In the end the original model can be rebuilt by loading obtained weight values.

This allows for fast prototyping of the custom top model, even if the whole model composes of many parameters in the frozen base model. For illustration see figure 4.19.

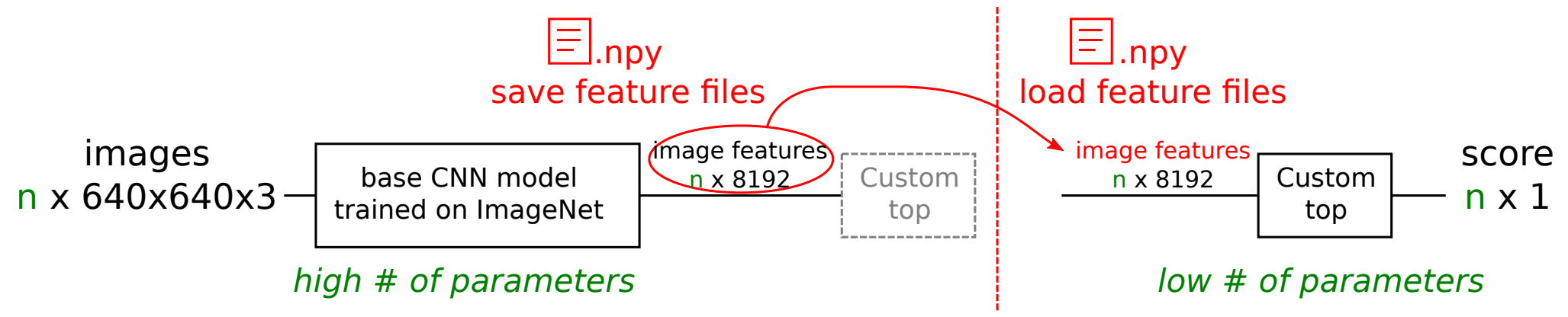

**Figure 4.19.** Reusing saved image features from a file instead of costly computations, which saves time as well as the hardware requirements. We need to maintain the same order of the input data in relation with its labels, which requires a deterministic shuffling of the data.

## 4.7 Model evaluation

As was mentioned in 4.6 we are using the practice of splitting dataset into training and validation dataset, with the k-fold cross-validation technique.

In order to prevent from being influenced by the selection bias, we split our entire dataset into k folds and then in sequence we use these to build training datasets and validation datasets. Every fold will take role of validation set for a model, which will train on data composed from all the remaining folds. Each of these will run a full training ended by an evaluation giving us score. Eventually we can calculate the average score with standard deviation.

This approach obviously increases the computational requirements, because it repeats the whole experiment for each fold. It is not used while prototyping models, but as a reliable method to later generate score. We have chosen the number of folds to be 10, as is a traditionally recommended approach.

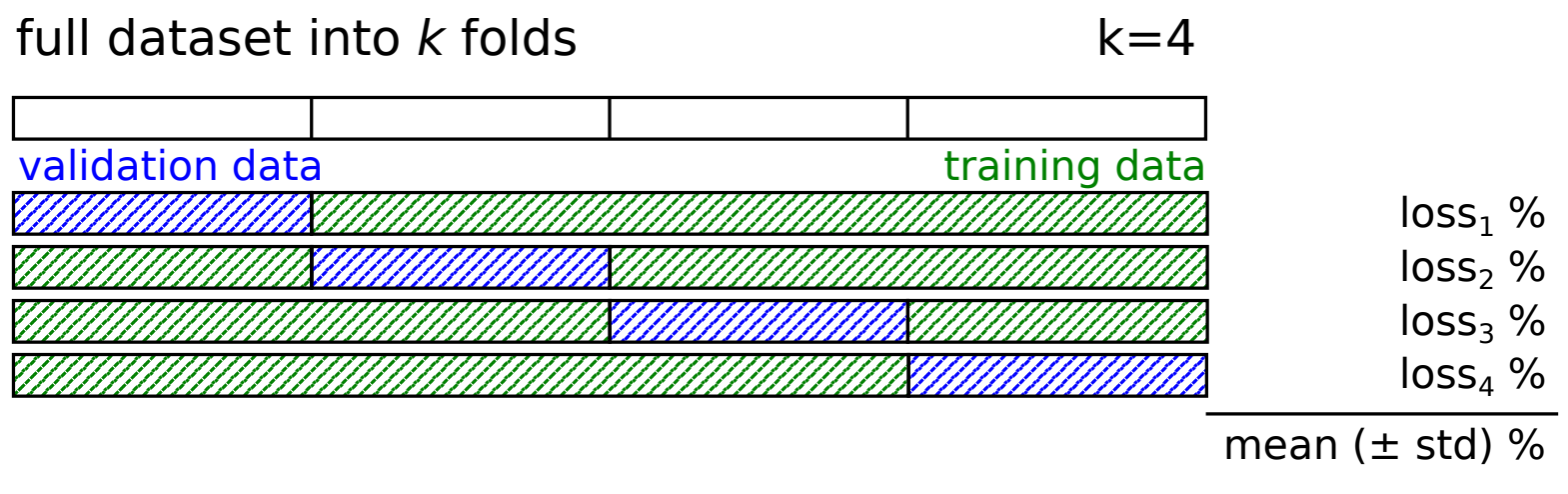

**Figure 4.20.** Splitting schema employed in the k-fold cross-validation. We are given average score alongside with its standard deviation. The chosen value of k will influence the size of the validation data.

# Chapter 5
# Implementation

Details of implementation as well as used frameworks and resources are mentioned in this chapter. We outline the project structure and go deeper into the description of pseudocode samples.

We have used the Keras [44] framework which supports fast model prototyping as well as efficient training and evaluation on custom datasets. We explain how to implement the designed model architectures in 5.8.

We present a pipeline of running experiments with customizable settings in 5.10.

## 5.1 Project overview

In planning of the composition of project code, we have somewhat separated the sections responsible for downloading the data, from those managing the dataset and modeling and running experiments.

Downloader is tasked to download images from Google Street View API to enhance the initial dataset of edges and nodes mentioned in 3.2.1. See 5.2 for the Downloader functionality description. Segment object works as a unit holding information about edge and its corresponding images and score.

However for later processing of the data, we have created a DatasetHandler which contains all the necessary functions. See 5.5.

To run more instances of models and later evaluate them, we have chosen to build individual experiments from custom written setting files in 5.9. Experiment Runner provides the common framework for all these more complicated computations in 5.10. In Settings folder we hold setting files defining each experiment.

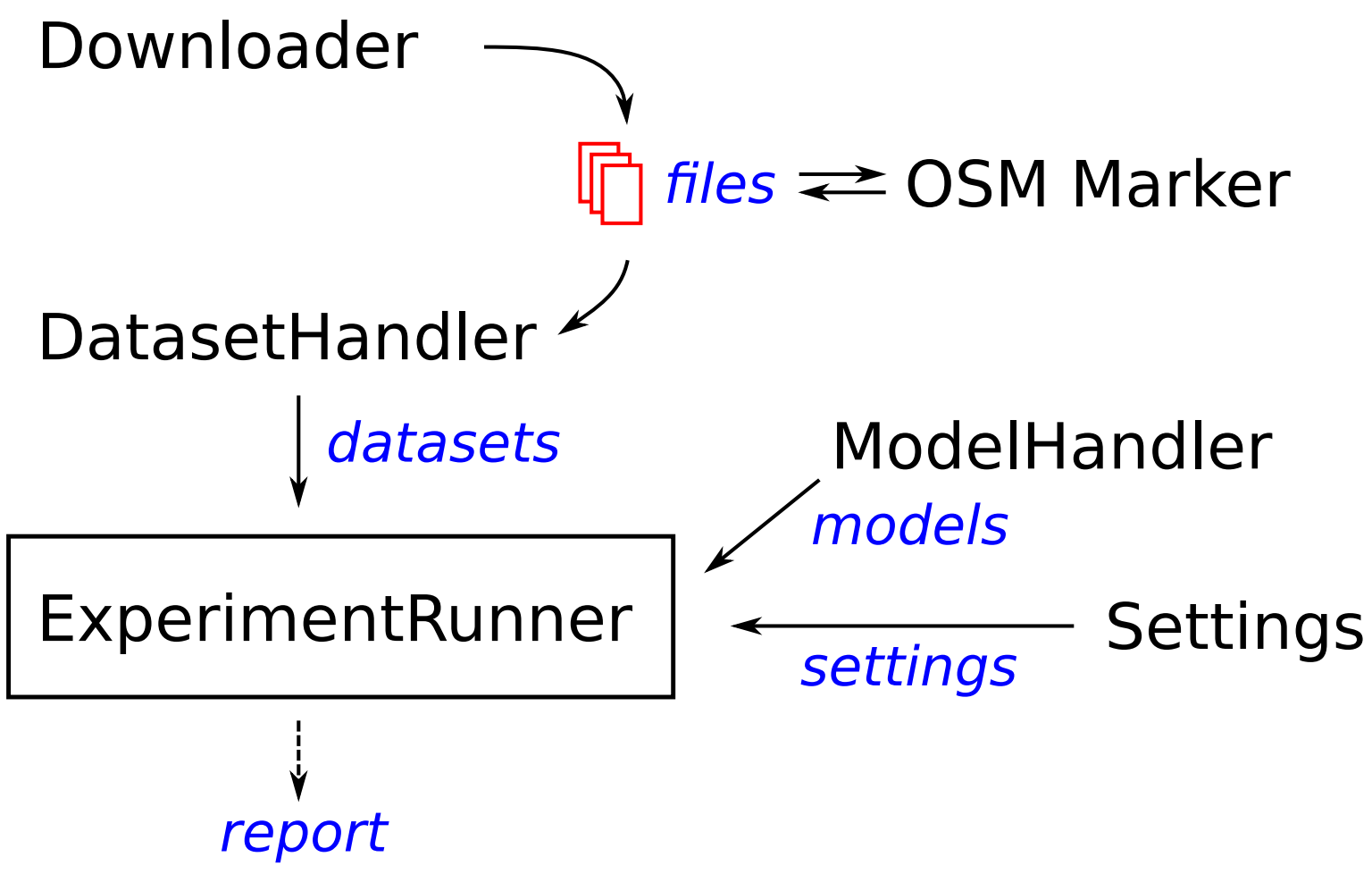

**Figure 5.1.** Project structure in a schema. See the details for each of these boxes in this chapter.

## 5.2 Downloader functionality

The Downloader is the main method of acquiring imagery data and preparation of the dataset. It first creates necessary folders according to the directory path and custom name we provide it with.

`Defaults.py` contains default settings for the downloader, such as default number of times the code should try downloading each image and internal codes with which to mark unsuccessfully downloaded segments. Besides these internal representation settings, it also controls the pixel size of downloaded images.

```
def RunDownload(segments_filename, json_file_path):
# Main downloading function
  Segments <- PrepSegments(json_file_path):
    # parses JSON file
    Segments <- Segment(Start, End, Score, Id)
    return Segments

  FilenameMap <- GenListOfUrls(Segments):
    for Segment in Segments:
      split long Segment into fractions
      for [point_start, point_end] in fractions:
        urls, filenames <- betweenPoints(point_start, point_end)
        # generates three images from standing at point_start
        # and three images from standing at point_end
        return urls, filenames
    # build list of urls and filenames to download

  DownloadUrlFilenameMap(FilenameMap, Segments):
    for [url, filename, segment_id, nth_image] in FilenameMap:
      loaded, error <- url_retrieve_with_retry(url, filename)
      Segments[segment_id].HasLoadedImages[nth_image] = loaded
      Segments[segment_id].ErrorMessage[nth_image] = error

  SaveDataFile(Segments, segments_filename) # save pickled array
```

**Code 5.1.** RunDownload code sample. It is the main function downloading new images from Google Street View service. We end up with a folder of images and generated Segments object saved in a pickled `.dump` file.

Downloading procedure is initiated in `RunDownload()` method (see Code 5.1), which parses node and edge data in provided GeoJSON files and builds an array of Segments.

For each segment in this array, we generate list of images to download. Long segments will be split by interpolating the start and end location to create smaller edge sections (see Code 5.3). We generate a url link in format corresponding to Google Street View API presented in Code 5.2 alongside with an unique file name.

```
http://maps.googleapis.com/maps/api/streetview?size=<width>x<height>
        &location=<lat>,<long>&heading=<angle from north>&key=<api>
```

**Code 5.2.** Exact Google Street View API url we need to generate. Note that we are controlling the dimensions of the image as well as location and the direction in which we are looking. We need a API key for repeating the requests frequently.

One by one we try to download each of these images, marking segments with error flag, if we were unable to access them. The code attempts to retry each request for default number of times, to prevent temporary instability of network to hinder the downloading process.

We also check for invalid images in areas where Google Street View doesn't have any data available, see Figure 5.2.

```
def getGoogleViewUrls(self, min_allowed_distance = 30)
        edge_length = 1000*distance_between_two_points(self.Start, self.End)
        number_of_fractions = max(floor(edge_length / min_allowed_distance),1.0)

        for (last_fraction, current_fraction) in fractions:
            # runs for example with 0.0 - 0.2, 0.2 - 0.4, ..., 0.8 - 1.0
            PointA = interpolation(self.Start, self.End, fraction=last_fraction)
            PointB = interpolation(self.Start, self.End, fraction=current_fraction)

            urls, filenames <- self.betweenPoints(PointA, PointB)

  return urls, filenames
```

**Code 5.3.** Code segment which breaks down long edges and generates lists of urls and filenames Segment.getGoogleViewUrls().

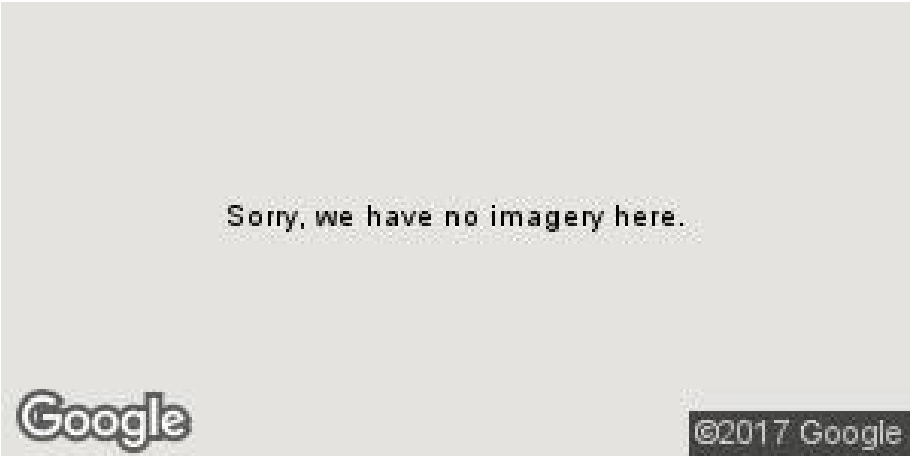

**Figure 5.2.** Location outside of streets with available photographic imagery. We have to detect these images and filter them out.

```
dataset folder: 5556x_example_dataset_299px/
SegmentsData.dump
images/0000_0.jpg
images/0000_1.jpg
...
images/5546_5.jpg
```

**Code 5.4.** Folder structure of dataset. We keep the images downloaded in a separated folder with information of their location and labels in the saved Segments object in `SegmentsData.dump` file. Later we also add the downloaded OSM data into the stored Segments file.

Eventually we save the pickled array into file for further processing. Folder structure is as follows in Figure 5.4.

Function `RunCheck()` in 5.5 allows us to load previously downloaded Segments file and check for erroneous data - for example in case of sudden network failure we can download only the missing files and then save the fixed Segments file.

## 5.3 OSM Marker

Our code works with data downloaded from Open Street Maps to estimate what is the neighborhood of each segment. We have downloaded the .osm data file of corresponding location and exported it into a PostgreSQL database following commands shown in Code 5.6.

Having a PostgreSQL database is advantageous, because it allows us sending repeated queries to database of objects with geographical location while using PostgreSQL macros and functions for getting distance and intersections between areas. PostgreSQL database has hierarchical structure of data storage which is necessary for fast data access. Our python code can access this database with simple queries and then process its responses.

```
def RunCheck(segments_filename):
# Check downloaded segments and fix them
  Segments = LoadDataFile(segments_filename)
  if (HasErroneousData(Segments, ERROR)):
    Segments = FixDataFile_FailedDownloads(Segments, ERROR)
  SaveDataFile(Segments, segments_filename)

def FixDataFile_FailedDownloads(Segments, ERROR):
  BrokenSegments = []
  for (i, Segment) in Segments:
    if Segment.ErrorMesages[i] == ERROR
      BrokenSegments.append(Segment)

  FilenameMapOfBroken <- GenListOfUrls(BrokenSegments)
  DownloadUrlFilenameMap(FilenameMapOfBroken, BrokenSegments)
  return Segments
```

**Code 5.5.** RunCheck code sample checks the downloaded dataset and redownloads missing images. As the downloading itself can take time and there is a daily limitation imposed by the Google Street View API, we want to fix the data on the fly.

```
createdb gisdb
psql -d gisdb -c 'CREATE EXTENSION postgis; CREATE EXTENSION
    postgis_topology;'
osm2pgsql --create --database gisdb example_data.osm.pbf -U example_user
```

**Code 5.6.** Loading data into PostgreSQL database. We are using the Osm2pgsql tool available at [45].

```
def Marker(Segments, radius = 100):
# OSM Marker
  global connection
  connection = ConnectionHandler() # handles Python <> PostgreSQL DB

  for Segment in Segments:
    MarkSegment(Segment, radius)
  return Segments

def MarkSegment(Segment, radius):
  for (i, distinct_location) in Segment.DistinctLocations:
    nearby_vector = connection.query_location(distinct_location, radius)
    Segment.mark_with_vector(nearby_vector, i)

class ConnectionHandler:
def query_location(self, location, radius):
  sql_command = self.sql_cmd_radius(location, radius)
  # builds the query
  # SELECT <interesting_columns> FROM table WHERE distance < radius

  rows, column_names = self.run_command( sql_command )

  pairs = extract_all_pairs(rows, column_names)
  # returns pairs of column names and their values
  # for example "highway=primary", ...

  nearby_vector = [0] * number_of_observed_pairs

  for pair in pairs
    if pair in observed_pairs:
      index = indice_dictionary[pair]
      nearby_vector[ index ] += 1
  return nearby_vector
```

**Code 5.7.** Marking data with OSM vector. We iterate over each Segment in Segments file marking its unique locations.

The code 5.7 loads array of Segments and one by one accesses all the distinct locations stored in each Segment. This could be only two locations for segments too short to be broken down, or more if the original segment was split. Minimally these are the starting and the ending locations of the segment.

We generate a SQL command, which targets particular columns which we chose to observe and also filters the data with `WHERE` clause for objects in distance smaller than chosen radius from segments location. See the actual SQL query on Figure 5.8.

```
SELECT * FROM (
  SELECT ST_Distance(ST_Transform(way, 4326), ST_MakePoint(14.4310,
      50.0631)::geography) AS dist_meters, "shop", "bridge", "amenity", "
      bicycle", "office", "surface", "cutting", "wetland", "waterway", ...
       FROM planet_osm_line
) AS A WHERE A.dist_meters < 50;
```

**Code 5.8.** SQL query for attributes which caught our interest. In this examples the coordinates of `14.4310, 50.0631` constitute for one of the unique locations of edge and value `50` for the radius. Edge has multiple unique locations as per the number of segments we broke it down into.

OSM data we imported into PostgreSQL database takes structure of four tables representing `point`, `line`, `polyline` and `road` objects. We look into each of these tables and combine the results.

Result of query gives us list of rows, each representing one object in vicinity of our location and columns containing attributes. We can produce a "`attribute=value`" pair from the values in each row alongside with the count of their occurrences.

Final neighborhood vector counts all these occurrences in fixed positions as was discussed in 3.3.1.

Each segment stores multiple neighborhood vectors, one for each distinct location. Marked array of Segments is saved into the pickled `.dump` file. We can store multiple versions of these files, each with different radius setting, while reusing all the downloaded images.

## 5.4 Dataset Augmentation

```
image_generator =
    ImageDataGenerator(
  rotation_range = 0,
  width_shift_range = 0.1,
  height_shift_range = 0.1,
  horizontal_flip = True,
  vertical_flip = False
)
```

Expansion scheme

```
image_generator =
    ImageDataGenerator(
  rotation_range = 10.0,
  width_shift_range = 0.2,
  height_shift_range = 0.2,
  horizontal_flip = True,
  vertical_flip = False
  shear_range=0.2,
  zoom_range=0.2,
)
```

Aggresive Expansion scheme

**Code 5.9.** Data augmentation with the Keras ImageDataGenerator syntax.

We made use of the Keras inbuilt object ImageDataGenerator which can be controlled with syntax presented in Code 5.9. We present two schemes of data augmentation. First scheme in which we allow for flipping and image shifting in limited amount of 10% of the whole image.

A more aggressive augmentation scheme is suggested for the second case, where we allow for flipping, shifting up to 20%, rescale of 20% and even shear and rotation of the image by 10 ° degrees.

For each image in our original dataset we create a fixed amount of images with random transformations allowed from this augmentation scheme. For practical reasons of limiting the size of our dataset we chose to generate 2 new images (effectively tripling the size of the original dataset). For reproducibility of the results we stored the expanded dataset into their own folders reusing the same randomly generated data.

## 5.5 Datasets and DatasetHandler

`DatasetObject` is a structure shielding us from low level manipulation with the segments stored in `.dump` file, which can then remain unchanged and be reused for many experiments.

Figure 5.3 shows us the structure of `DatasetObject`, with its most important functions and variables exposed.

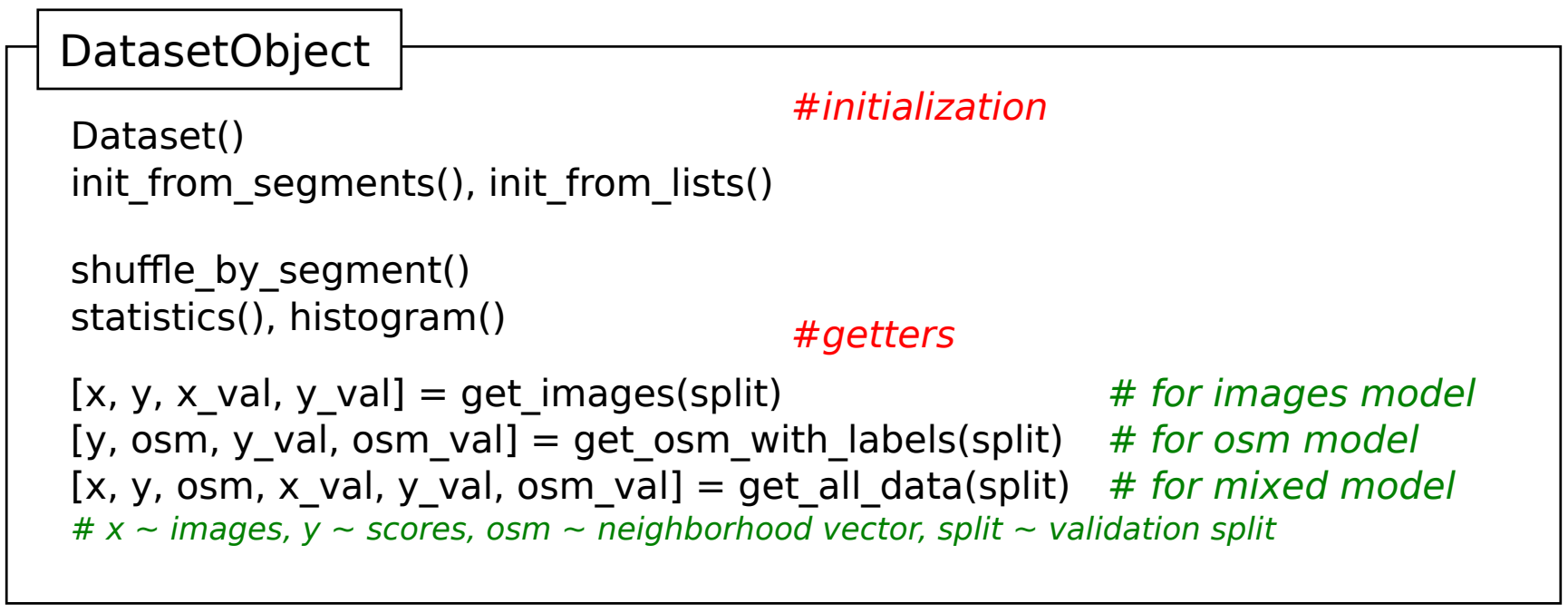

**Figure 5.3.** Dataset object with important functions highlighted.

Dataset is initialized in Code 5.10 by reading a `SegmentsData.dump` file, however it stores data in its own four internal arrays. These are: list of urls to images, labels marking the scores, OSM vectors if we have loaded a marked Segments file and unique IDs of original segments. Note that we are not directly loading all of the images yet, rather we are keeping only their filenames. We can access the images when necessary, saving us from wasteful memory allocation.

We also provide multiple getter functions which split data into training and validation datasets for the purpose of testing experiments. Each model type will require access to different data – image only model will for example omit all the OSM neighborhood vectors.

## 5.6 Keras syntax

Keras [44] is a framework which supports fast prototyping of models with easy API to underlaying backend using low level Theano and TensorFlow. We can build entire

```
def load_dataset(path_to_segments_file):
# part of DatasetHandler which creates the Dataset object
  dataset = Dataset()
  dataset.init_from_segments(path_to_segments_file):
    # loads 4 important lists:
    Segments = LoadDataFile(path_to_segments_file)
    list_of_images, labels, osm, segment_ids = self.load_data_from_segments(
        Segments)
    self.init_from_lists(list_of_images, labels, osm, segment_ids)

  dataset.shuffle_by_segments()
  # shuffles data while respecting that the data from one segment should be next
      to each other

  return dataset

def load_data_from_segments(Segments):
  for Segment in Segments:
    for i_th_image in Segment.number_of_images:
      index = Segment.location_index[i_th_image]

      list_of_images <- Segment.get_image_filename(i_th_image)
      labels         <- Segment.get_score()
      osm            <- Segment.distinct_nearby_vector[index]
      segment_ids    <- Segment.get_id()
  return list_of_images, labels, osm, segment_ids
```

**Code 5.10.** Initialization of Dataset in DatasetHandler from saved segments file.

```
Dataset.statistics()
# prints statistics over data scores
# min |-- 25th_percentile { mean } 75th_percentile --| max

Dataset.histogram()
# generates histogram of data scores
```

**Code 5.11.** Functions for data statistics visualization.

CNN or Machine Learning models by listing individual layers of the model from prebuilt list of commands. For most of models mentioned in 2.2 which performed well in the ILSVRC competition there exists a model description in Keras. Models, which we show how to load in Keras syntax in Code 5.16, have weights available from learning on the large ImageNet dataset. This makes Keras perfect for our task, even for the feature transfer design we intend to use for our models.

Keras also has a training and testing API prepared to use models efficiently with good hardware support including computation on GPUs.

Generic code for building models in Keras works in two modes – Sequencial and Functional. We will use Functional in these particular examples as it is convenient both for writing and understanding.

We will mention some of the layers we can use to build Keras models.

```
Out = Dense(256, activation='relu')(In)
Out = Dense(1, activation='sigmoid')(In)
```

Dense layers stand for fully connected layers mentioned in 4.1.2. Here we can see two possible uses for this type of layer. With width set to `1`, this layer represents one neuron which can be used as the final output of our classifier.

```
Out = Dropout(probability)(In)
```

Dropout layer is used accompanying the Dense layers and follows the process mentioned in 4.1.5. With selected probability the connection of neurons will be dropped during the training period.

```
from keras.models import Model
from keras.layers import Input, Dense, Dropout

def build_generic_model(input_shape):
# part of ModelHandler
  input = Input(shape=input_shape)
  x = Dense(256, activation='relu')(input)
  x = Dropout(0.5)(x)
  ... # more layers defined

  output = Dense(1, activation='sigmoid')(x)
  model = Model(inputs=input, outputs=output)
  return model
```

**Code 5.12.** Building generic model in Keras. This particular model will take input of any size and use Fully-connected layer followed by Dropout layer. The final output will be a single score per each input.

After building the model we need to compile it specifying settings which will be used for training such as the optimizer, loss function and additional measured metrics.

```
Model.compile(optimizer, loss, metrics)
```

For details about which settings we use for training, check 4.6.

Finally our model is trained on dataset with fit command.

```
model.fit(data, labels, validation_data = (data_val, labels_val))
```

The full code to build and train a generic Keras model on a dataset can be seen in Code 5.12 and 5.13.

```
def run_generic_experiment():
  dataset = DatasetHandler.load_dataset(path_to_segments_file)
  [images, score, images_val, score_val] = dataset.get_imgs_with_labels(...)

  input_shape = dataset.get_input_shape()
  model = ModelHandler.build_generic_model(input_shape)

  model.compile(optimizer, loss, metrics)
  # optimizer, loss, metrics,
  # epochs, batch_size are loaded from Settings

  history = model.fit(images, score, epochs, batch_size,
                      validation_data=(images_val, score_val))

  ModelOI.save_history( history )
  ModelOI.visualize_history( history )
```

**Code 5.13.** Fit generic model to data in Keras and then save and plot the history of the training.

## 5.7 ModelHandler

`ModelHandler` is composed of three functional sections. `ModelOI` provides the functions handling input and output – including interaction with `DatasetHandler` and reporting functions. `ModelGenerator` is responsible for building models in Keras syntax depending on settings we choose for current model. `ModelTester` provides functions controlling the training and testing capabilities of the model.

These individual blocks are controlled from `ExperimentRunner`, which ties them together and manages shared Settings information.

Figure 5.4 illustrates the various functions which are part of the `ModelHandler`. In Code 5.14 we can see snippets of code dealing with building and testing models. Note that in one run we are handling multiple instances of models and datasets, as is explained in 5.10.

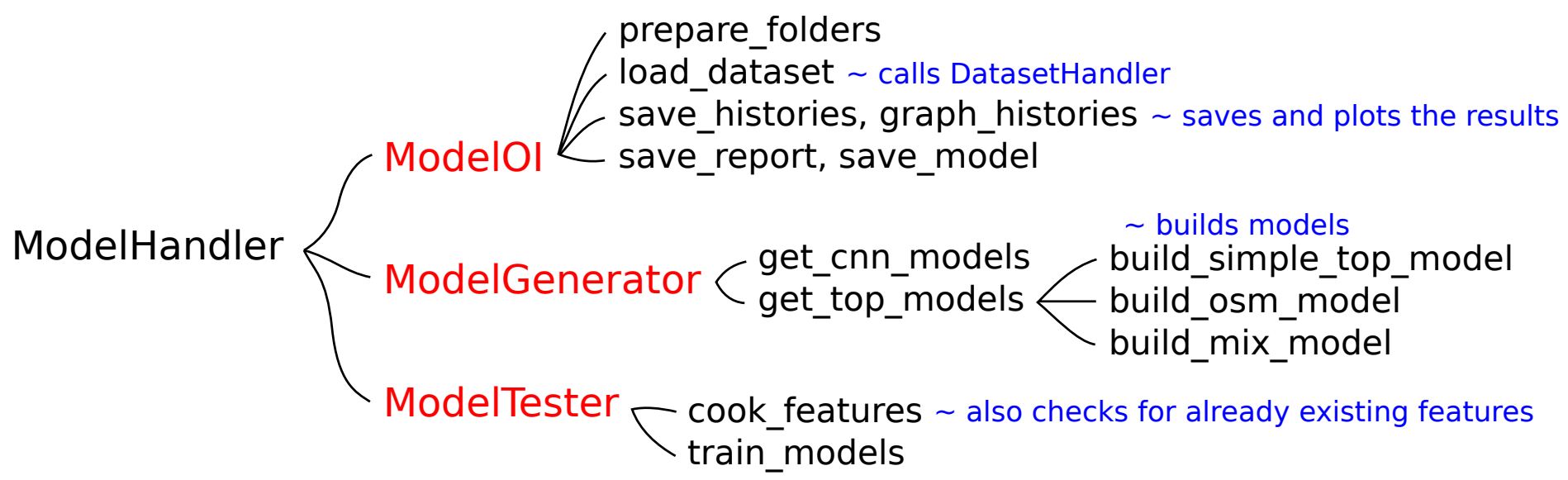

**Figure 5.4.** Illustration of `ModelHandler` structure using its three functional parts - `ModelOI`, `ModelGenerator` and `ModelTester`.

```
def get_top_models():
  models = []

  for model_setting in Settings["models"]:
    model_type = model_setting["model_type"]
    model = build_img_model / build_osm_model / build_mix_model

    models.append( model )
  return models


def train_models():
  histories = []

  for model_setting in Settings["models"]:
    model <- models
    dataset <- datasets

    history = train_model(model, dataset, model_setting)
    histories.append( history )
  return histories
```

**Code 5.14.** `ModelHandler` functions. One experiment run interacts with multiple models each with their individual settings and dataset. Here we can see the pseudocode of model generation and training - we process one by one.

## 5.8 Models

Particular models described in section 4 are shown here with functions which generate them. We have stated that their structure can be parametrized, and that we can alter these parameters to change their shape and performance. We can change the depth and width of these models.

### 5.8.1 OSM model

See Code 5.15, which builds the model discussed in 4.2. As input we are using the OSM neighborhood data. Number of fully-connected Dense layers as well as their width is variable.

```python
from keras.models import Model
from keras.layers import Input, Dense, Dropout

def build_osm_only_model(input_shape, model_depth, model_width=256):
    osm_features_input = Input(shape=input_shape)
    top = Dense(model_width, activation='relu')(osm_features_input)
    top = Dropout(0.5)(top)
    for i in range(0,model_depth-1):
        top = Dense(model_width, activation='relu')(top)
        top = Dropout(0.5)(top)
    output = Dense(1, activation='sigmoid')(top)

    model = Model(inputs=osm_features_input, outputs=output)
    return model
```

**Code 5.15.** Function producing a Keras OSM model described in 4.2.

### 5.8.2 Images model

When we are working with model described in 4.3, we make use of the feature transfer method and as such the training process can be made easier as was discussed in 4.6.4.

Also the Keras syntax will be correspondingly altered. We create two models instead of a single whole model. One represents the feature extractor section of the model and is mostly created by the reused base of pretrained CNN model. We can load these as base of our models with weights initialized from model trained on ImageNet dataset, with syntax presented in Code 5.16.

```python
# Keras Applications, deep CNN models with weights trained on ImageNet

model = ResNet50(weights='imagenet', include_top=False, input_shape=input_shape)

model = VGG16(weights='imagenet', include_top=False, input_shape=input_shape)

model = VGG19(weights='imagenet', include_top=False, input_shape=input_shape)

model = InceptionV3(weights='imagenet', include_top=False, input_shape=input_shape)

model = Xception(weights='imagenet', include_top=False, input_shape=input_shape)
```

**Code 5.16.** Available CNN models in Keras for loading as base models.

Our second model is then made of just our custom top architecture discussed in 4.3.3. We will feed it with outputs of features from the first base model. Note that for repeated experiments with the same dataset and model setting, we can save these features into files and spare a lot of valuable computation time. Code snippet 5.17 shows us an example of the use of these two models.

### 5.8.3 Mixed model

Mixed model uses the same techniques of feature transfer and is fully described in 4.4.

Code 5.18 shows how we build this model. Note that we can even reuse the saved feature files from image only model, as long as we maintain the same order of data in our dataset. `DatasetHandler` can shuffle data in deterministic manner which allows us to do so.

## 5.9 Settings structure

Throughout the whole run of experiments, we keep a shared Setting structure in form of python dictionary. Each experiment we run first loads default settings values and then it alters them depending on the experiment description file.

```
def img_model_with_feature_cooking():
  dataset = DatasetHandler.load_dataset(path_to_segments_file)
  [images, score, images_val, score_val] = dataset.get_imgs_with_labels(...)

  input_shape = dataset.get_input_shape()
  model_base = ResNet50(weights='imagenet', include_top=False, input_shape=input_shape)

  training_features = model_base.predict(images)
  validation_features = model_base.predict(images_val)

  # here it's possible to split the computation
  # save the training_features, validation_features
  # and load them repeatedly after

  model_top = build_simple_top_model( input_shape=training_features.shape[1:] )
  model_top.compile(...)

  history = model_top.fit(training_features, score, ...
      validation_data=(validation_features, score_val))
```

**Code 5.17.** Images only model using the feature cooking method. We can save the intermediate product of feature files to save on future computations.

```
from keras.models import Model
from keras.layers import Input, Dense, Dropout, Flatten, concatenate
from keras.applications.resnet50 import ResNet50

def build_full_mixed_model(input_shape_img, input_shape_osm, number_of_repeats):
  model_cnn = ResNet50(input_tensor=input_shape_img, weights='imagenet', include_top=False)
  img_features = Flatten()(model_cnn.output)

  osm_features_input = Input(shape=input_shape_osm)
  osm_features = Dense(256, activation='relu')(osm_features_input)
  osm_features = Dropout(0.5)(osm_features)

  top = concatenate([osm_features, img_features])
  for i in range(0,number_of_repeats):
    top = Dense(256, activation='relu')(top)
    top = Dropout(0.5)(top)
  top = Dense(1, activation='sigmoid')(top)

  model = Model(inputs=[model_cnn.input, osm_features_input], outputs=top)
  return model
```

**Code 5.18.** Building mixed model. Note that instead of calculating `img_features` on the fly from the selected base model, we can load the feature files shared with Image models.

Each experiment has Settings defined for the whole run and also a list in `Settings["models"]` further specifying multiple models per experiment. Each of these can be using different dataset, or it can reuse already loaded one without wasting resources.

Finally the reasoning behind experiments is to join these individual model runs and have an easy method of linking their results together while maintaining order in what are we testing. We can graph the learning progress of defined models into same plot for easier visualization.

```
DefaultSettings = {}
DefaultModel = {}
# ... specifying default values
DefaultSettings["models"] = []
DefaultSettings["models"].append(DefaultModel)
# DefaultSetting contains settings applicable for the whole experiment
# Each item in list of DefaultSettings["models"] contains specific setting
    for one individual model

# load Setup function from a file
Settings = Setup(DefaultSettings, DefaultModel)
```

**Code 5.19.** Default Settings initialization. Setup function is loaded from a customized setting file and alters the default values where it is needed.

Settings is initialized from default values, built as a python object dictionary as seen in Code 5.19. Then a custom Setup function is called which specifies differences from the default settings. This allows for descriptions of new experiments to be relatively small and to reuse already written code. This also allows for very small settings description files specifying either simple problems, while also being able to design complex experiments going into great detail. See Code 5.20 for example of a minimal experiment specification.

```
def Setup(Settings, DefaultModel):
  # minimal_model.py
  Settings["experiment_name"] = "minimalExperiment"
  # specifies the name of this experiment as well as folder name where the results will be stored
  Settings["models"][0]["dataset_name"] = "5556x_minlen30_640px"
  # specifies the folder name of dataset we want to use
  Settings["models"][0]["epochs"] = 50
  # the number of epochs we want to train our model for. Default value is 150
  return Settings
```

**Code 5.20.** Minimal settings experiment definition file specifying a simple experiment with custom dataset and number of epochs.

```
Settings[name_of_parameter] = default_value
Settings["models"][i][name_of_parameter] = default_value
```

**Figure 5.5.** Settings syntax. Python dictionary object with list of models each also pointing to a dictionary. This structure allows for high degree of customization.

Following are tables of important values of Settings syntax. Note that we write these following syntax shown in Figure 5.5. Table 5.6 contains list of parameters which are applied over the whole experiment run. All other parameters are model specific and are listed in Tables 5.7 and 5.8.

| parameter name | default value | description |
|---|---|---|
| experiment_name | 'basic' | Name of the experiment as well as name of the folder used to store the results in. |
| graph_histories | ['all', 'together'] | Accepts values of 'all', which produces one graph for each model, 'together', which draws graphs of all models into one shared graph. Also accepts lists of numbers, which specify which models we would like to have plotted together. For example [0,2] would plot histories of the first and third model together. |
| interrupt | False | Internal flag used to interrupt the whole experiment in case of error. |
| filename_features_train, filename_features_test | '' | Internal string flags used to share file paths amongst different parts of code in ModelHandler. |

**Table 5.6.** List of parameters shared throughout the whole experiment.

| parameter name | default value | description |
|---|---|---|
| **Dataset specific** | | |
| dataset_pointer | -1 | If left at value '-1' we instantiate a new dataset for this model. Otherwise we use the dataset of the indicated different model. For example if the first model has -1, then the second model can have dataset_pointer set to 0 to reuse the same dataset. |
| dataset_name | '1200x_markable _299x299' | States the name of folder we want to use for loading the dataset from (this folder has to contain SegmentsData.dump and a folder 'images' with Google Street View images). Downloader can prepare these folders for us. |
| dump_file_override | '' | Can be used to indicate usage of nonstandard segments file. For example we can keep multiple versions of OSM marked data in different files and access them via this setting (for example loading file 'Segments-Data_marked_R100.dump'). |
| pixels | 299 | Pixel dimension indicator of the loaded images (image files have dimension of pixels*pixels*3). |
| number_of_images | None | Indicates if we want to use only a subset of the dataset. When left at None, whole dataset will be used, otherwise a uniform sampling will be used to give us indicated number of images. |
| seed | 13 | Seed for maintaining deterministic nature of certain random sampling operations (such as the initial reordering of loaded dataset). This value is important as it guarantees our ability to reuse precomputed feature files. |
| validation_split | 0.25 | Under which fraction we split data in simple (those not guided by k-fold cross-validation scheme) experiments. Ranges from 0 to 1. With 0.25 one quarter of data will designated as validation set and the remainder of three-quarters will be the training set. |
| shuffle_dataset | True | Indicator that we want to shuffle our dataset in deterministic manner (using the seed value). |
| shuffle_dataset _method | 'default-same-segment' | Shuffling method which maintains the order of images from the same segment to be kept together, which is important not to bring in dualities into our data. |

**Table 5.7.** List of parameters specific to individual models.

| parameter name | default value | description |
|---|---|---|
| **Dataset specific** | | |
| edit_osm_vec | ‘’ | Additional transformation of the OSM data, accepts ’booleans’, which turns all quantitative entries into 0 or 1. Setting ’low-mid-high’ turns all entries into three categorical system of low, mid and high depending on percentiles of data distribution in dataset. These three categories are encoded as one-hot vector producing variants: 001, 010 or 100. Note that this effectively triples the length OSM vector. |
| **Model specific** | | |
| unique_id | | Unique name for this model, will be used for plotting graphs and later for identification which plotted history belonged to which model. |
| model_type | ’simple_cnn_with_top’ | Specifies which model type will we use. Accepted values are: ’simple_cnn_with_top’, ’img_osm_mix’, ’osm_only’. |
| cnn_model | ’resnet50’ | Which basic CNN model will we use in case of image_only and mixed model. Allowed values are: ’vgg16’, ’vgg16’, ’resnet50’, ’inception_v3’, ’xception’. |
| cooking_method | ’generators’ | Internally used flag to indicating which method will we use to generate feature files. ’generators’ tends to be less memory demanding, but consumes longer time when compared with the other allowed value ’direct’. |
| top_repeat_FC_block | 3 | Specifies depth of classifier in image_only and osm_only models. |
| osm_manual_width | 256 | Manually setting the width of osm_only model. |
| save_visualization | True | Whether we want to plot graphs for this experiment. |
| **Training specific** | | |
| epochs | 150 | Indication of how many epochs we want to spend on training of this model. |
| optimizer | ’rmsprop’ | Choice for the Keras optimizer. Suggested values are ’rmsprop’ or ’adam’, however also accepts the object of Keras optimizer such as customizable optimizers.SGD(lr=0.01, decay=1e-6, momentum=0.9) |
| loss_func | ’mean_squared_error’ | Metric used as a loss function during training of this model. |
| metrics | [’mean_absolute_error’] | List of other metrics which we want to also track into history. |

**Table 5.8.** List of parameters specific to individual models (continued).

## 5.10 Experiment running

`ExperimentRunner` is special scheme which uses other modules mentioned in this chapter. It starts with loading a custom Settings file and then follows the same structure for various experiments eventually evaluating the tested model and dataset combination.

```
def experiment_runner(settings_file=None, job_id):
    Settings = SettingsDefaults.load_settings_from_file(settings_file,
        job_id, verbose=False)

    # preparation
    datasets = ModelOI.load_dataset(Settings)
    Settings = ModelOI.prepare_folders(Settings, datasets, verbose=True)
    models = ModelGenerator.get_cnn_models(Settings)

    # building
    ModelTester.cook_features(models, datasets, Settings)
    models = ModelGenerator.get_top_models(models, datasets, Settings)

    ModelOI.save_visualizations(models, Settings)

    # training
    histories = ModelTester.train_models(models, datasets, Settings)

    # save and report results
    ModelOI.save_histories(histories, Settings) # to .npy files
    ModelOI.graph_histories(histories, Settings) # to png images

    ModelOI.save_report(Settings)
    ModelOI.save_models(models, Settings)

    ModelOI.send_mail_with_graph(Settings)
    ModelOI.save_metacentrum_report(Settings)
```

**Code 5.21.** ExperimentRunner pseudocode outlining the run of whole experiment, we are making heavy use of the units explained in this chapter.

The structure of ExperimentRunner is outlined in Code 5.21

## 5.11 Training

Basic model functionality is supported by Keras with commands in Code 5.22. We specify settings of the training procedure by choosing parameters of optimizer, loss function and additional metrics. We chose optimizers `adam` and `rmsprop`, which work well with their default settings, as is discussed in 4.6.

As a loss function we have selected “mean squared error” with formula in Figure 4.17. Keras allows us to specify additional metrics, which can be used to simply track progress, but don’t actually influence the training process.

Finally we are given a history stored in dictionary of arrays with tracked values (both of the loss function and measured metrics). We make use of these histories to plot the progress of training into graphs.

## 5.12 Testing

For more relevant results we use the advanced method of k-fold cross-validation. In code we actually use the same syntax as the one of 5.11, however with data generation scheme described in 4.7.

Note that we can use the results of graphed histories even without using the k-fold cross-validation, however this gives us more in depth view.

```
def train_model(model, dataset, model_setting):
  [...] = dataset.get_data(...)
  model.compile(optimizer, loss, metrics)
  history = model.fit(...)

  return history
```

**Code 5.22.** Model training syntax. Model is trained on a specific dataset according to the settings. History is saved for later graph plotting.

## 5.13 Reporting and ModelOI

ModelOI is a programmatic unit responsible for inputs and outputs of the larger experiment runs. We use it to communicate with DatasetHandler while loading datasets at the beginning of experiments. We also use methods of ModelOI for reporting after the training has ended.

Histories produced during training in ModelTester are processed and plotted into graphs via the Matplotlib library.

For convenience we also chose to send interesting results via email after the often lengthy computations have finished.

## 5.14 Metacentrum project and scripting

Thanks to the Metacentrum project [46] we had access to machines with high enough computational power needed to teach even complicated models.

Metacentrum is a project supporting research groups including the academic research groups at our university. They provide a server cluster capable of running our experiments on wide array of machines with powerful hardware. We control the individual runs of experiments by special tasking and job queuing system, PBS Professional. We also make use of its storage units with image datasets containing many items and large feature files.

Scripts controlling the job queuing on Metacentrum servers are using the PBS Professional job scheduler with couple of commands described in Code 5.23.

Note that further customization of commands is available, however these commands were all we needed for our experiments.

| | |
|---|---|
| qsub task.sh | basic specification to run bash program in task.sh |
| -l walltime=<time> | reserved time (for example walltime=23:30:00) |
| -l mem=<memory> | memory requirements (for example mem=32gb) |
| -l ncpus=<n> | number of CPUs used (for example ncpus=4) |
| qstat -u <user> | command to get list of tasks run by a specific user |
| qstat <task_id> | get information about task specified by unique task_id |
| qdel <task_id> | delete planned job |
| qsig -s SIGINT <task_id> | interruption of running task by the SIGINT signal |

**Code 5.23.** Metacentrum planner scripts

Contents of task.sh written in bash are in Code 5.24. We first load Anaconda environment and step into the correct directory. Then we run python code which is supplied

```
#!/bin/bash

export PATH="/<home_directory>/anaconda2/bin:$PATH"
cd /<home_directory>/MGR-Project-Code/

python run_on_server.py Settings/sample_settings_description_file.py $PBS_JOBID
```

**Code 5.24.** Bash code to run python file with custom targeted Settings file. Note that `$PBS_JOBID` is an unique job id given by the tasking and scheduling software.

with path to Settings file and an unique job id. Settings files are described in 5.9 and allow us to customize runs of experiments relatively freely.

# Chapter 6
# Results

This chapter is divided into several sections. First we will look at the efficiency of various strategies used when generating the dataset in 6.2 and then when building models in 6.3.

## 6.1 How to read graphs in this section

We have implemented a k-fold cross-validation discussed in 4.7 and used it to evaluate performance of the tested model and dataset combination. We present a higher level view of the experiments by plotting the evolution of error over training epochs. Note that we can follow both the training and validation error to see how much is our model overfitting.

We also provide a more detailed view into the situation after the training was finished. We look at the box plot graphs of errors in the last epoch of training.

Note that in some cases we may want to hold the model with the best achieved validation error, instead of the one from last iteration. This applies in the case when with further training iterations the model overfits. For this view we may show box plot graphs of the epoch where each model fared the best.

## 6.2 Strategies employed in dataset generation

In this section we focus on the results of various attempts to alter the dataset generation scheme.

As we described in 3.2 we have multiple ways to construct our dataset. We have tried varying the size of downloaded images in 6.2.1.

We also explore two types of increasing the images available for our models. In 6.2.2 we try splitting long edges to smaller segments, while maintaining a limit of how small can resulting segment be. The smaller the minimal segment length, the higher the number of segments and consequentially the number of images. See 3.2.3 and Code 5.3 for details.

In 6.2.3 we try an alternative approach to dataset augmentation from data already obtained by applying additional transformations such as flipping the image, shifting or scaling. In this way we create new images from an already existing set.

### 6.2.1 Dimensions of downloaded images

We chose to try two settings to illustrate the effect of our choice in the matter of image dimensions. Following the original article of [12], we chose our first image size to be 299x299 pixels. Our second dataset is made of images sized 640x640 pixels, which was the limitation of Google Street View as the most detailed image resolution available freely via their API.

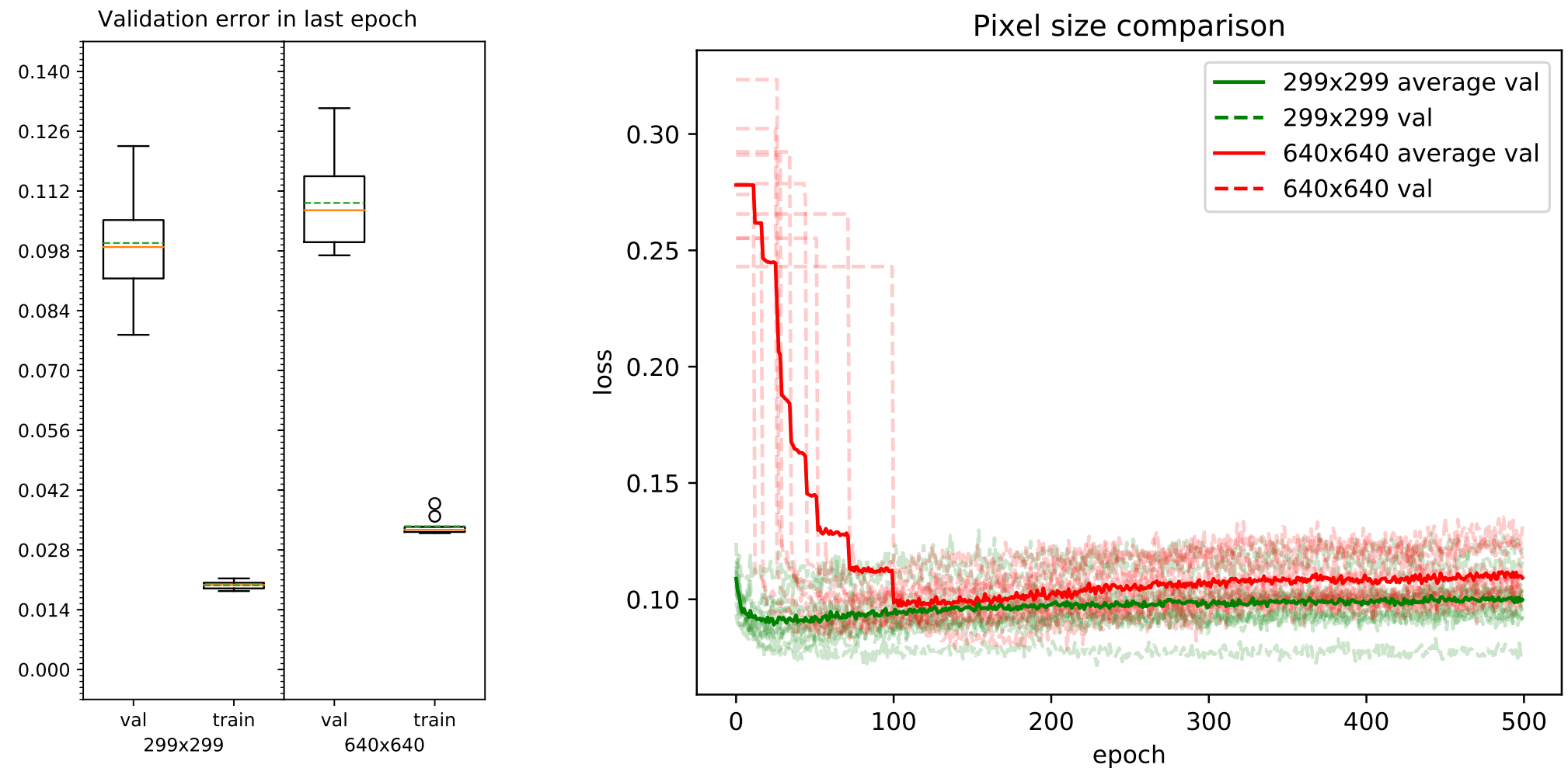

**Figure 6.1.** Model: *Mixed*, Datasets: *299x299*, *640x640*. Results of experiment with pixel size of input images. Here we see the graph of evolution of loss function over epochs. K-fold cross-validation scheme was used with k=10. We can see error on validation dataset of all the 10 runs with average results highlighted.

Note that the 640x640 dataset is bigger in its size and it also takes more time for the models to be trained on it. Also the consequent feature files which we store in between experiments to allow their reuse are bigger.

We have used datasets internally marked as `5556x_mark_res_299x299` for 299x299 dataset and `5556x_markable_640x640` for the 640x640 dataset. Their only difference is in the image pixel dimensions.

As for the choice of models we don't include the OSM model as it is not affected by image pixel size. Figure 6.1 shows the results when choosing Mixed model (see 4.4). Figure 6.2 shows the results when using Image model (see 4.3).

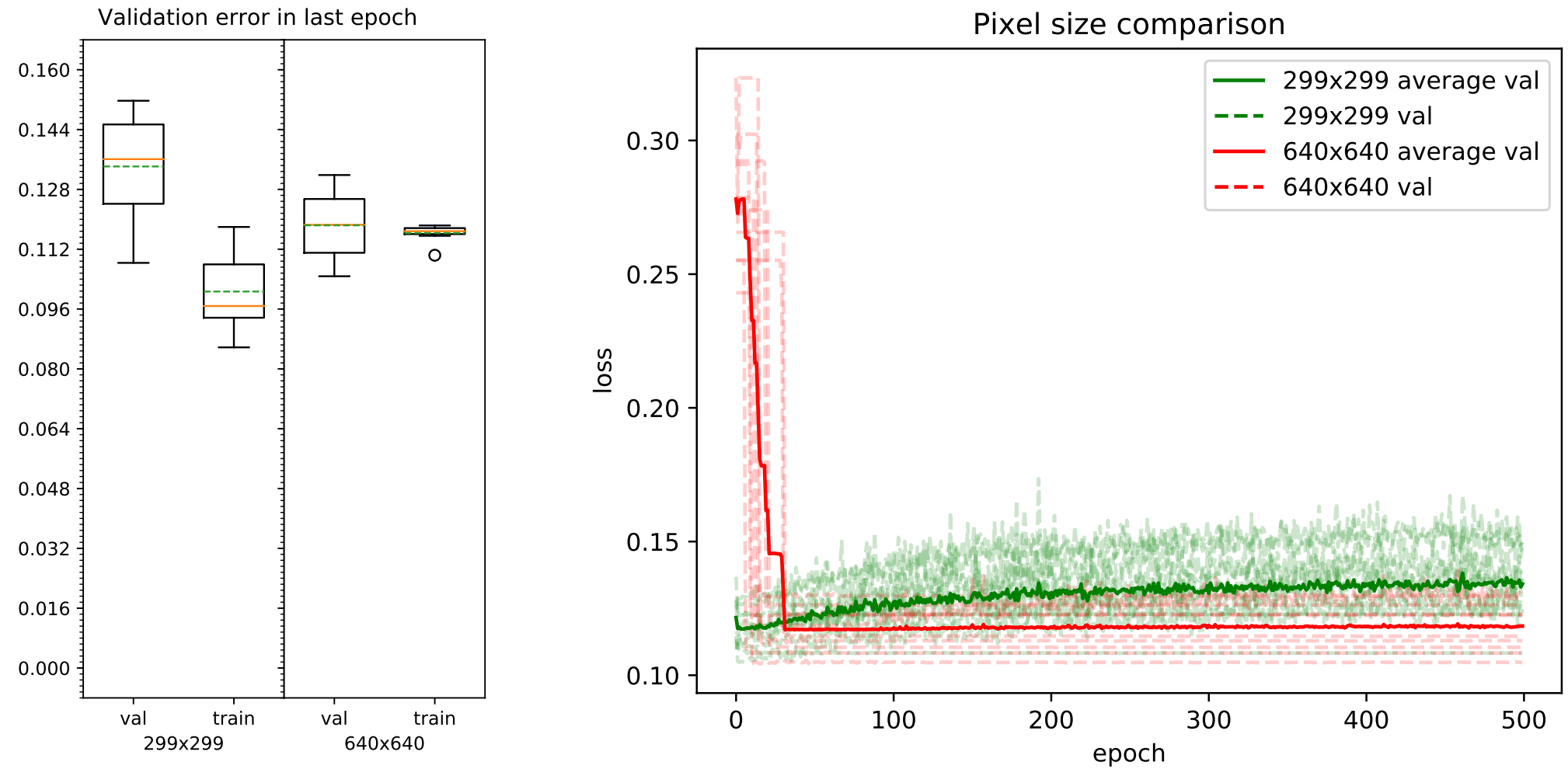

**Figure 6.2.** Model: *Image*, Datasets: *299x299*, *640x640*. Results of experiment with pixel size of input images. Here we see the graph of evolution of loss function over epochs. K-fold cross-validation scheme was used with k=10. We can see error on validation dataset of all the 10 runs with average results highlighted.

We can see both the validation error as well as training error evolution over time of epochs as well as box plots with more detailed view of the last epoch.

In order to compare the performance of both the two used datasets and models, we use the box plots in Figure 6.3, which look at the best validation error score of the four models.

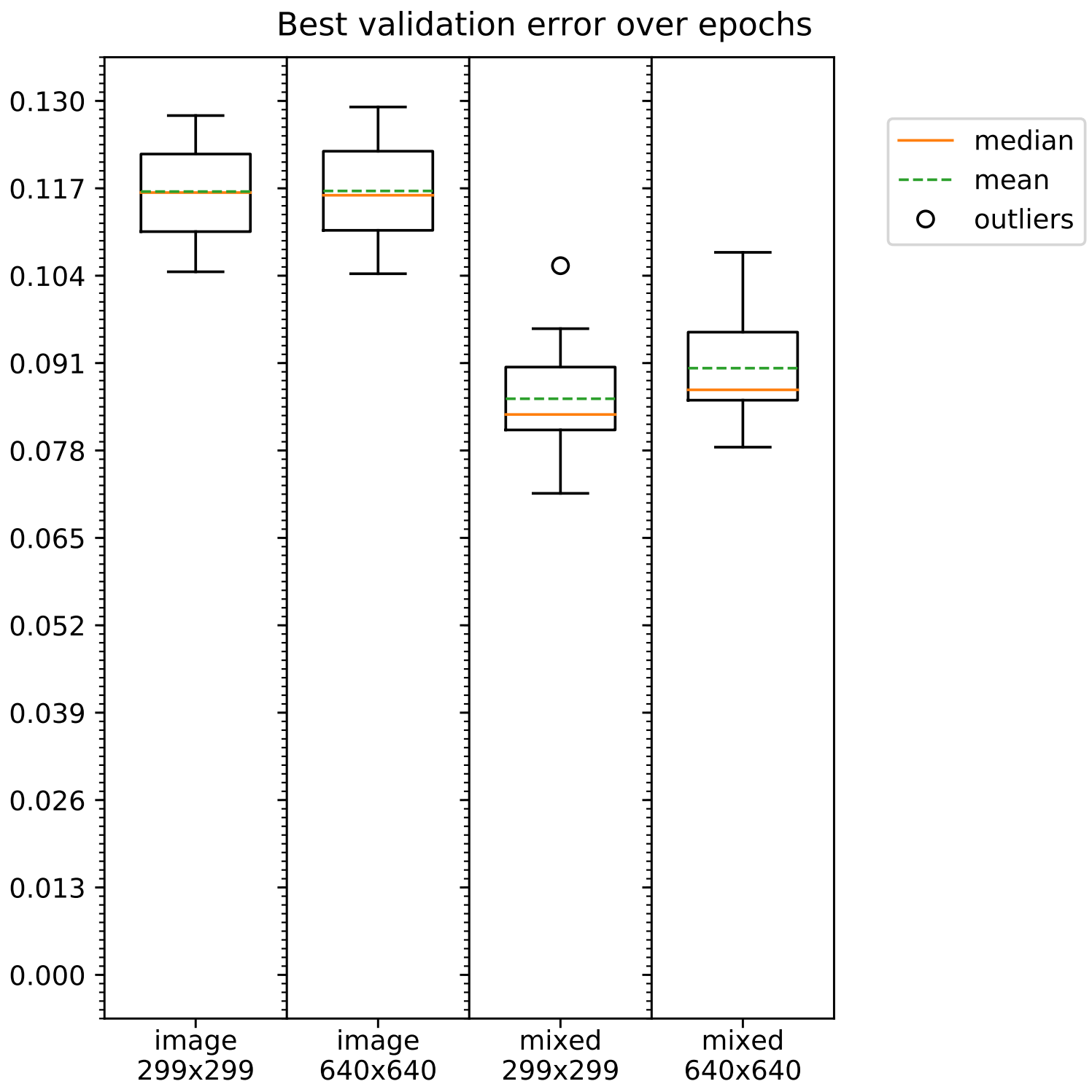

**Figure 6.3.** Models: *Image*, *Mixed*, Datasets: *299x299*, *640x640*. Comparison of four model and dataset combinations used in the pixel size experiment. We look at the best achieved validation error throughout their evolution over epochs.

As was already mentioned on Figure 4.10 the input image dimension will influence the dimension of intermediate features produced by the base CNN model. Consequent structure of our models has to deal with this variability in size.

As we can see in the case when we used Image only model on Figure 6.2, the use of bigger images is advantageous - in the last epoch the statistics of validation error speak in favor of dataset 640x640.

Somewhat surprising is the result over the Mixed model, where as can be seen on Figure 6.1 the validation error of 640x640 dataset is actually worse than that of 299x299 in terms of median, mean and also the spread of lower and upper percentiles. Here we should consider the situation, where the 299x299 dataset produces features of size (1, 1, 2048), whereas the 640x640 dataset produces size (2, 2, 2048).

Regardless of the chosen dataset, the performance of Mixed model is better than that of Image only model. However the 299x299 dataset seems to gain more from this transition than the 640x640 one. We end up with the Mixed model on 299x299 dataset beating other combinations on its last iteration. Figure 6.3 confirms this finding also for the best achieved validation error along all epochs for all these model and dataset combinations.

| Dataset | # of images |
|---|---|
| Original 5556x_markable_640x640 | 8376 |
| min length 10m 5556x_minlen10_640px | 24078 |
| min length 20m 5556x_minlen20_640px | 18534 |
| min length 30m 5556x_minlen30_640px | 13458 |

**Table 6.4.** Dataset overview for the experiment of splitting long edges into smaller segments. Note that the number of images corresponds with the size of inputs for models.

### 6.2.2 Splitting long edges

See Table 6.4 with a list of tested datasets. When we choose to split long edges, we generate more smaller segments and from each of these segments even more images. Note that we count only valid segments, where there is imagery data available and where we had a label in the initial dataset.

We chose to run the experiment with the original dataset 5556x_markable_640x640 as comparison of how models fared without long edge splitting.

For this experiment it again makes sense to test on the Image and Mixed models as those are the ones affected by imagery data input.

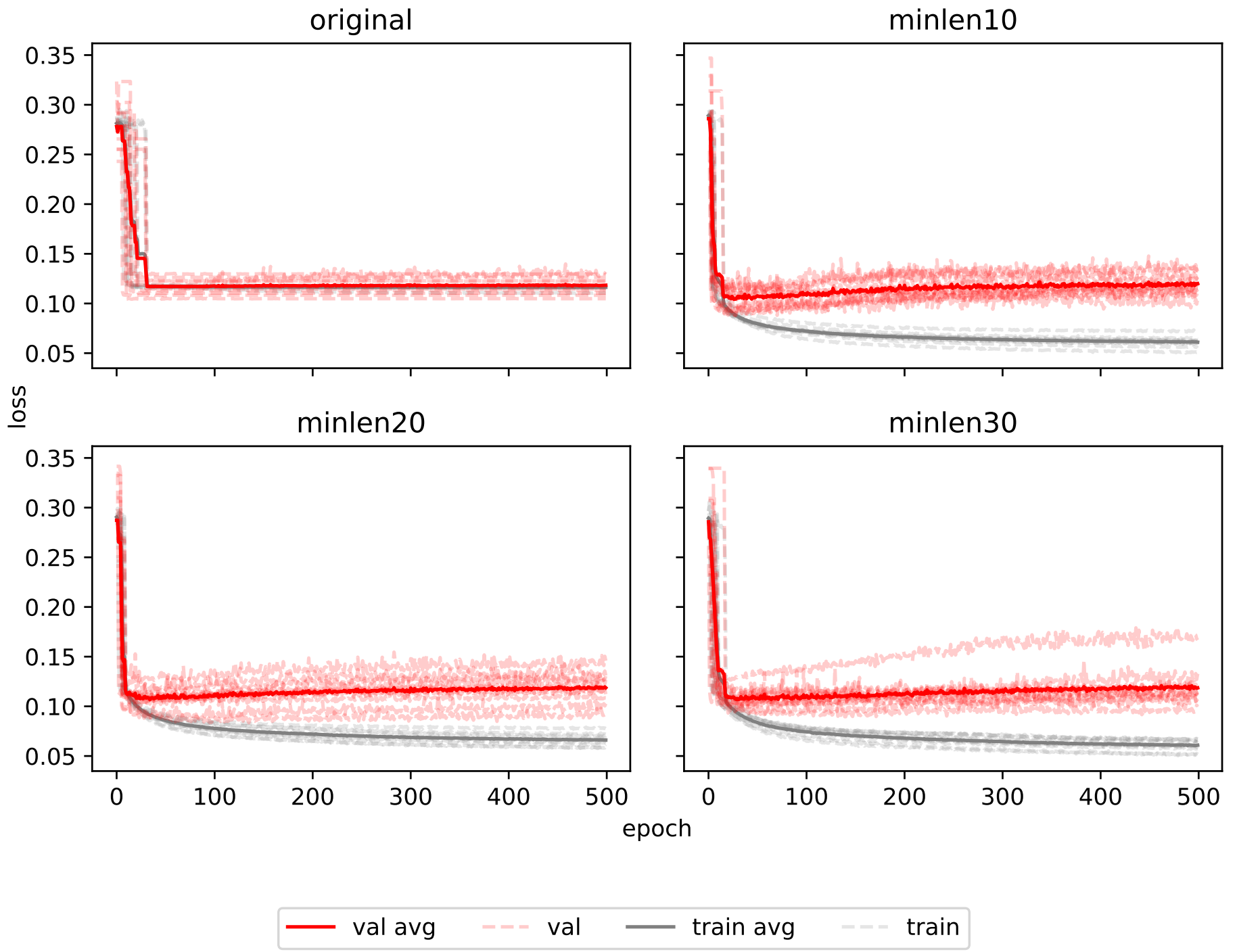

**Figure 6.5.** Model: *Image*. Overall illustration of the training process of three datasets which were enhanced by long edge splitting and an original dataset for reference.

Starting with Image model, we chose to present the detailed information about the progress of training in Figure 6.5, which shows all four dataset variants alongside with its validation and training error. In Figure 6.6 we see the average validation errors plotted together in order to better compare their performances. Lastly in 6.7 we can see the comparison of all four datasets with their best score over all epochs.

Note that we also tested the Mixed model in Appendix with Figures B.1 and B.2.

When looking at Figure 6.7 which portrays the comparison of Image model over the four datasets, we can see quite clearly that the expanded datasets perform better with their median and mean values always improved over the original dataset. In the case of "minlen30" dataset we can see vast improvement with the exception of several outliers.

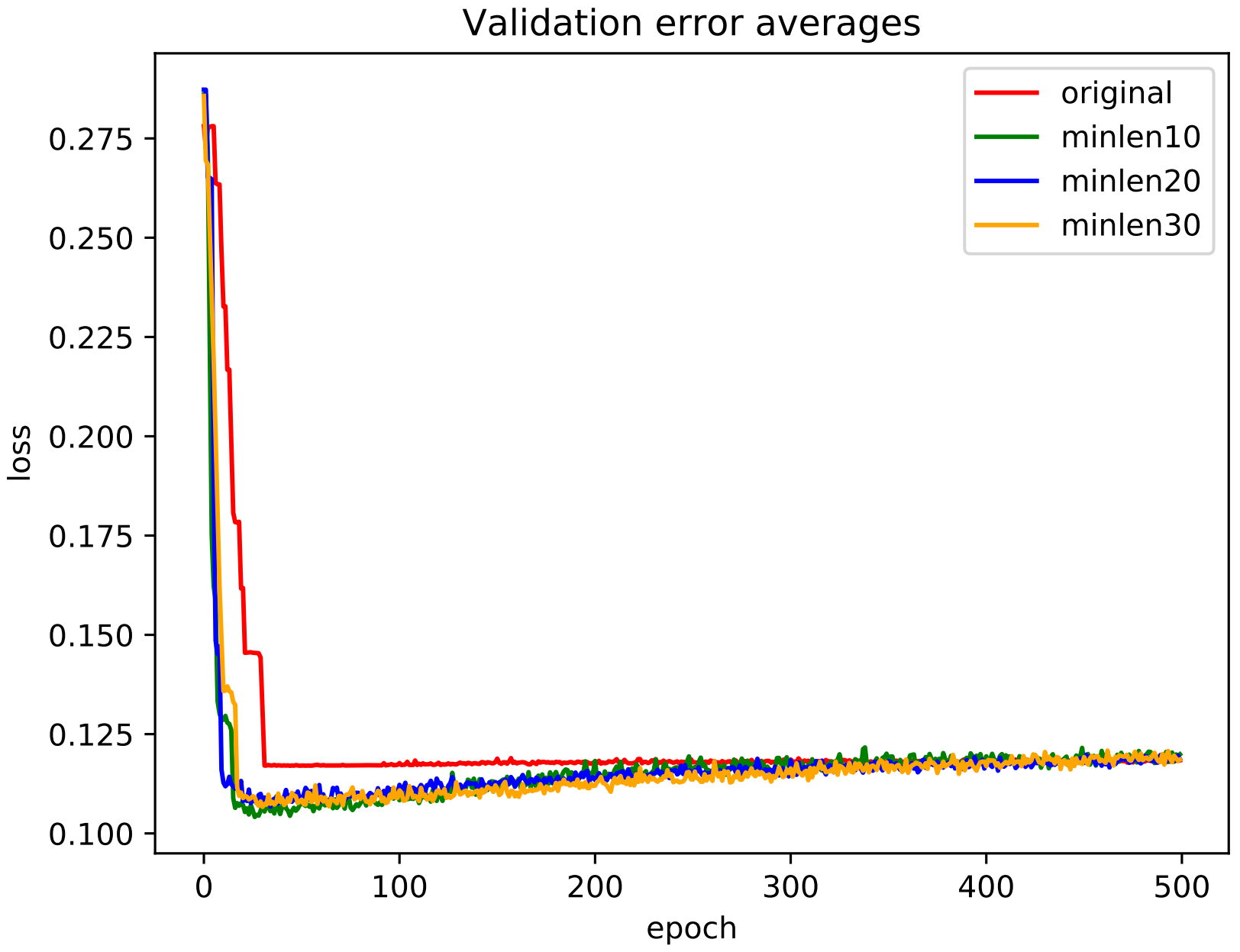

**Figure 6.6.** Model: *Image*, Datasets: *original*, *minlen10*, *minlen20*, *minlen30*. Long edge splitting experiment. For better clarity we present only the averages of validation errors over the 10 runs of k-fold cross-validation with k=10.

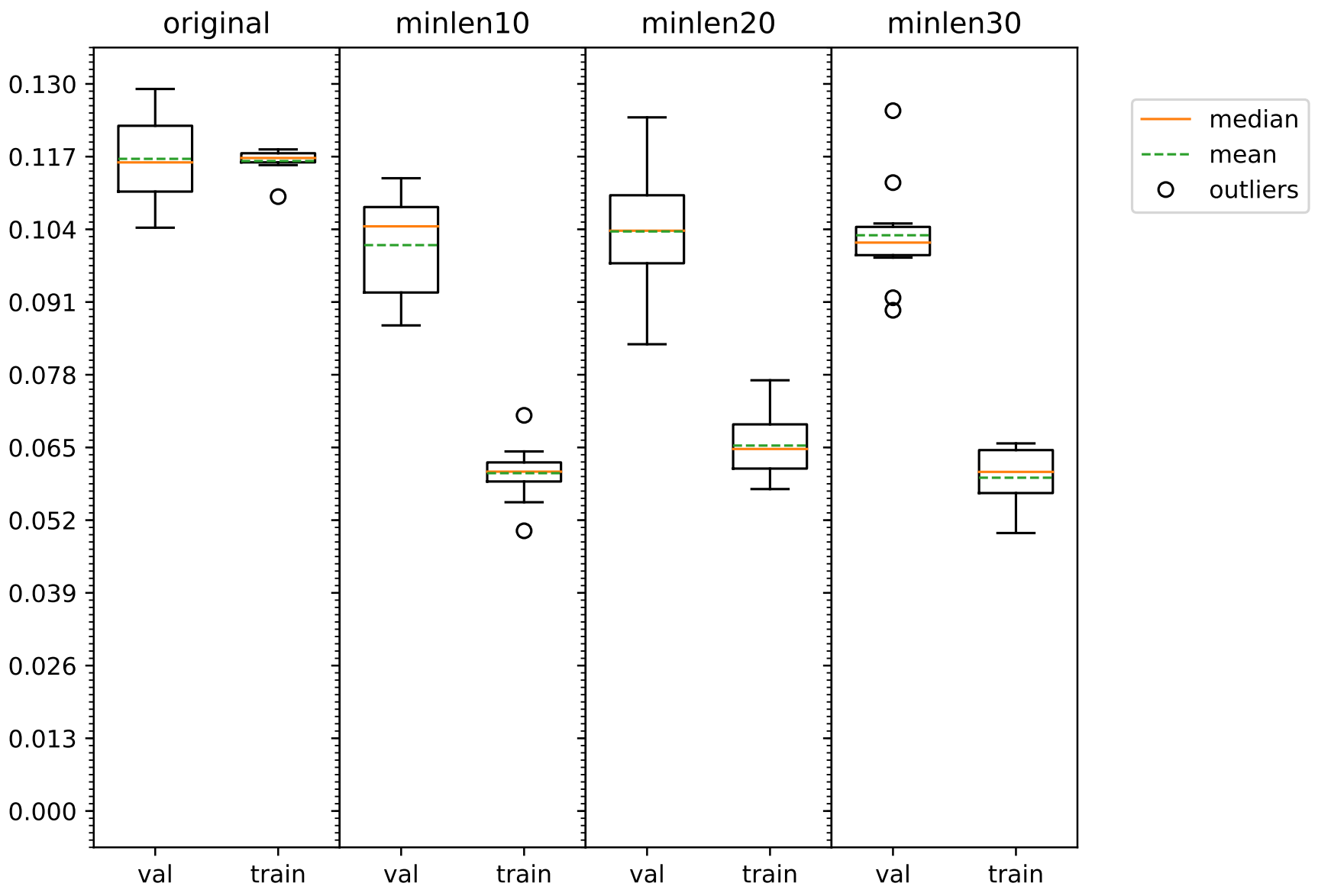

**Figure 6.7.** Model: *Image*, Datasets: *original*, *minlen10*, *minlen20*, *minlen30*. Comparison of the results of splitting long edges into smaller segments. The smaller the minimal edge length was, the more images were generated. Dataset `5556x_markable_640x640` under the label original was used for comparison. Results are from the epoch with best validation error.

However note that for the Image model the validation error averages seem to converge to the same value in their last epoch as can be seen on Figure 6.6. It is also worth noting, that in the initial 100 iterations the Expanded datasets performed better and only overfit towards the same mean average value with their 400th to 500th iteration. The best validation errors in Figure 6.7 clearly come from these initial iterations.

For several further experiments we chose dataset “5556x_minlen30_640px” to accompany the original “5556x_markable_640x640”.

Situation in Figure B.1 is different and the results show that the Mixed model is not as easily helped by the long edges splitting. Look at Figure B.2 shows the best result over all epochs and reveals that the altered datasets of “minlen10”, “minlen20” and “minlen30” didn’t bring much improvement over the original one. The spread of results in the non-original datasets is wider and seems to reach better score values, however both the median and mean of all expanded datasets has slightly worsened.

### 6.2.3 Dataset augmentation

| Dataset | # of images |
|---|---|
| Original `5556x_markable_640x640` | 8376 |
| Expanded | 25128 |
| Aggresively expanded | 25128 |
| Original `5556x_minlen30_640px` | 13458 |
| Expanded | 40374 |
| Aggresively expanded | 40374 |

**Table 6.8.** List of datasets used for the experiment of data augmentation with the number of valid images they contain. Note that the “Expanded” and “Aggresively expanded” dataset followed different generation scheme, even though their number of images is the same (3 times their original dataset).

As was explained in 5.9 we chose two schemes for data augmentation which gives us three datasets to work with as listed by Table 6.8.

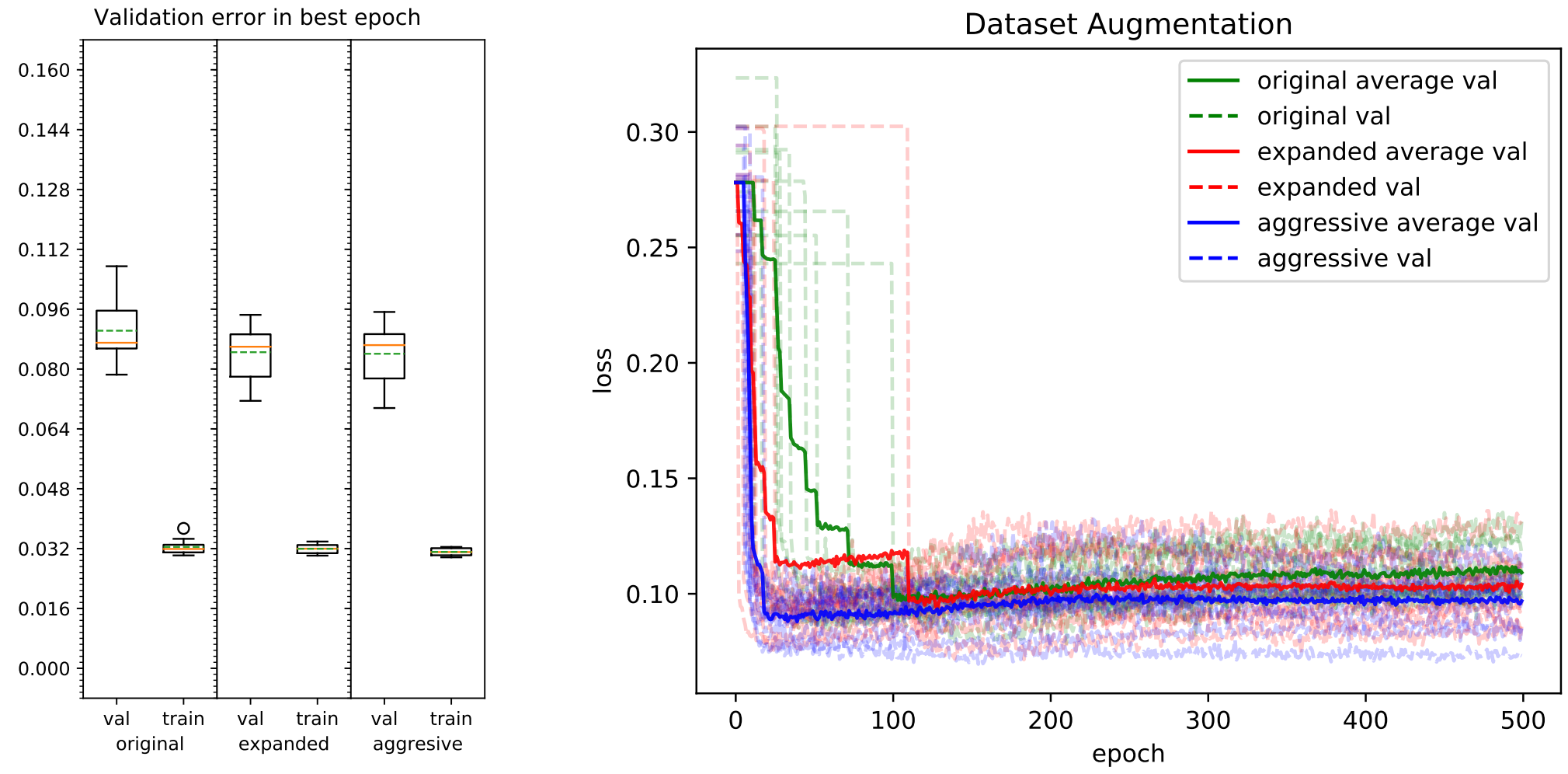

**Figure 6.9.** Model: *Mixed*, Dataset: *5556x_markable_640x640*. Comparison of the two augmented datasets with the original referential one. Expanded dataset used the transformations of image flipping and shifting. Aggresively expanded dataset used additional transformations of shear, scale and rotation.

Figure 6.9 shows the evolution of validation error over the epochs, as well as the validation and training error distribution from k-fold cross-validation in its best epoch. On Figure 6.10 we have the same experiment using `5556x_minlen30_640px` as the original dataset.

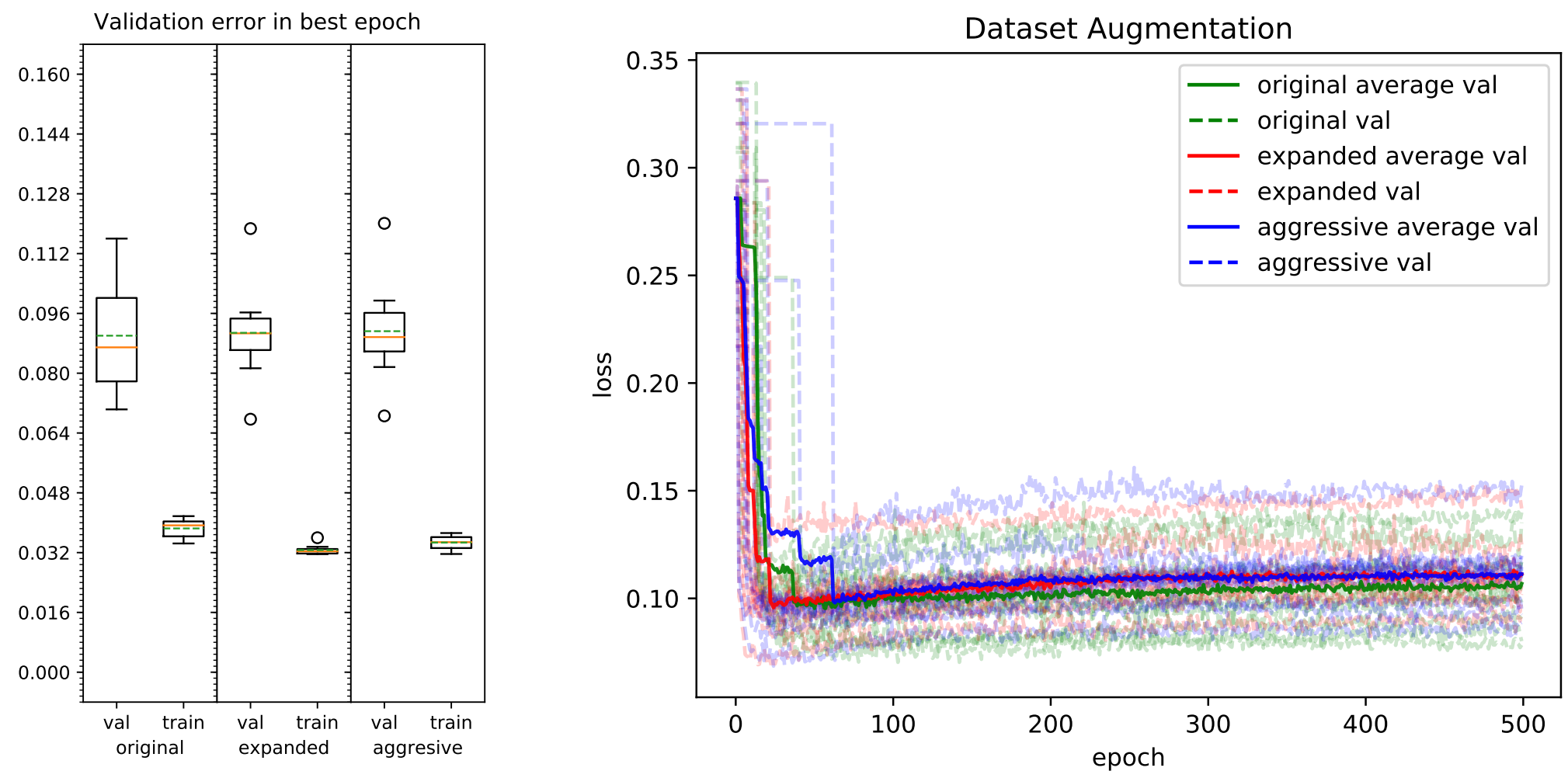

**Figure 6.10.** Model: *Mixed*, Dataset: *5556x_minlen30_640px*. Comparison of the two augmented datasets with the original referential one.

Let us first focus on the effect of data augmentation over the basic `5556x_markable_640x640` dataset. When comparing the three used datasets in left plot of Figure 6.9, we see better results in terms of both median and mean an the distribution of errors - both expanded datasets improve the original one. When consulting the overall view on the right side of Figure 6.9 we see, that the average validation error is improved significantly for the expanded datasets over the whole training period.

Situation with dataset `5556x_minlen30_640px` taken as the base for further augmentation hasn't brought the same amount of performance boost, as can be see in Figure 6.10. In fact both the median and mean of validation error has slightly worsened in the cases of both expanded datasets. However the spread of results is lower in the expanded datasets. Overall view on the right side of Figure 6.10 underscores that in this case the data augmentation didn't significantly help.

It is interesting to compare the result of data augmentation on the original dataset, versus on the dataset with more data samples in `5556x_minlen30_640px`. When looking at Table 6.8, we can see the exact numbers of images. It seems that the dual data enhancement, first with long edge splitting followed by augmentation by random image transformations didn't bring in the expected error reduction. The byproduct was in fact vastly increased time needed for training and evaluation of models with more than 4 times the images than the initial dataset `5556x_markable_640x640`.

## 6.3 Strategies employed in model architecture

We already could see changing behavior of different model types interacting with differently generated datasets in 6.2, but here we will focus directly on the changes of model architectures and their influence on performance.

Note that we will mostly use the datasets which showed promising results in 6.2 ignoring those which lead to decrease of performance.

When using the Mixed and Image models, we are using an already existing base CNN model for feature transfer (see 4.3 and Figure 4.9), so in the our first experiment in 6.3.1 we will explore the possible types of used CNN base models.

In 6.3.2 we look into the comparison of performances of the three designed models – Image only, OSM only and Mixed model.

In 6.3.3 we will look at a OSM model specific setting of depth and width of the CNN design.

### 6.3.1 Different CNN base model for feature transfer

As was mentioned in Code 5.16, Keras allows to use multiple pretrained models alongside with weights initiated from the ImageNet dataset.

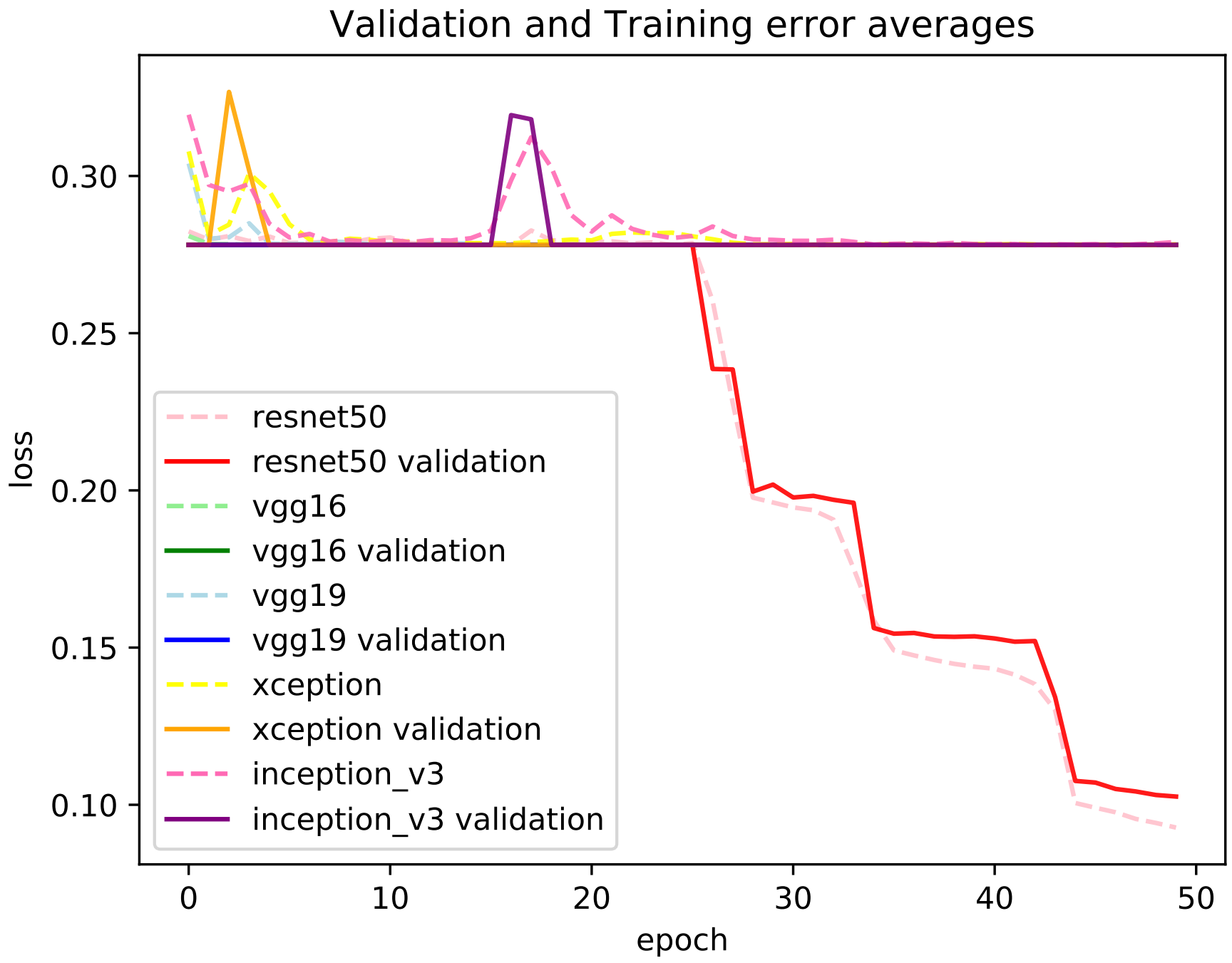

**Figure 6.11.** Model: *Mixed with variable base*, Dataset: *5556x_markable_640x640*. Illustration of five models in one graph. For better clarity we chose to show only the averages of validation and training errors. These Mixed models used different CNNs as their base.

Note that these models plotted in 6.11 have different number of parameters in their structure and therefore the whole evaluation of experiment can take different amount of time and exhibit different memory requirements.

We have also made the same comparison between different base CNN models on dataset with 299x299 pixels in Figure B.3.

Plot of Figure 6.11 shows us the first 50 epochs into training of Mixed models built with different base CNNs from the list which Keras allows us to use. All the models except for `resnet50` have great difficulties to learn on the dataset. Table 6.12 provides us with possible explanation. With chosen dataset of images sized 640x640 pixels, the dimensionality of features we save at the output of used base CNN varies dramatically between model `resnet50` with (2, 2, 2048) and other models with values up to (20, 20, 2048). Subsequently the training process of top model with an unchanged architecture is greatly hindered with too large features. We would need to alter the architecture of top models appropriately using pooling layers to subsample the dimension, or make use of dataset with smaller input images.

| Mixed model with ... | # of trainable parameters of base CNN | dimensionality of features |
|---|---|---|
| resnet50 | 23,534,592 | ($n$, 2, 2, 2048) |
| vgg16 | 21,768,352 | ($n$, 18, 18, 2048) |
| vgg19 | 20,024,384 | ($n$, 20, 20, 512) |
| inception v3 | 14,714,688 | ($n$, 20, 20, 512) |
| xception | 20,806,952 | ($n$, 20, 20, 2048) |

**Table 6.12.** List of models available in Keras framework with the number of parameters the Mixed model using it as its base CNN model will have. Note that $n$ in the dimension of features signifies the number of images. Dataset used for this measurement contained images of 640x640 pixels.

After all these models were initially used with images of pixels size in range of 224x224 to 299x299. However Figure B.3 confirms that we can see similar behavior even on smaller dataset `5556x_mark_res_299x299` which contains the same amount of images, just resized to pixel size of 299x299.

### 6.3.2 Model competition - Image vs. OSM vs. Mixed

We have proposed three major CNN architecture schemes to build models - OSM only described in 4.2 and built with Keras by Code 5.15, Image only described in 4.3 and in Code 5.17 and finally the Mixed model described in 4.4 and built by Code 5.18.

We chose one dataset to compare these three models on. We use the default setting for building these models.

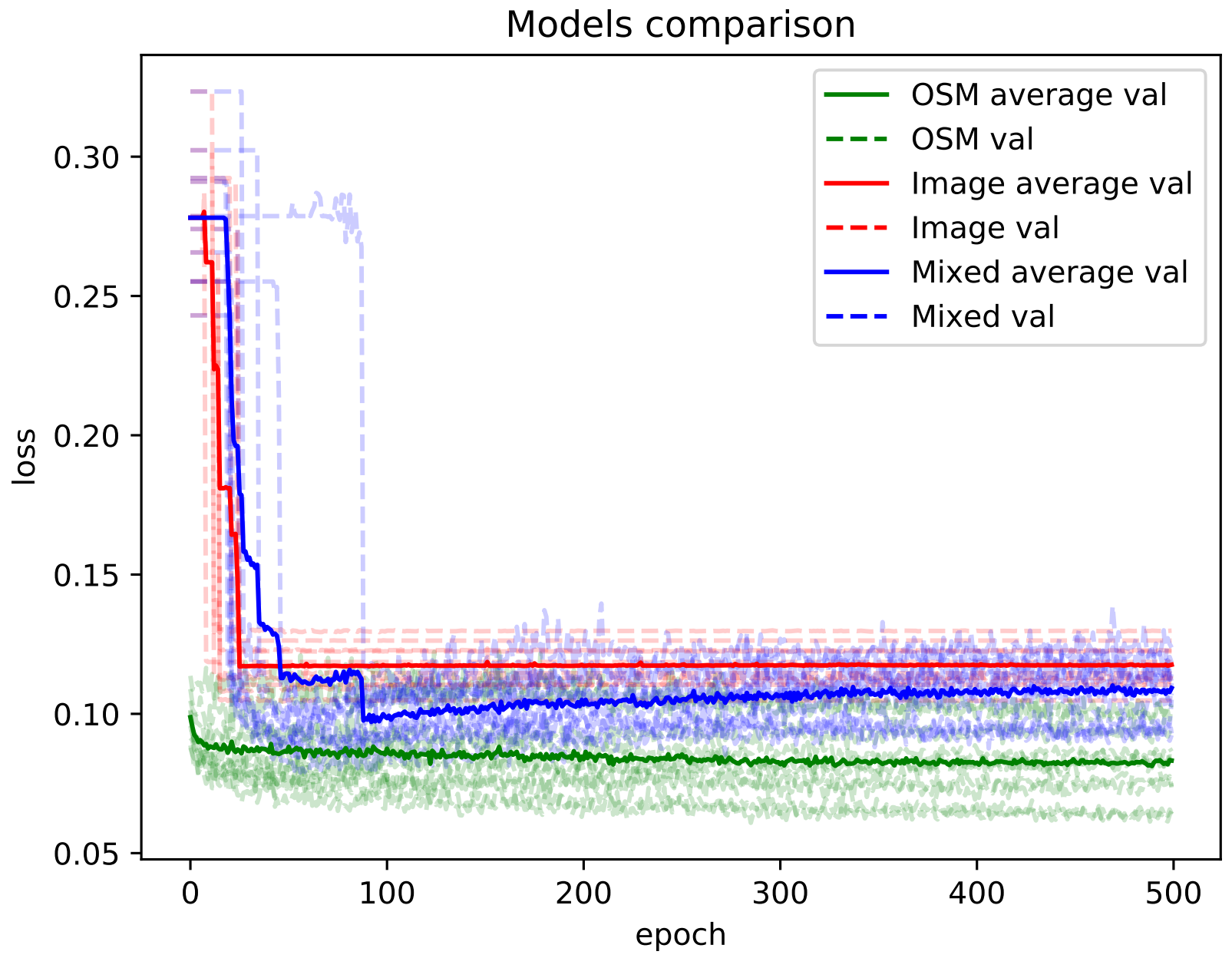

**Figure 6.13.** Models: *OSM*, *Image*, *Mixed*, Dataset: *5556x_markable_640x640*. Model competition, evolution over epochs.

See Figure 6.13 for high level view of evolution of each models performance over time of epochs. For detailed view of models state in the final epoch see 6.14. To be fair for evaluating the models, we also look at Figure 6.15 comparing the best performance of each model when just looking at its validation error. Note that the exact epoch when the model performed the best can be different for these three models.

The results of `5556x_minlen30_640px` dataset are shown in Appendix B.1.3.

We will start by looking at the models performance on the basic `5556x_markable_640x640` dataset on Figure 6.13. We can see that on the basic dataset the OSM model outperforms all the other models by quite large margin. In detail we can explore the performance in Figure 6.14 in last iteration of the training process, or to be more fair in Figure 6.15 which looks at the best result over the whole training. The OSM model has better median and mean of validation error as well as lower and upper quartile spread than the other two models.

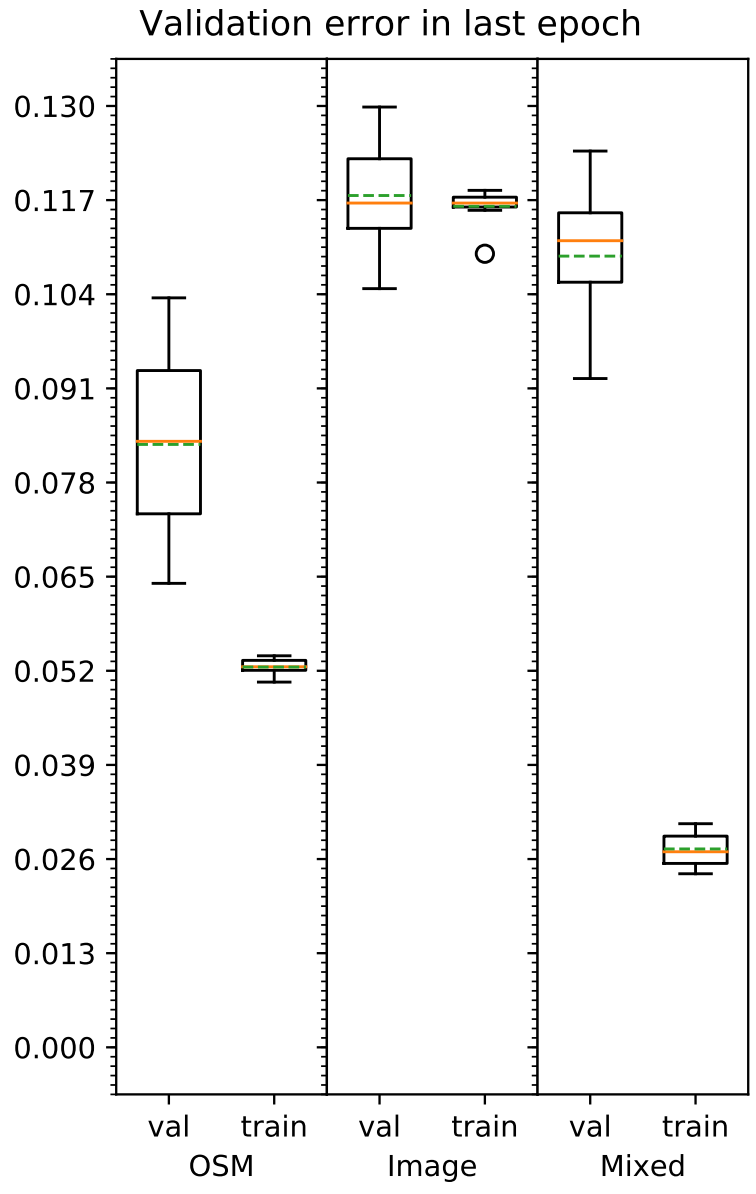

**Figure 6.14.** Model competition, last epoch. Dataset: *5556x_markable_640x640*.

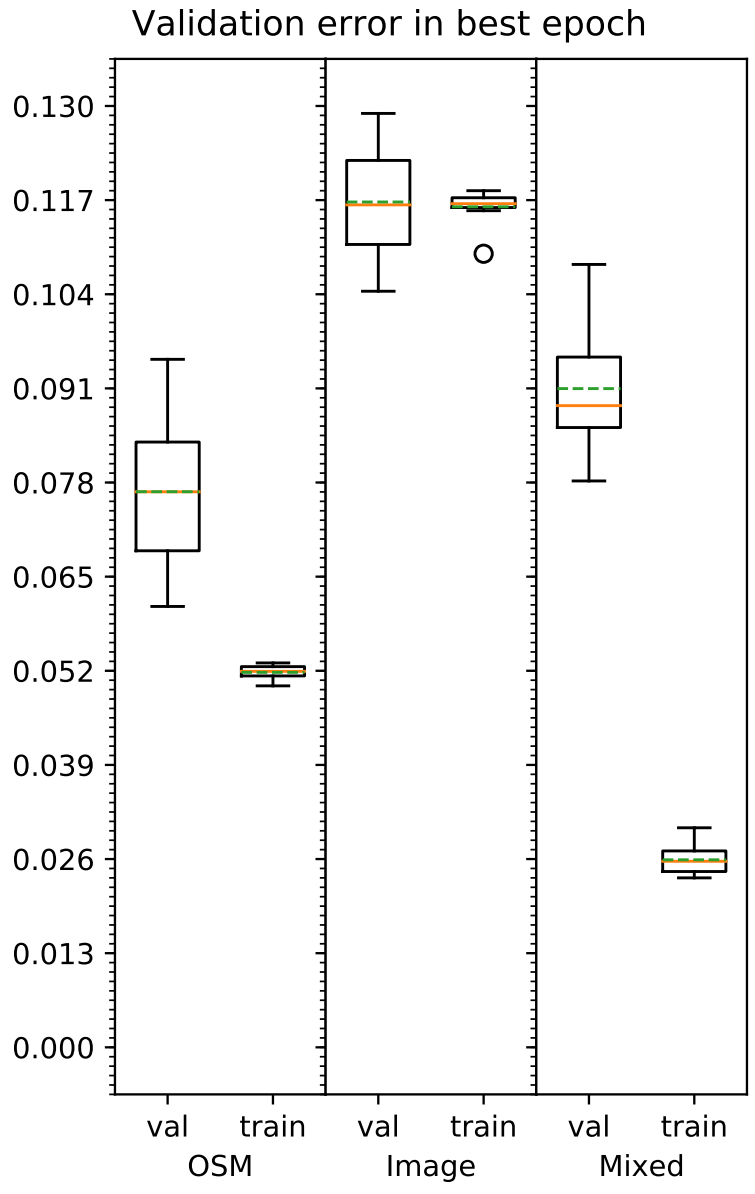

**Figure 6.15.** Model competition, best epoch. Dataset: *5556x_markable_640x640*.

When we compare the results of the basic unexpanded dataset `5556x_markable_640x640` with other expanded datasets, the performance of OSM model actually suffers in the expanded datasets. This is due to the model using only the neighborhood vector information without taking images into account. By the manner of data generation, when splitting long edges, we are also creating new distinct locations, which will be marked with their own distinct OSM vector. But when we simply apply image transformations mentioned in Code 5.9 we are not adding any new unique locations. Therefore the OSM model cannot gain any new information by the data augmentation.

When looking at the model comparison over dataset `5556x_minlen30_640px`, the performance of Mixed model is much closed to the OSM model. Looking directly at Figure B.6, which shows the best achieved average validation error over the whole training process, we see that the Mixed model has in fact very slightly lower validation error in terms of both lower and upper percentile than the OSM model. We can see similar result of B.4, where the Mixed and OSM models have comparable performance over the evolution of epochs, whereas the Image model is lacking.

We can conclude that the neighborhood information brings in useful performance boost, when comparing both the Mixed and OSM models with the Image model. We also encountered the OSM model outperforming the other two more complicated models, only being caught up to by the Mixed model in cases where we used smart dataset enhancing techniques. It should be noted that the OSM model is more compact in terms of number of parameters of the NN as well as being more time efficient for evaluation.

We have taken more care to tweak the performance of this OSM model design in the next experiment.

### 6.3.3 OSM specific - width and depth

This experiment focuses on the performance of OSM model (see 4.2) in different settings of its width and depth. See Figure 4.8, where we can see what these two parameters influence. Note that by increasing width, we are setting the same width for all layers and that there are possible unexplored architectures of models with variable width in different depths (imagine for example a model with first layer wide 128 neurons and second and third layer 32 neurons wide).

We are trying values of (1, 2, 3, 4) for depth and (32, 64, 128, 256) for width. Each model was ran for 10 times with k=10 folds of cross-validation. Out of these runs we preserved the score best achieved validation over the evolution of epochs (to get the score before the model started overfitting). We have a grid of 4x4 possible combinations plotted in Figure 6.16 each with statistics evaluation of the best achieved validation errors.

We chose to display a cut of this table with fixed value of depth to 2 and variable width in Figure 6.17 which follows also their performance over epochs.

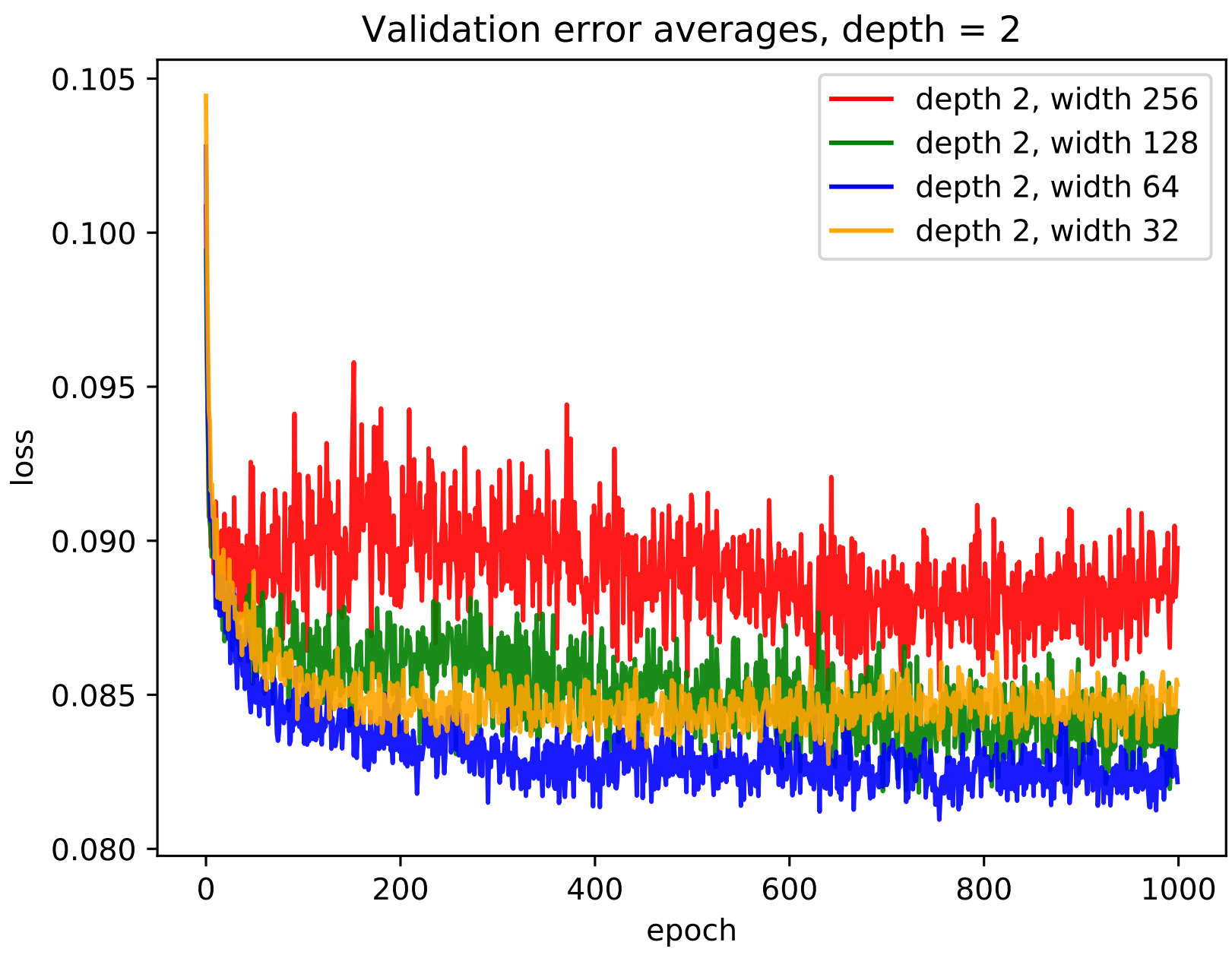

**Figure 6.17.** Model: *OSM*, Dataset: *5556x_markable_640x640*. OSM model with depth fixed at value 2 and variable width.

For more combinations see section B.1.4 in Appendix, where are other cuts (with one attribute fixed while the other one variable) available.

In Figure 6.16 we generated a grid comparing the performances of models with variable depth on the $x$ axis and variable width on the $y$ axis. Width influences the number of neurons each layer has and the depth influences the number of layers. Variation between the results of models with different setting is not very pronounced - even the most basic model with *depth = 1* and *width = 32* is viable. In general we can see the models with *depth* set at lower extreme value *1* to be somewhat lacking behind more balanced models. Similarly on the other end of graph models with *depth* set to *4* seem to have slightly worse results than the simpler models.

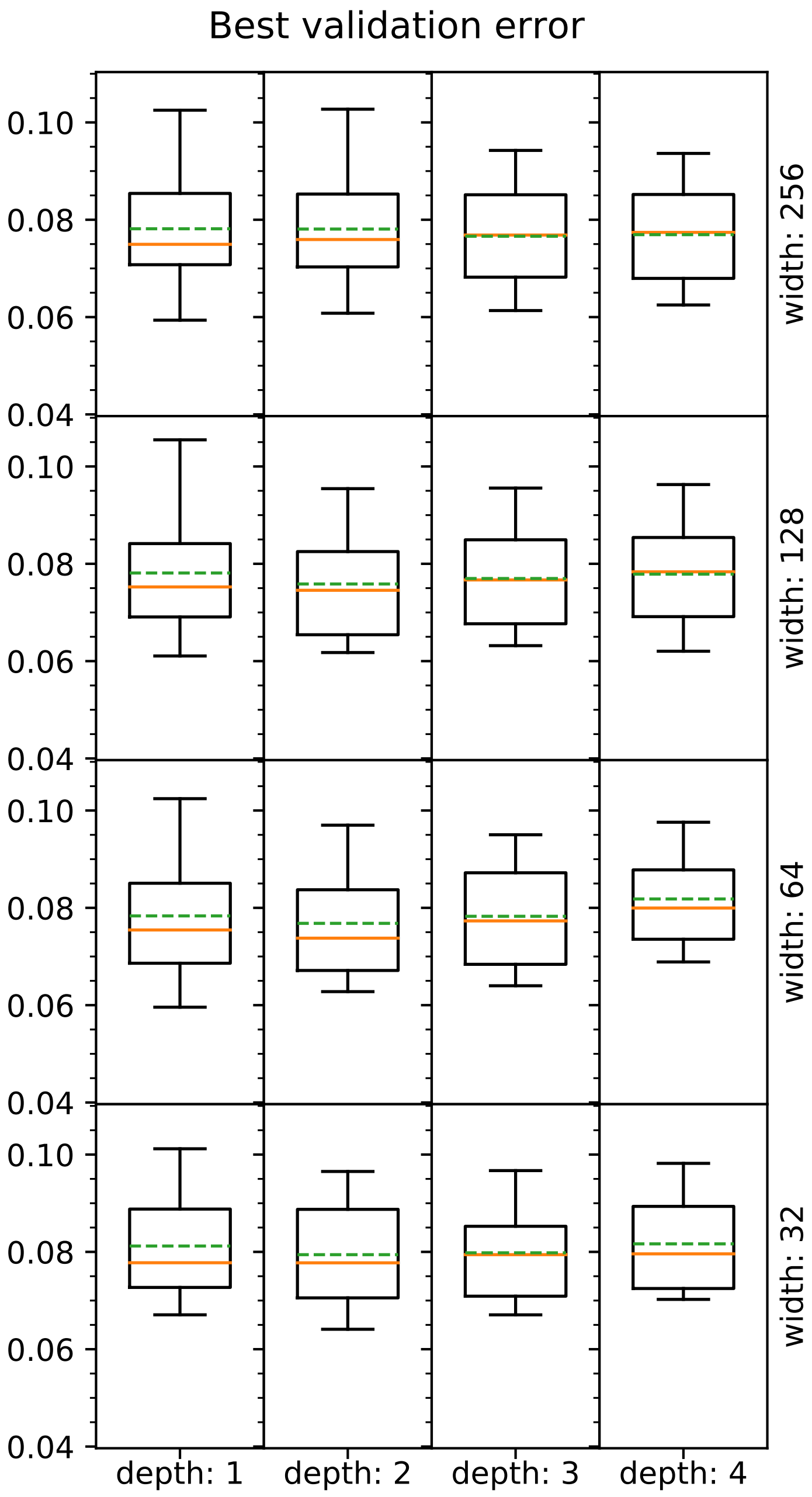

**Figure 6.16.** Model: *OSM*, Dataset: *5556x_markable_640x640*. OSM model alterations in depth and width.

Figure 6.17 allows us to compare amongst each other the models with fixed depth and variable width. Simple models in terms of width can achieve better results through-out the whole training process. Furthermore the smaller the model is in terms of its trainable parameter size, the faster it is taught.

Note that the best performing model can be considered in various terms. When we look at the average best achieved validation error of the 16 models like in Figure 6.16, where average of best results of 10 runs was taken, then model which was built with *depth = 2* and *width = 128* ranks first with mean: 0.0759 +/- 0.0117. See Table 6.18.

If we instead consider the minimal value of average validation error like in Figure 6.17, then the model with *depth = 2* and *width = 64* ranks first with value: 0.0809. See Table 6.19.

| # | w | d | mean(best) | ±std(best) |
|---|---|---|---|---|
| 1 | 128 | 2 | 0.0759 | ±0.0117 |
| 2 | 256 | 3 | 0.0766 | ±0.0107 |
| 3 | 64 | 2 | 0.0768 | ±0.0117 |
| 4 | 256 | 4 | 0.0769 | ±0.0104 |
| 5 | 128 | 3 | 0.0770 | ±0.0110 |
| 6 | 128 | 4 | 0.0779 | ±0.0111 |
| 7 | 256 | 1 | 0.0781 | ±0.0125 |
| 8 | 128 | 1 | 0.0781 | ±0.0128 |
| 9 | 256 | 2 | 0.0781 | ±0.0130 |
| 10 | 64 | 1 | 0.0783 | ±0.0128 |
| 11 | 64 | 3 | 0.0783 | ±0.0109 |
| 12 | 32 | 2 | 0.0794 | ±0.0110 |
| 13 | 32 | 3 | 0.0798 | ±0.0099 |
| 14 | 32 | 1 | 0.0812 | ±0.0109 |
| 15 | 32 | 4 | 0.0816 | ±0.0099 |
| 16 | 64 | 4 | 0.0818 | ±0.0095 |

**Table 6.18.** Model comparison by average best validation error (average over minimal validation error of each fold). Model: *OSM*, Dataset: *5556x_markable_640x640*.

| # | w | d | min(avg) |
|---|---|---|---|
| 1 | 64 | 2 | 0.0809 |
| 2 | 256 | 3 | 0.0810 |
| 3 | 128 | 3 | 0.0811 |
| 4 | 256 | 4 | 0.0812 |
| 5 | 128 | 2 | 0.0818 |
| 6 | 128 | 4 | 0.0822 |
| 7 | 64 | 1 | 0.0826 |
| 8 | 32 | 2 | 0.0828 |
| 9 | 64 | 3 | 0.0828 |
| 10 | 32 | 3 | 0.0833 |
| 11 | 256 | 1 | 0.0834 |
| 12 | 128 | 1 | 0.0839 |
| 13 | 256 | 2 | 0.0847 |
| 14 | 32 | 1 | 0.0848 |
| 15 | 32 | 4 | 0.0864 |
| 16 | 64 | 4 | 0.0873 |

**Table 6.19.** Model comparison by minimal average validation error (minimal of average validation errors over all folds). Model: *OSM*, Dataset: *5556x_markable_640x640*.

## 6.4 Edge evaluation visualization

Once we have trained our model and saved it in its last epoch iteration, we can use it to estimate scores of a custom dataset. We chose to load the original GeoJSON Edge file mentioned in 3.1 and have our model evaluate the edges with missing score. As can be seen in Table 6.20, there is large amount of edges which didn't have initial score. Note that for various models we have to make distinction, of if we can obtain images in the location. For 498 edges we weren't able to download images and only the OSM vectors could have been collected. This reveals another advantage of the OSM model over Image and Mixed models, which depend on presence of imagery data.

Each edge is represented by multiple data entries, in our initial dataset where we didn't use edge splitting this is maximally 6 images or vectors per edge. In some cases we couldn't obtain all 6 images, because Google Street View didn't have imagery information available at the location.

We have used OSM model from experiment 6.3.3 with parameters *width=64* and *depth=2* on dataset *5556x_markable_640x640* to evaluate score of all images with initially missing score. We saved this information to a new GeoJSON file.

Figure 6.21 shows our best performing OSM model with tweaked width of 64 and depth 2 from experiments in 6.3.3. The online resource Carto.com was used to visualize our GeoJSON data with relative ease. Note that in this example we were able to evaluate all present edges, even those without any Google Street View images. This visualization is avilable online at `https://goo.gl/SdNKoC`.

| *Imagery data* | *Edges with images* | *Images* |
|---|---|---|
| Scored | 1396 | 8376 |
| Without score | 3662 | 21696 |
| *OSM vector data* | *Edges with OSM vectors* | *OSM vectors* |
| Scored | 1396 | 8376 |
| Without score | 4160 | 24960 |

**Table 6.20.** Breakdown of scores in the initial dataset. We make a distinction between edges which were initially scored and also between those edges, where we can access imagery information versus where only the OSM vector is available.

Figure 6.22 shows Mixed model taught on the aggressively expanded original dataset which showed great promise in experiment 6.2.3. Note that in this case we can't evaluate all edges as some lacked the imagery information where Google Street View service didn't have data. This visualization is avilable online at `https://goo.gl/GAcFpg`.

These two models evaluate some of the edges differently, which could be expected as their error rate differs. For easier comparison of these two models, we provide another visualization with edges assigned with the value of absolute difference between the scores given by the OSM model from Figure 6.21 and by the Mixed model from Figure 6.22. This auxiliary visualization can be accessed online at `https://goo.gl/iTKML9`. Note that most of the edges exhibit only very small absolute differences.

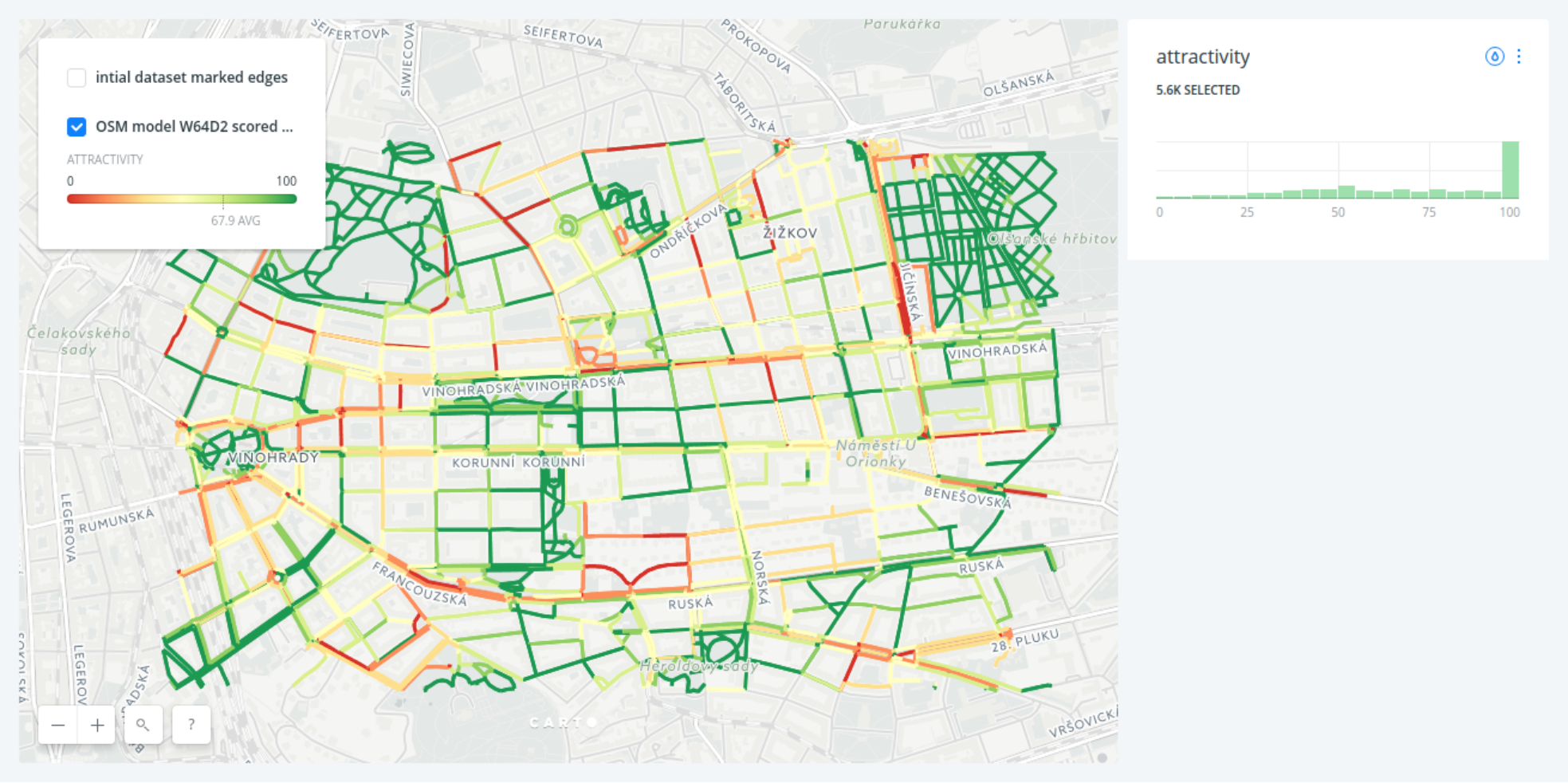

**Figure 6.21.** Model: *OSM model width 64, depth 2*, Dataset: *5556x_markable_640x640* from experiment 6.3.3. GeoJSON file visualization provided by the online service Carto.com. Note that all edges were evaluated as OSM model doesn't depend on presence of imagery information. See this visualization online at `https://goo.gl/SdNKoC`.

## 6.4.1 Edge analysis

Some streets in the initial dataset had relatively low scores which could be explained by the supposed high traffic in the area. This rather low score has spread to close by initially unevaluated edges.

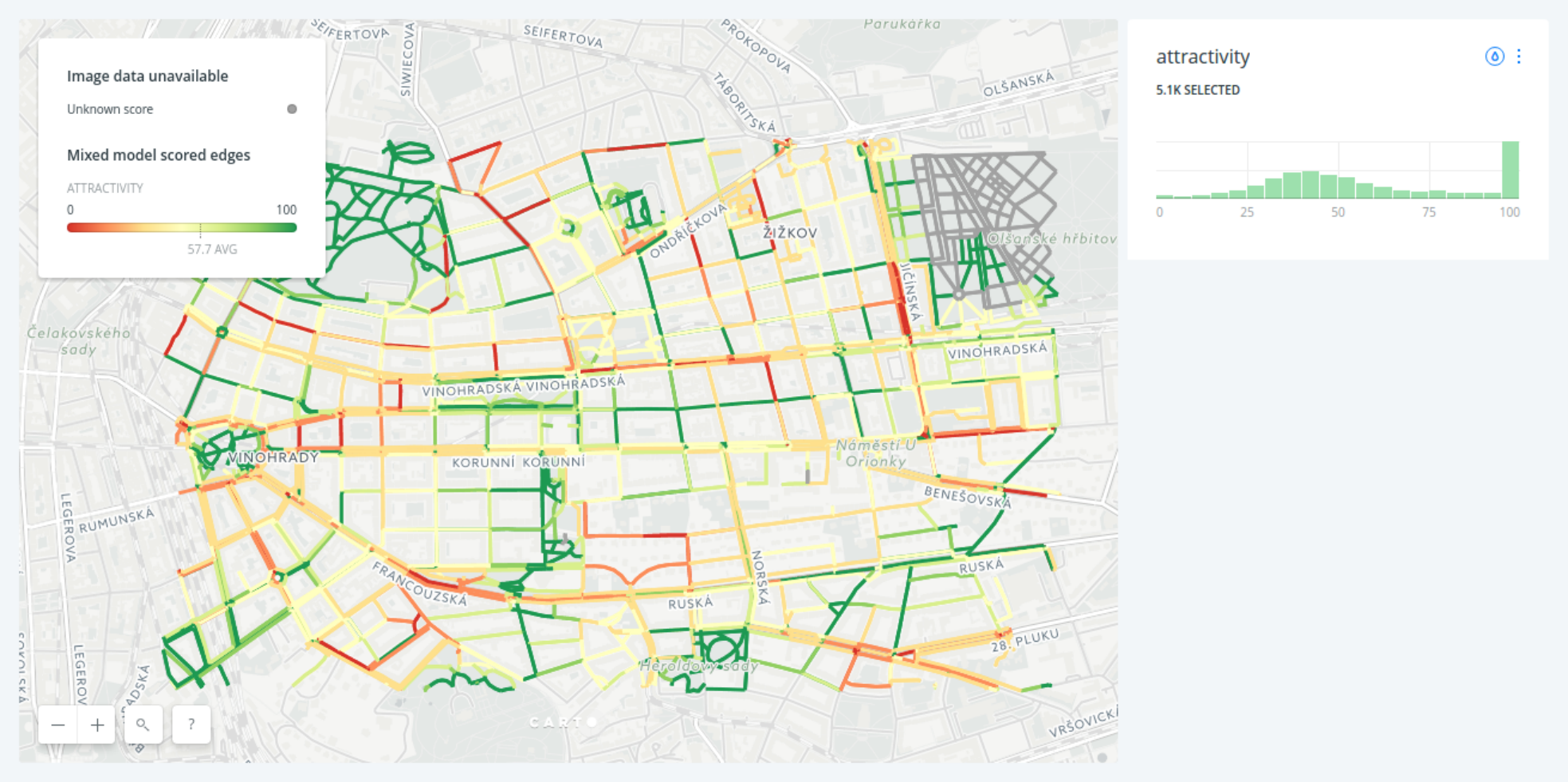

**Figure 6.22.** Model: *Mixed model*, Dataset: the aggressively expanded dataset *5556x_markable_640x640* from experiment 6.2.3. GeoJSON file visualization provided by the online service Carto.com. Note that in this case we rely upon imagery information and thus edges where we couldn't download images are left unscored. See this visualization online at `https://goo.gl/GAcFpg`.

Opposite effect can be seen in parks or locations with the proximity of large amount of natural objects. We can notice the entire area of park will achieve a very high score. See Figure 6.23 for examples of these extreme values.

Locations at the edges of parks and regions with lots of large and frequently used streets seem to achieve mixture of these scores and end up as being neutral (scoring around 50). Similarly streets inside blocks of town areas without parks and without large highways or primary roads will get relatively neutral score. See these neutral edges on Figure 6.24. Note that there is still variability in the score depending on the subtle changes in their neighborhoods.

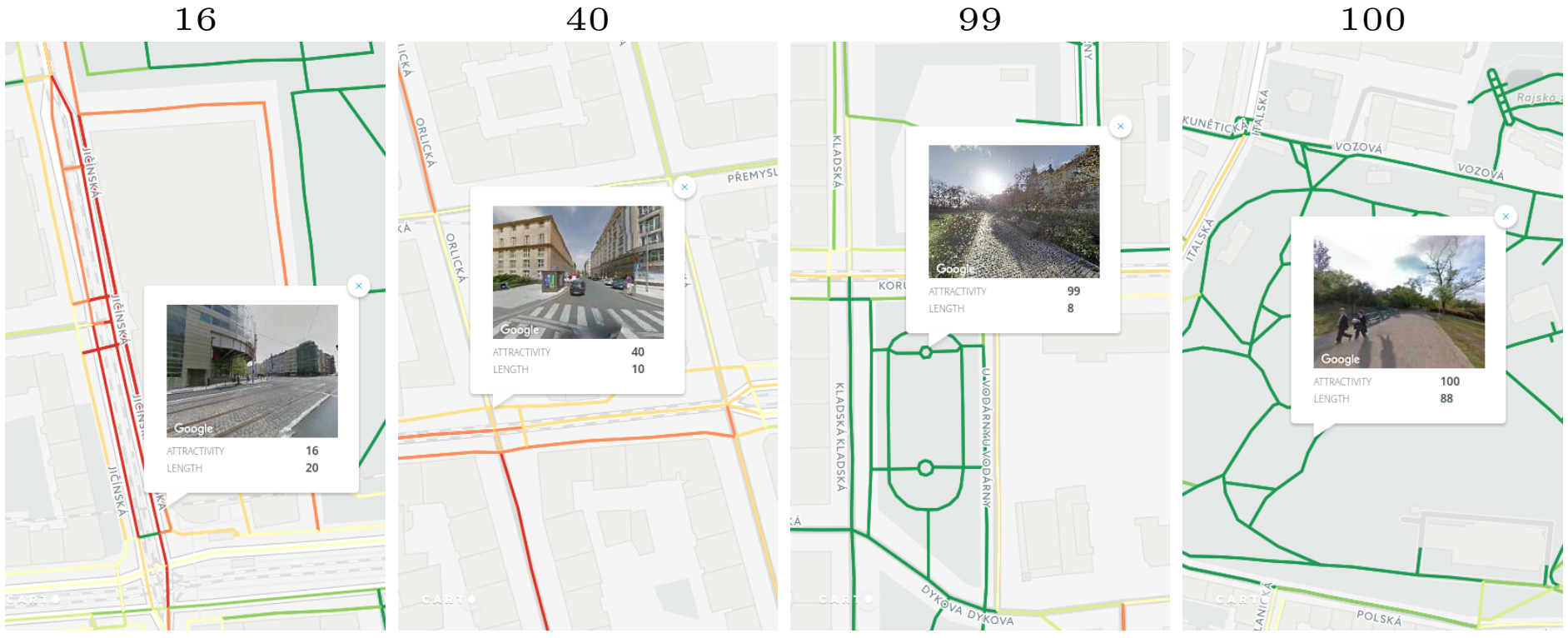

**Figure 6.23.** Model: *OSM model width 64, depth 2*, Dataset: *5556x_markable_640x640*. Details of the evaluated dataset. We can notice the tendency to give parks and locations in proximity of greenery high scores. Note that this is a OSM model rating solely on the neighborhood vector data.

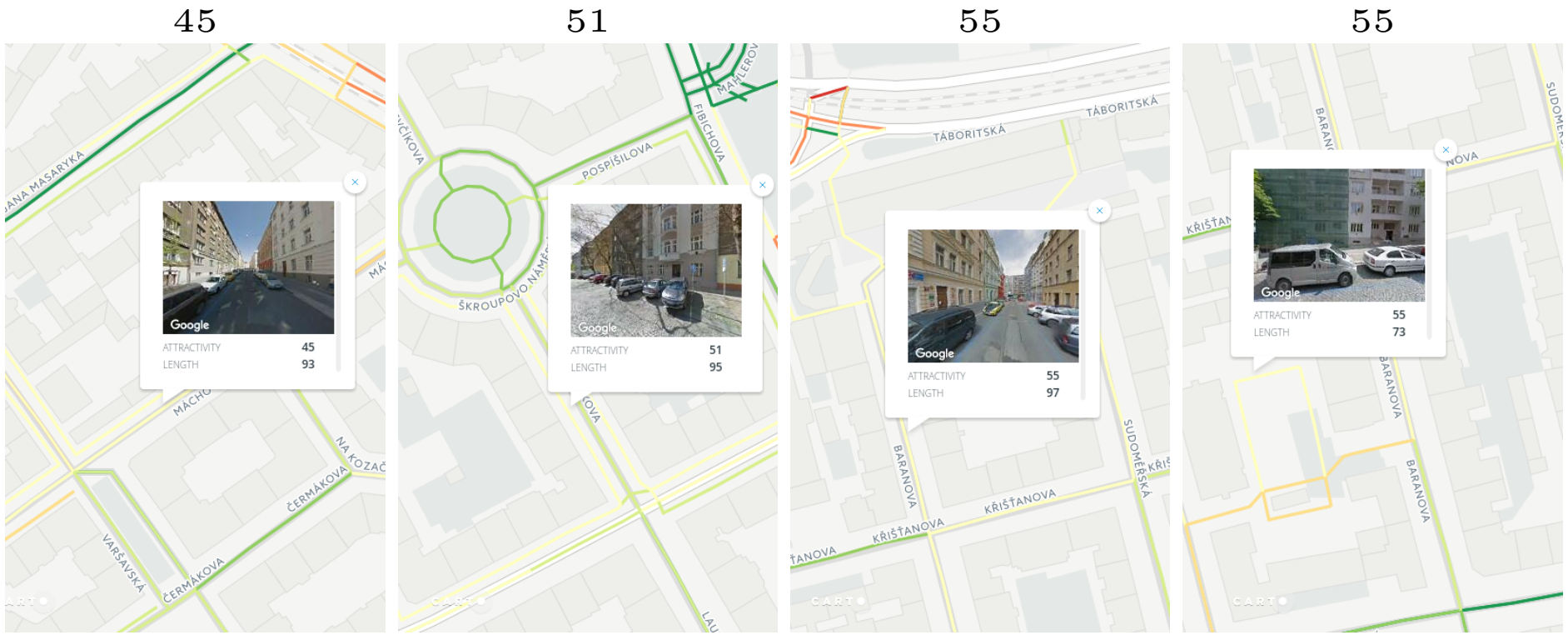

**Figure 6.24.** Model: *OSM model width 64, depth 2*, Dataset: *5556x_markable_640x640*. Details of the evaluated dataset. We have selected samples with neutral score (around 50, neither bad nor good to take).

## 6.5 Limits of indiscriminate dataset expansion

There are two main fields of strategies we can analyze when approaching our task – dataset augmentation in 6.2 and model design in 6.3. As was discussed in section 3.2, there are some ways to alter the scheme of generating data from edges. We understand data as lists of items each with a triple (image, OSM vector, score). When expanding the initial dataset with imagery information, we chose a scheme of generating six non-overlapping images per segment. We rotate by $120^\circ$ degrees two sides of each segment. Furthermore as described in 3.4 we chose to split long edges into smaller segments and then generate images from them.

It could be argued that it is in our interest to generate as many images as possible, which would encourage breaking long edges into many very short segments and perhaps also adapting the amount of degrees we rotate around on the same spot. In theory the continuous quality of real world could make us want to sample it in the most detailed way, generating an infinity of distinct images. However the practical reality we are faced with is very much different. In fact we are working with a discrete representation of real world into a limited set of images. If we ask Google Street View for samples too close to each other, we will in fact obtain the same images from its API. Similarly choosing to rotate for less than $120^\circ$ degrees would produce images with vastly overlapping content.

Both methods of new data generation in 6.2.2 and 6.2.3 can face the same problem of being wasteful and not bringing in any meaningful data. A balance needs to be found in all these cases.

Note that increasing number of images inevitably also introduces an increase in time and resource consumption. We should consider the advantages of being economical with resources as well as watching the validation error.

## 6.6 Discussion

We have tried experimenting both with the definitions of models and with the construction of datasets in this chapter. Some methods were noticeably effective in terms of models performance, while other were negligible. In some case we chose the only feasible solution for the rest of the experiments, such as in the case of the base CNN model choice in 6.3.1. The example of 6.2.1 revealed that in our case bigger doesn't always mean better. The experiments of 6.2.2 and 6.2.3 showed us that in most cases our attempts to enlarge the dataset by either data augmentation or more complex method of preprocessing initial dataset by edge splitting are rewarded with better precision.

When compared, the results of edge splitting in Figure B.1 fall behind the more pronounced performance boost of data augmentation in Figure 6.9, especially in the case of more aggressive augmentation scheme.

However the results of Figure 6.10 reveal that some of the attempts to improve our dataset cannot be efficiently stacked. The already enlarged dataset `5556x_minlen30_640px`, produced by refined edge splitting didn't benefit from subsequent data augmentation scheme.

Results of the model competition in section 6.3.2 reveal that the simple OSM model fared well even in comparison with its more complex competition. While also carrying the benefits of decreased computation and memory requirements, this result is particularly interesting. The Mixed model didn't fall very far behind and further was able to deal very successfully in otherwise difficult situations, where for example the Image model failed. We can see this when we compare the results of Mixed model with Image model on Figure 6.3 and also particularly well on the Figure 6.15.

The OSM model showed its capability when compared with other more complicated models. Furthermore in Figure 6.16 we can see that there is space for improvement in the way it is built. Balancing the complexity of the model, in the sense of amount of layers and number of neurons per layer (parametrically the width and depth of the model), shows that some configurations work better than other.

It can be noted that researching both the Mixed and OSM model is perspective.

## 6.7 Future Work

While the result overviewed in 6.6 are encouraging, they also show that there is space for further improvement in future work.

We propose the usage of more varied datasets in terms of their original locations. Having more rich data background, we could interact with more general features of images across differing geographical locations and temporal timestamps. We suggest that the usage of more varied dataset could be beneficial even for the case of data based from a singular source. Teaching our model on datasets from various cities, landscape types and seasons could improve the precision of the road scoring in any particular city. Models should be less prone to overfitting with richer datasets.

The shapes of our selected CNNs could be explored in more detail, as the results of 6.16 are already promising. Hyper parameter optimization over model settings could introduce further refinement of already proposed models.

The results of Mixed model which combines two data sources are favorable, even if they lag behind the simpler OSM model. We could try adjusting the imagery data influence in an attempt to have the Image model section of the Mixed model give us more precision rather than to weaken the contribution of the vector neighborhood data. We argue that there is a more balanced architecture to be found, which would improve over the OSM model by combining multiple sources of data. We suggest that altering the dimensionality of the images could be a promising approach as is supported by the results of section 6.2.1 and also as it would not drastically increase the resource requirements of the model.

# Chapter 7
# Conclusion

Following the research of the state of the art techniques on object identification in imagery data, we have successfully adjusted Convolutional Neural Network models for the task of route attractivity score estimation. We thoroughly researched the advances of CNN models in the ImageNet Large Scale Visual Recognition Competition as well as other related works.

We have enhanced the initial dataset of scored edges and nodes with imagery information downloaded from Google Street View and with vector representation of neighborhood structures aggregated from Open Street Maps. We introduced two major ways of further extending the dataset with the method of long edge refinement into smaller segments and by the dataset augmentation.

To overcome the limitations imposed by relatively small dataset size in terms of richness of unique images, we followed the method of feature transfer. We reused feature extractor section of complex CNN model `resnet50` trained on the large `ImageNet` dataset while adapting it for our own task by attaching a custom classifier and fine-tuning it on the task of score estimation.

Models and experiments were constructed with efficiency in mind by reusing the once computed image features across multiple runs enabling faster model prototyping. As a result we were able to experiment with various CNN model designs and variable generation schemes for generating and augmenting datasets.

We proposed three CNN model architectures according to the data types used to asses the informational value of each data source. We designed Image model using only the imagery information, OSM model using only the vector neighborhood representation and Mixed model using both data sources. We have made the model descriptions customizable and experimented with changing parameters influencing their shape, width and depth.

We have tested various setting combinations extensively while comparing their results with k-fold cross-validation. Our experiments explored the search space of model definition settings as well as dataset augmentations. The results were studied and compared to find underlying principles and suggest possible improvements.

We report the best model achieving mean squared error measurement of 0.076 +/- 0.012 on validation data.

Lastly we have evaluated the initial dataset with selected models and filled in the attractivity score to the formerly unlabeled edges. We provide online dataset visualization using the Carto.com service.

# Appendix A
# Abbreviations

## A.1 Abbreviations

In some cases we used abbreviations to simplify the text – here is the whole list.

| | |
|---|---|
| ATG | Agent Technology Center at CTU |
| CTU | Czech Technical University in Prague |
| CNN | Convolutional Neural Network |
| CPU | Central processing unit |
| GPU | Graphics Processing Unit |
| ILSVRC | ImageNet Large Scale Visual Recognition Competition |
| MNIST | Modified National Institute of Standards and Technology |
| SQL | Structured Query Language |

# Appendix B
# Additional graphs

## B.1 Additional graphs

### B.1.1 Splitting long edges

Figures 6.6 and 6.7 show additional graphs to the experiment in 6.2.2.

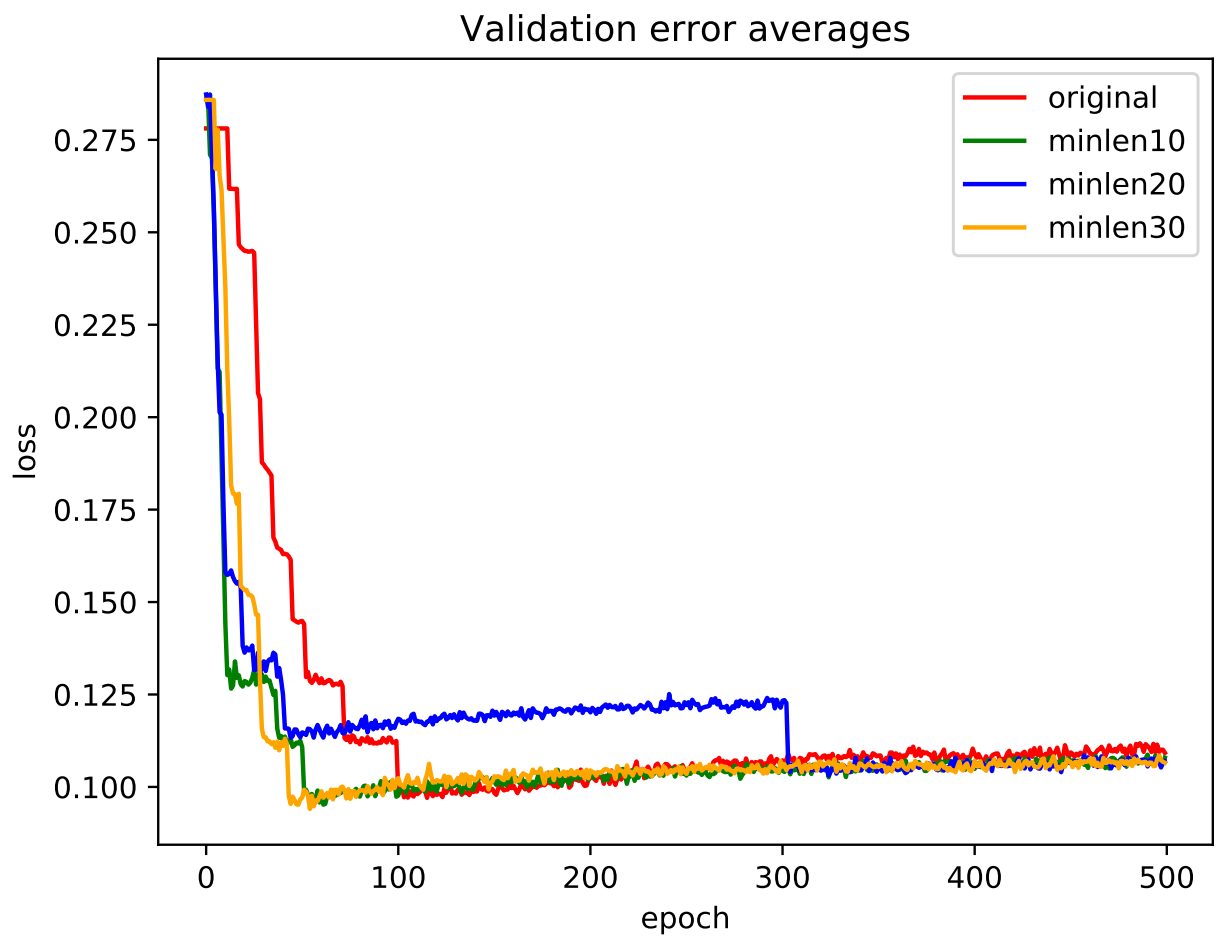

**Figure B.1.** Model: *Mixed*, Datasets: *original*, *minlen10*, *minlen20*, *minlen30*. Averages of validation errors with edge splitting experiment.

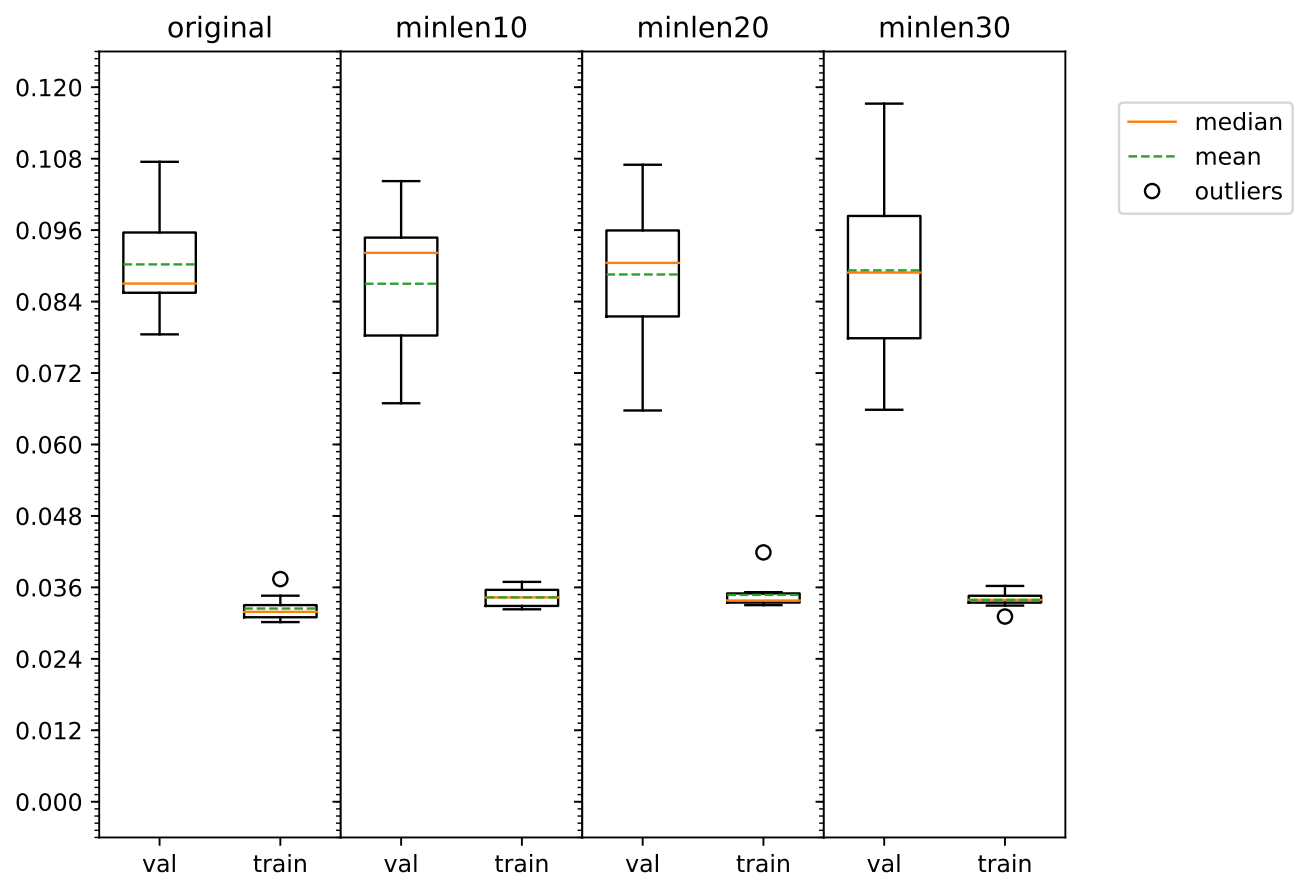

**Figure B.2.** Model: *Mixed*, Datasets: *original*, *minlen10*, *minlen20*, *minlen30*. Best epoch of the edge splitting experiment.

### B.1.2 Different CNN base model for feature transfer

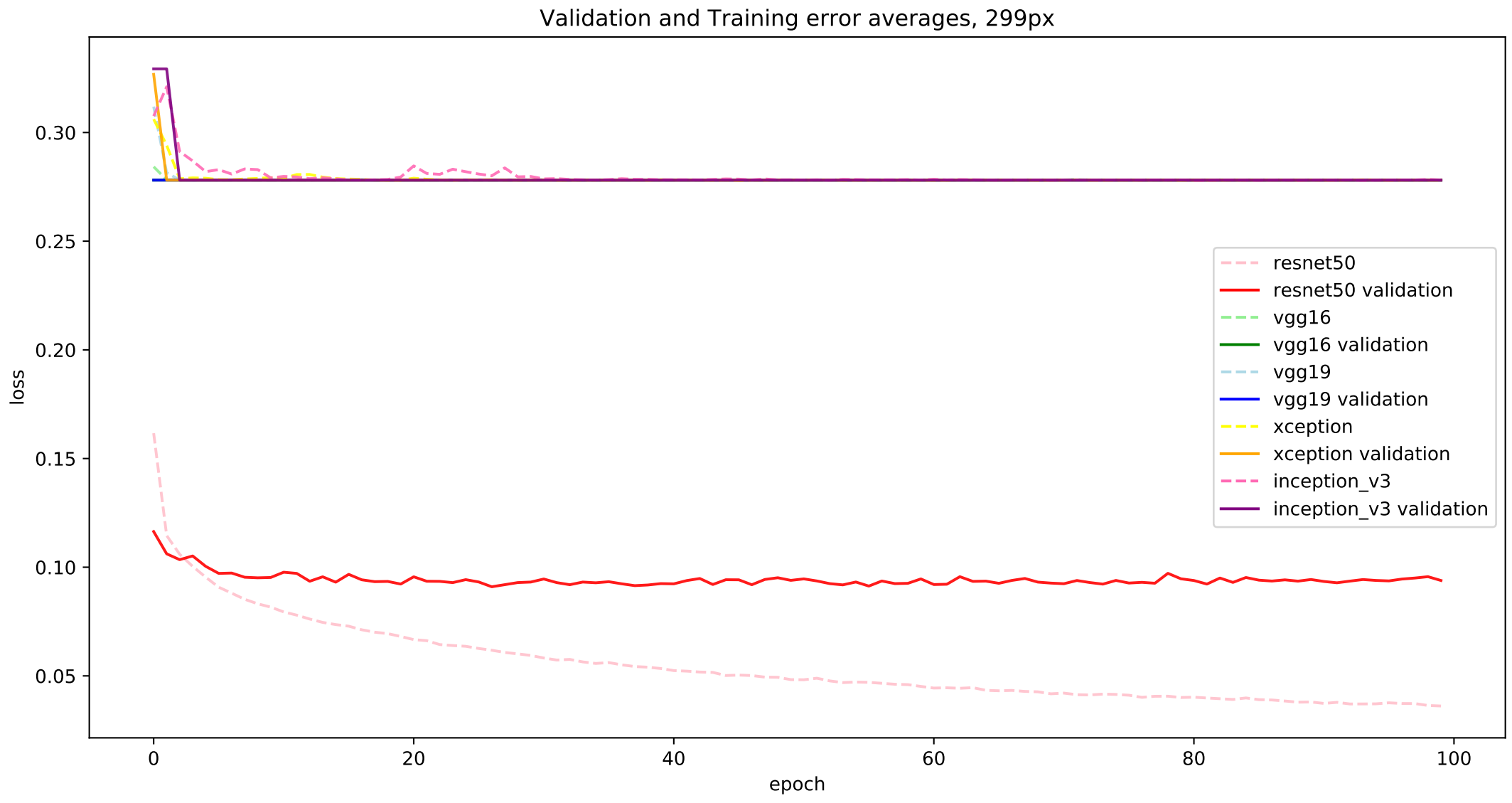

**Figure B.3.** Model: *Mixed with variable base*, Dataset: *5556x_mark_res_299x299*. Illustration of five models and their averages of validation and training errors. Note that this dataset uses images of 299x299 pixel size.

### B.1.3 Model competition - Image vs. OSM vs. Mixed

Figures B.4, B.5 and B.6 provide additional graphs for the section 6.3.2. Here we can see the `5556x_minlen30_640px` dataset.

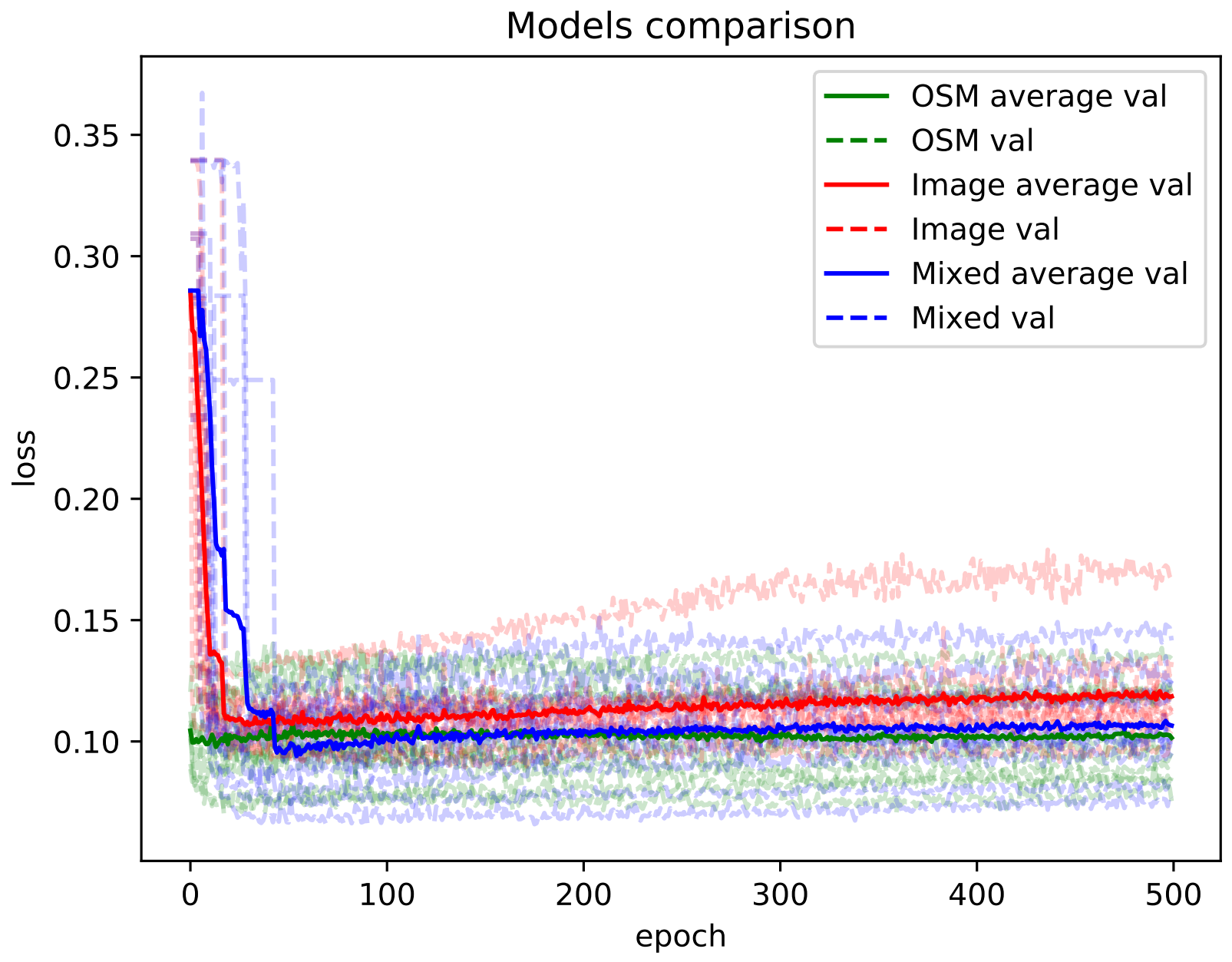

**Figure B.4.** Models: *OSM*, *Image*, *Mixed*, Dataset: *5556x_minlen30_640px*. Model competition, evolution over epochs.

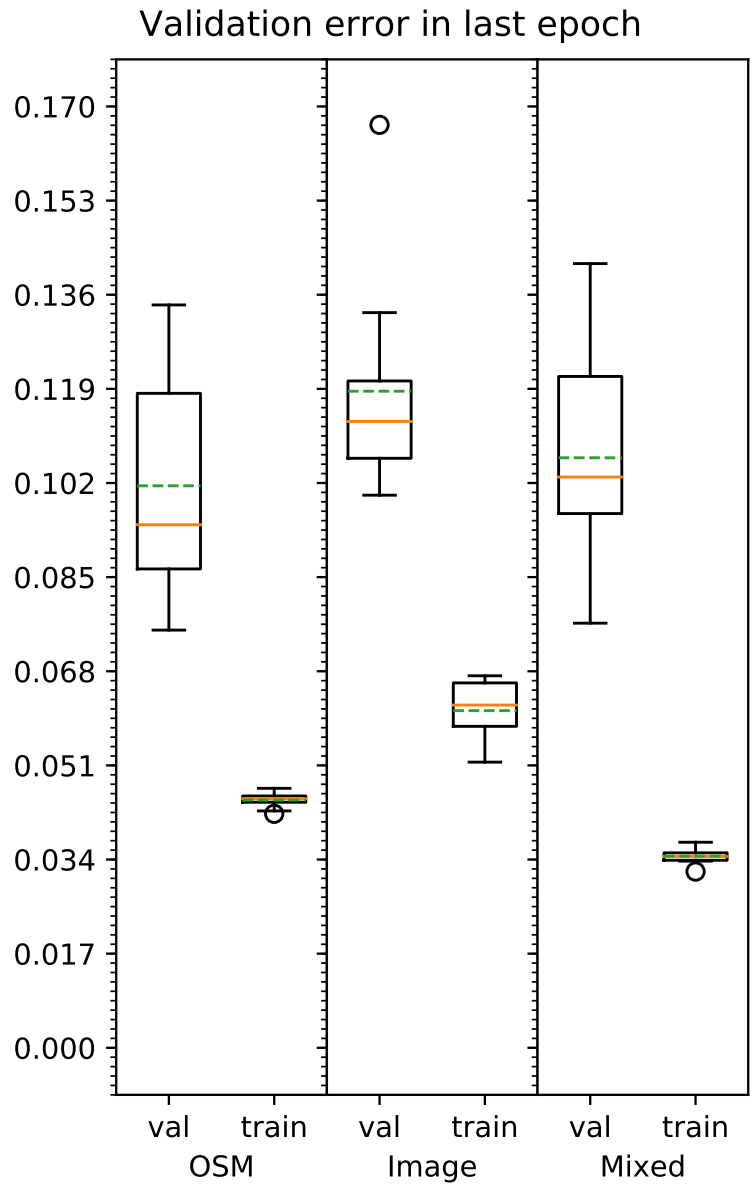

**Figure B.5.** Model competition, last epoch. Dataset: *5556x_minlen30_640px*.

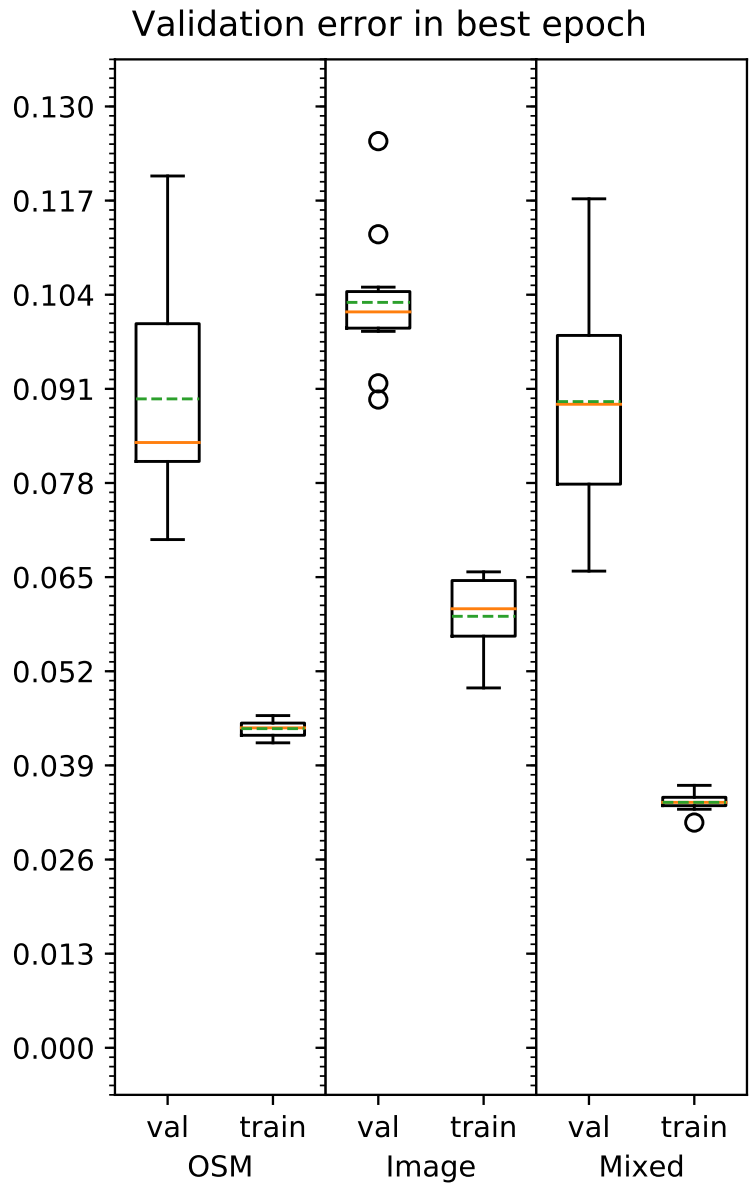

**Figure B.6.** Model competition, best epoch. Dataset: *5556x_minlen30_640px*.

## B.1.4 OSM specific - width and depth

Figure B.7 adds to the experiment of 6.3.3

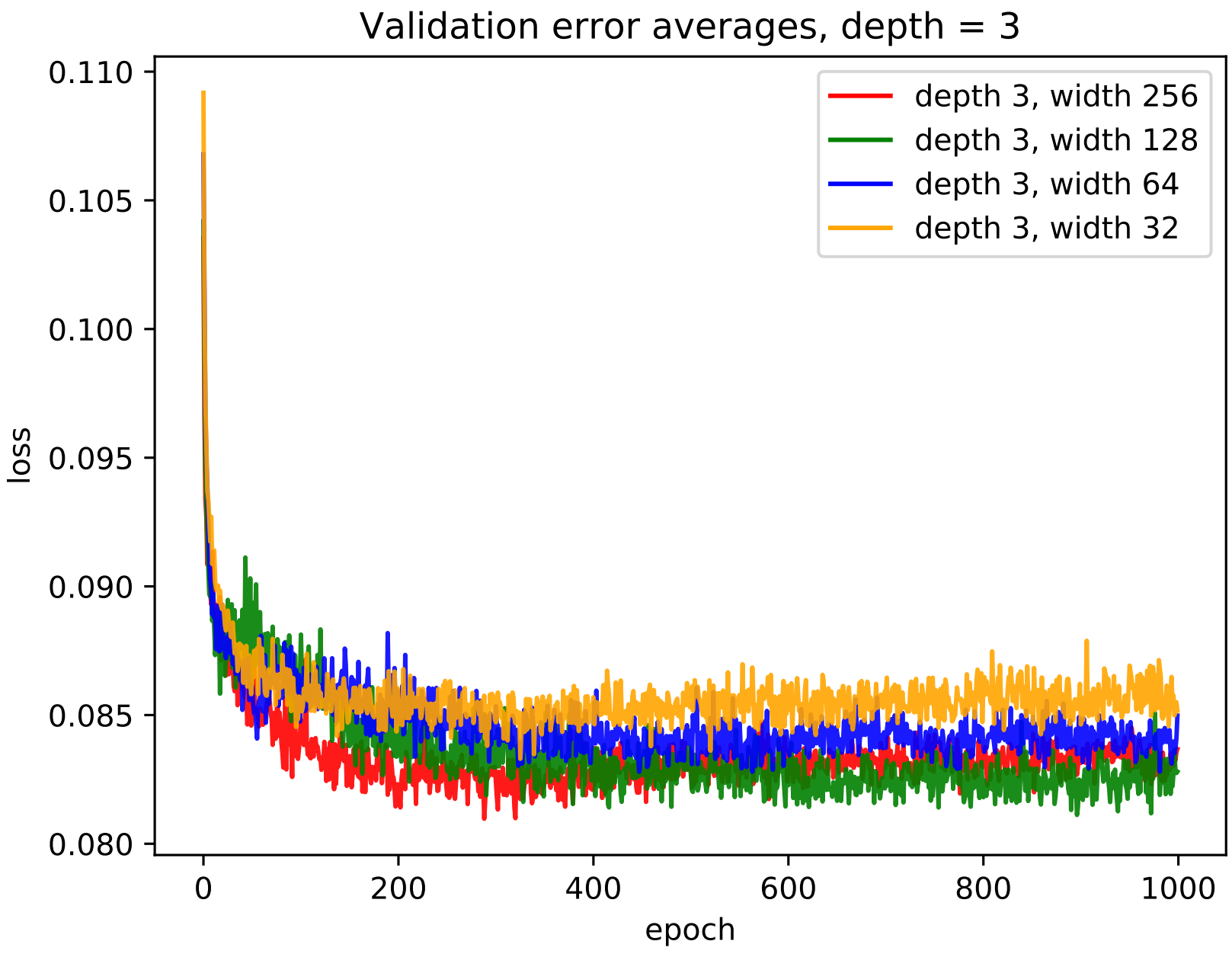

**Figure B.7.** Model: *OSM*, Dataset: *5556x_markable_640x640*. OSM model with depth fixed at value 3 and variable width.

# Appendix C
# Dataset overview

## C.1 OSM vector details

Table C.8 shows us the list of attribute and value pairs, which build up the OSM vectors. We are interested only in the occurances of the pairs in this list. Note that we show only a start of the vector, which is in total made of 594 attribute-value pairs. Refer to section 3.3.1 for details.

| # | *attribute=value pair* | # | | # | |
|---|---|---|---|---|---|
| 0 | building=yes | 30 | barrier=fence | 60 | natural=coastline |
| 1 | highway=residential | 31 | amenity=parking | 61 | building=detached |
| 2 | building=house | 32 | landuse=grass | 62 | amenity=place_of_worship |
| 3 | highway=service | 33 | surface=paved | 63 | barrier=hedge |
| 4 | highway=track | 34 | bicycle=yes | 64 | highway=living_street |
| 5 | waterway=stream | 35 | highway=crossing | 65 | amenity=school |
| 6 | highway=unclassified | 36 | building=apartments | 66 | building=shed |
| 7 | power=tower | 37 | highway=primary | 67 | area=yes |
| 8 | oneway=yes | 38 | landuse=meadow | 68 | highway=trunk |
| 9 | natural=tree | 39 | highway=bus_stop | 69 | building=roof |
| 10 | natural=water | 40 | barrier=wall | 70 | bicycle=no |
| 11 | highway=footway | 41 | railway=rail | 71 | service=alley |
| 12 | building=residential | 42 | natural=scrub | 72 | highway=motorway |
| 13 | surface=asphalt | 43 | highway=turning_circle | 73 | highway=street_lamp |
| 14 | highway=path | 44 | natural=wetland | 74 | amenity=bench |
| 15 | highway=tertiary | 45 | boundary=administrative | 75 | highway=steps |
| 16 | access=private | 46 | intermittent=yes | 76 | religion=christian |
| 17 | natural=wood | 47 | place=locality | 77 | tunnel=yes |
| 18 | landuse=residential | 48 | building=hut | 78 | public_transport=platform |
| 19 | power=pole | 49 | waterway=ditch | 79 | amenity=restaurant |
| 20 | landuse=forest | 50 | barrier=gate | 80 | leisure=swimming_pool |
| 21 | landuse=farmland | 51 | surface=ground | 81 | foot=designated |
| 22 | surface=unpaved | 52 | leisure=pitch | 82 | leisure=park |
| 23 | highway=secondary | 53 | place=village | 83 | waterway=drain |
| 24 | bridge=yes | 54 | place=hamlet | 84 | landuse=farmyard |
| 25 | service=driveway | 55 | building=industrial | 85 | admin_level=8 |
| 26 | foot=yes | 56 | surface=gravel | 86 | highway=motorway_link |
| 27 | building=garage | 57 | waterway=river | 87 | railway=level_crossing |
| 28 | service=parking_aisle | 58 | highway=traffic_signals | 88 | tunnel=culvert |
| 29 | oneway=no | 59 | highway=cycleway | ... | ... |

**Table C.8.** Examples of attribute=value pairs used to filter out which values are of interest to us while building the OSM vector.

## C.2 Used datasets overview

Table C.9 shows the list of the used datasets. These can be roughly divided by their purpose and method of generation. We can see those created by the edge splitting technique from 6.2.2 and those generated with data augmentation schemes from 6.2.3. We also have several datasets used purely for fast prototyping and experimentation, which can serve as simple examples.

| | |
|---|---|
| Proper datasets: | |
| `5556x_markable_640x640` | Original dataset |
| `5556x_mark_res_299x299` | Resized original dataset to the dimension of 299x299 pixels |
| `5556x_minlen30_640px`, `5556x_minlen20_640px`, `5556x_minlen10_640px` | Edge splitting technique used with variable minimal edge length |
| `5556x_reslen30_299px` | Resized version of 5556x_minlen30_640px dataset to 299x299 pixels |
| Expanded datasets: | |
| `5556x_markable_640x640_2x_expanded` | Expanded original dataset |
| `5556x_markable_640x640_2x_agressive_expanded` | Secondary scheme of augmentation on original dataset |
| `5556x_minlen30_640px_2x_expanded` | Expanded dataset with split edges |
| `5556x_minlen30_640px_2x_agressive_expanded` | Secondary scheme of augmentation on dataset with split edges |
| Subsets of datasets for fast prototyping | |
| `1200x_markable_299x299` | Roughly one fifth of all usable data, sized 299x299 |
| `miniset_640px` | Minimal dataset example with just 24 images. |

**Table C.9.** List of used datasets divided by their use.

## C.3 Dataset examples

### C.3.1 Original dataset

Original dataset `5556x_markable_640x640` containing images of pixel size 640x640 in Figure C.10.

### C.3.2 Expanded dataset

Expanded dataset `5556x_markable_640x640_2x_expanded` with images generated expanding the original dataset using the first augmentation scheme of 5.9 in Figure C.11.

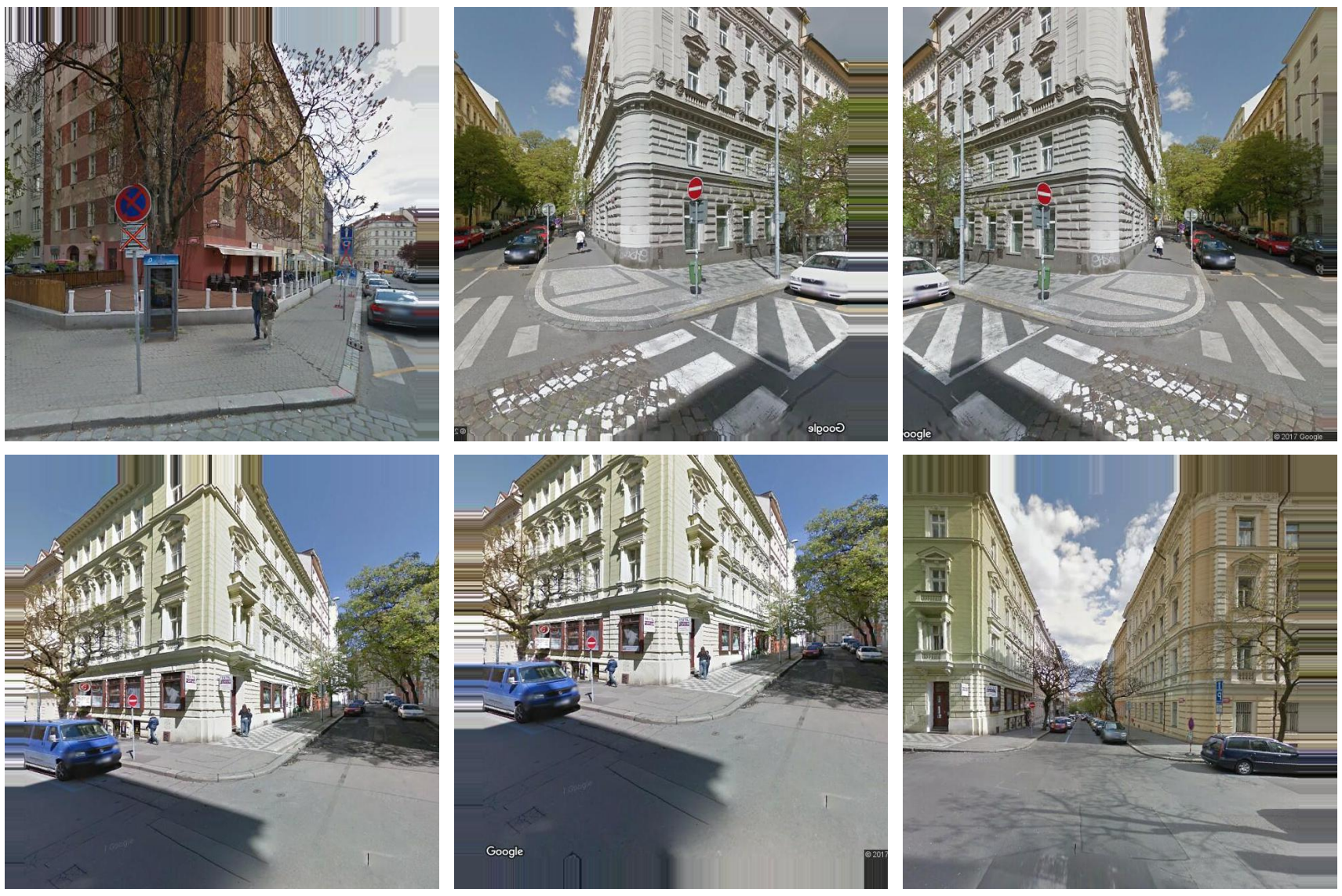

**Figure C.11.** Examples of the dataset `5556x_markable_640x640_2x_expanded`.

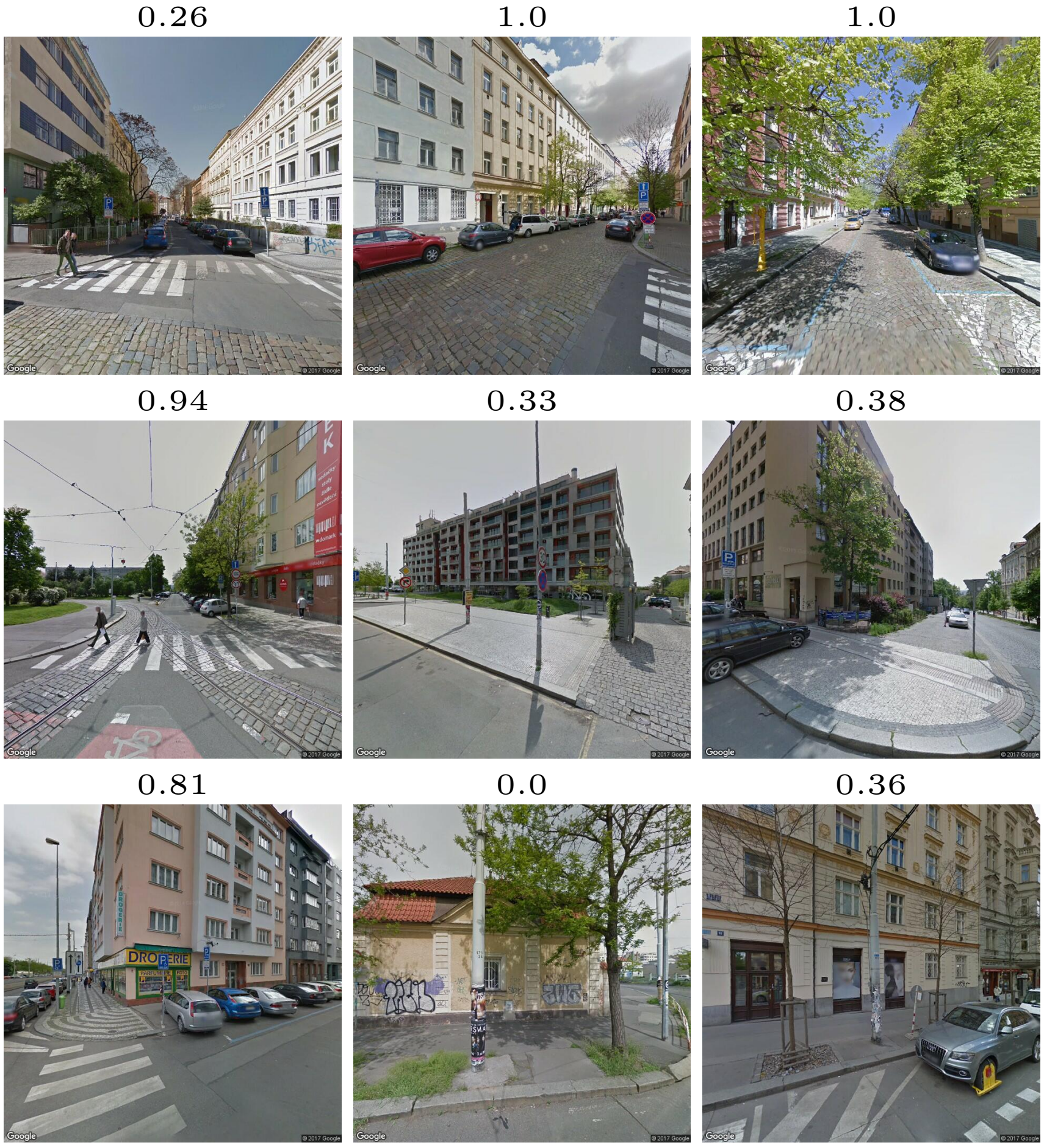

**Figure C.10.** Scored examples of the dataset `5556x_markable_640x640`. Note that the score labels come from the initial dataset.

### C.3.3 Aggressively extended dataset

Expanded dataset `5556x_markable_640x640_2x_agressive_expanded` with images generated expanding the original dataset using the second, more aggressive augmentation scheme of 5.9 in Figure C.12.

## C.4 Dataset analysis

We look at the composition of dataset entries in terms of edge length before their splitting in Figures C.13, C.15 and C.17. We look at the score distribution in Figures C.14, C.16 and C.18.

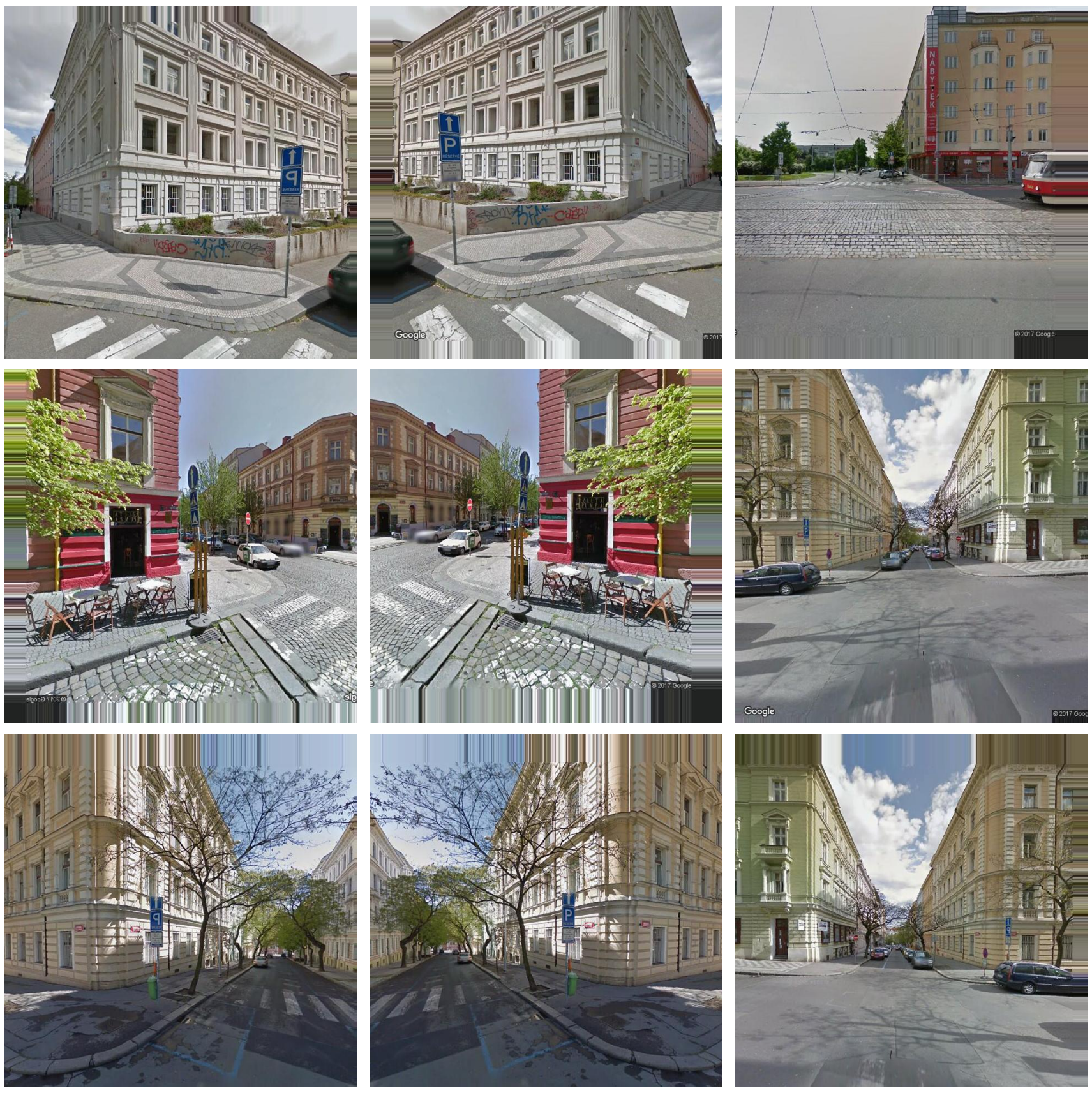

**Figure C.12.** Examples of the dataset `5556x_markable_640x640_2x_agressive_expanded`.

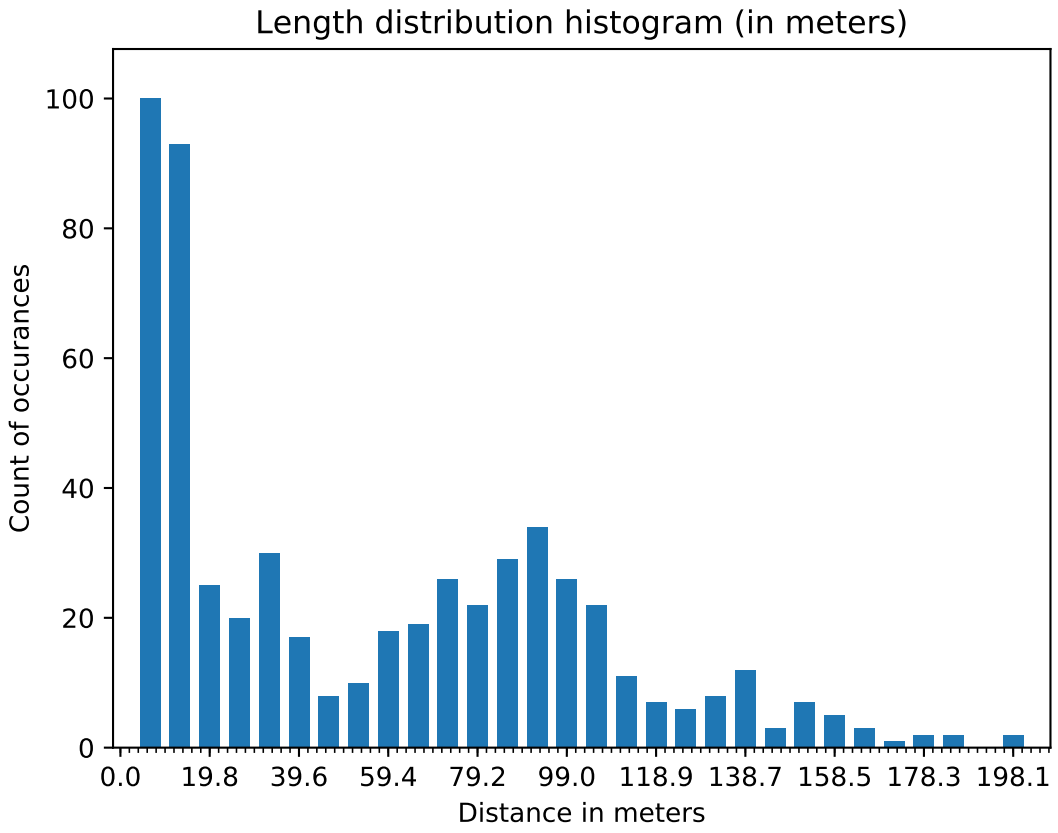

**Figure C.13.** Histogram of edge lengths from the initial dataset *5556x_markable_640x640*.

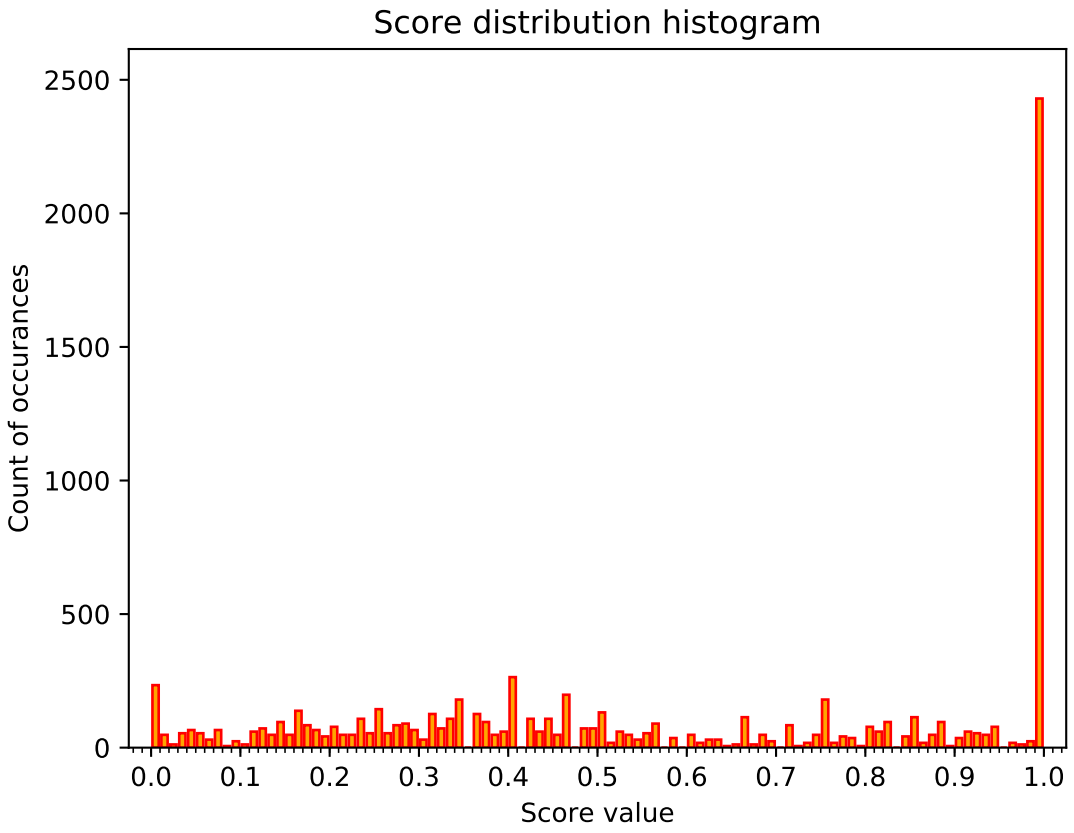

**Figure C.14.** Histogram of image scores from the initial dataset *5556x_markable_640x640*.

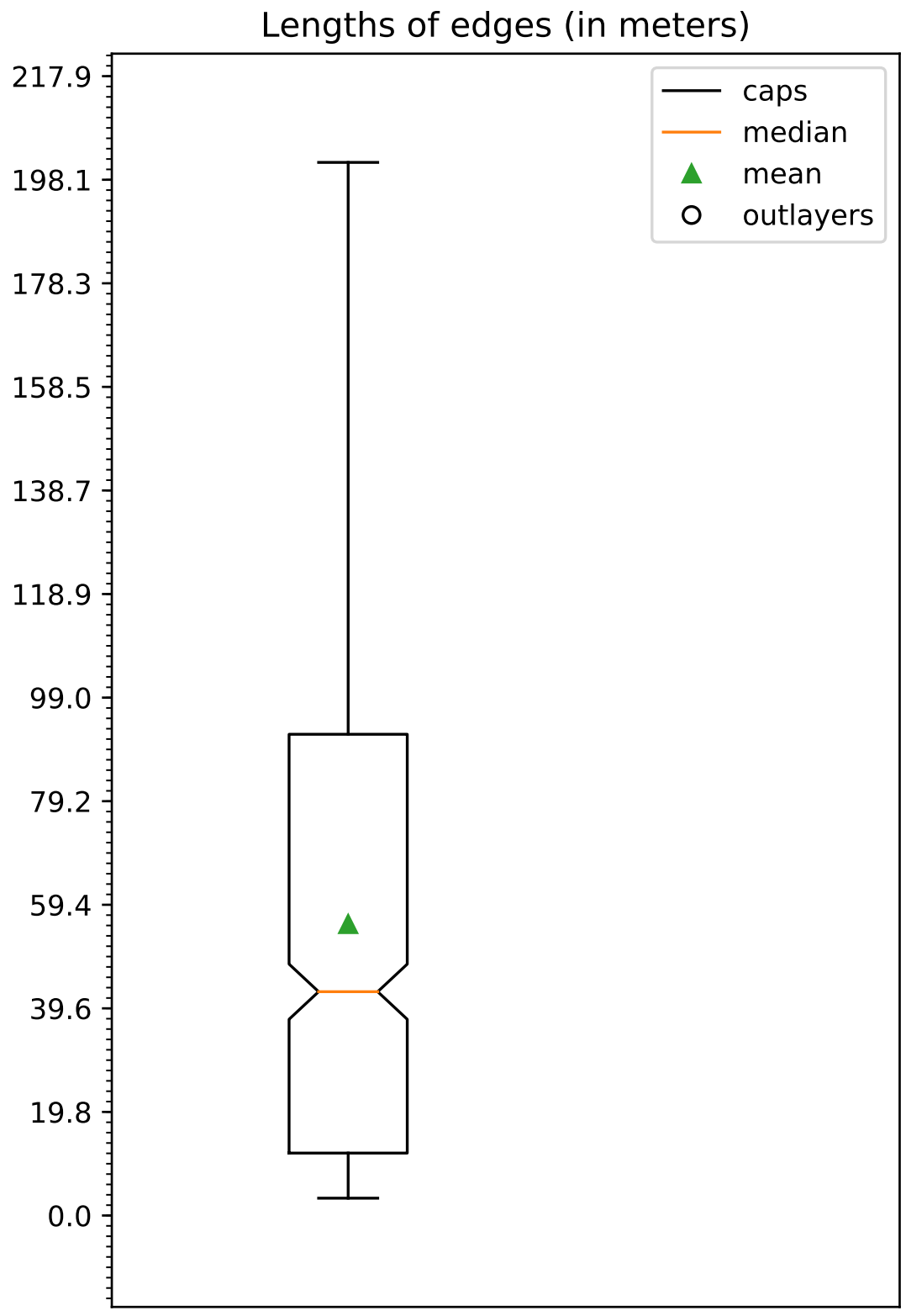

**Figure C.15.** Edge lengths in the initial dataset *5556x_markable_640x640*. We split this dataset in experiment 6.2.2.

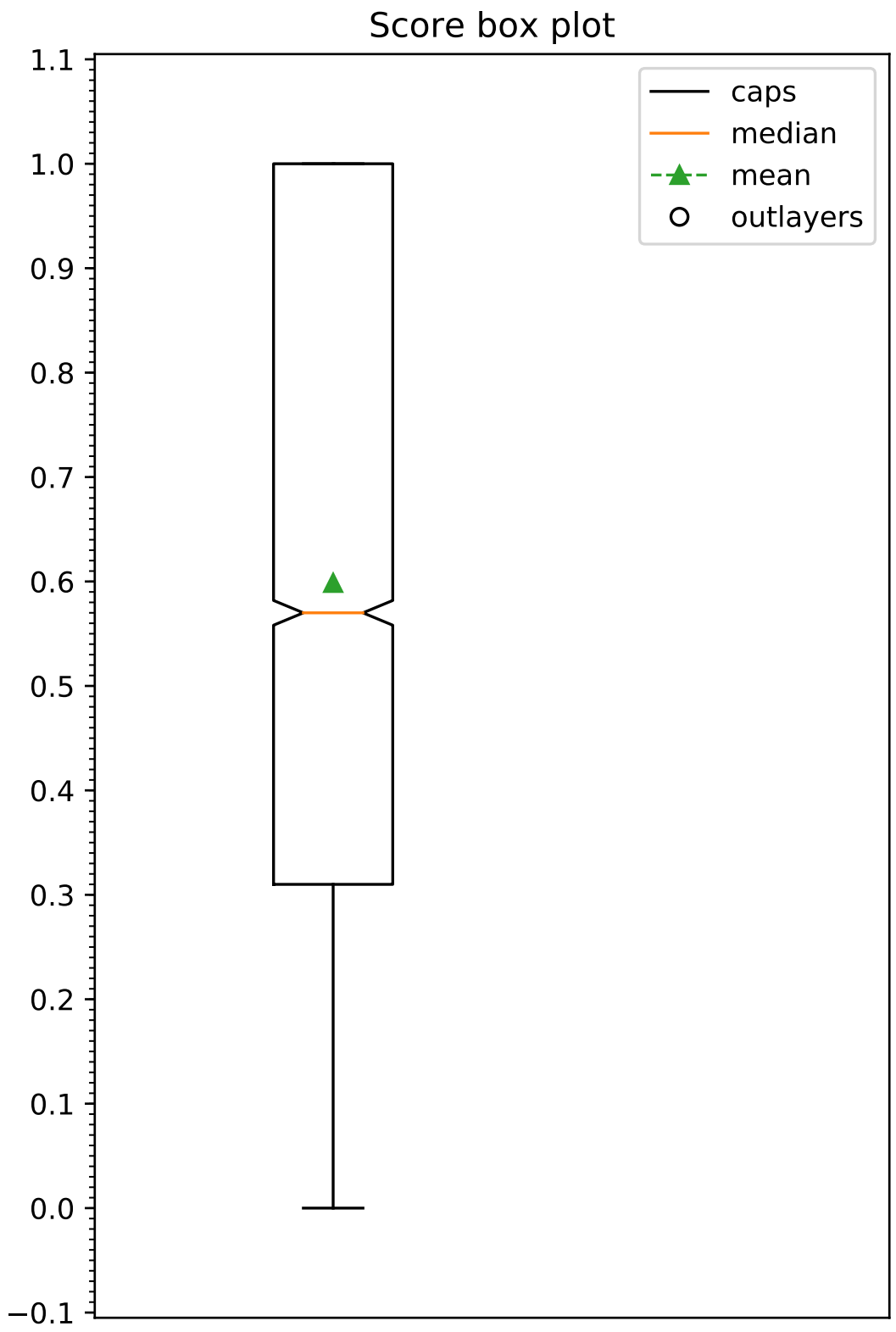

**Figure C.16.** Images score distribution in the initial dataset *5556x_markable_640x640*.

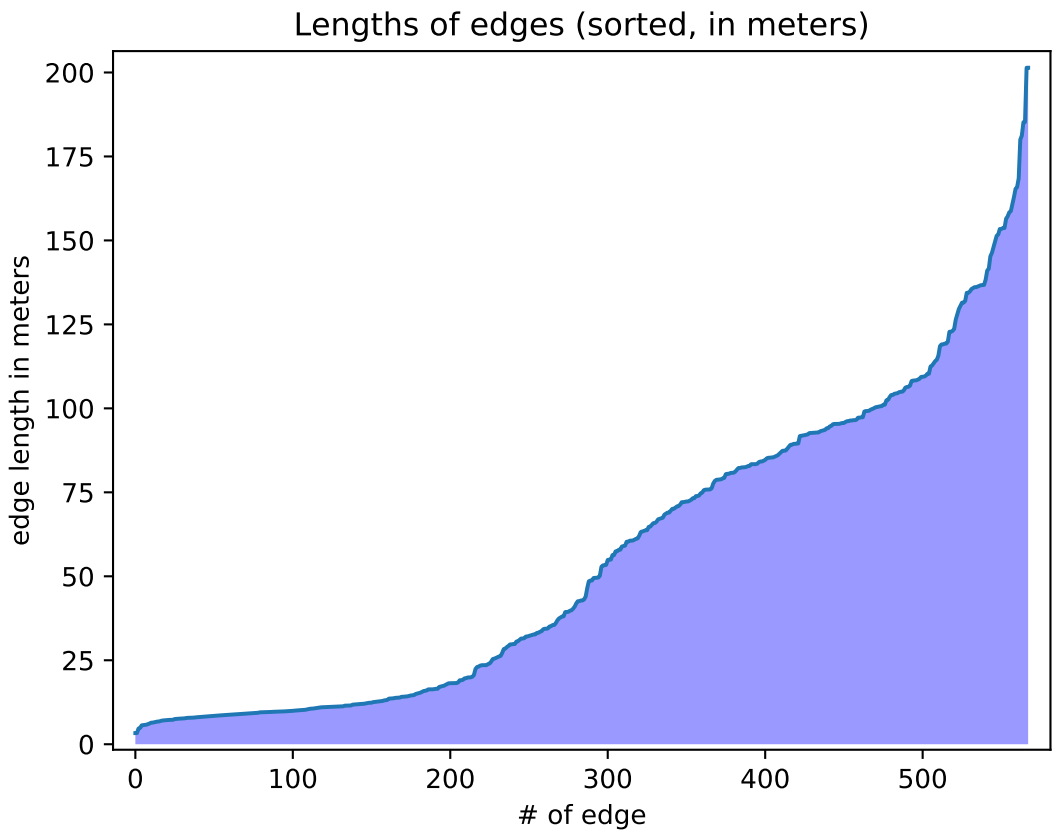

**Figure C.17.** Edge lengths from the initial dataset *5556x_markable_640x640* sorted by value.

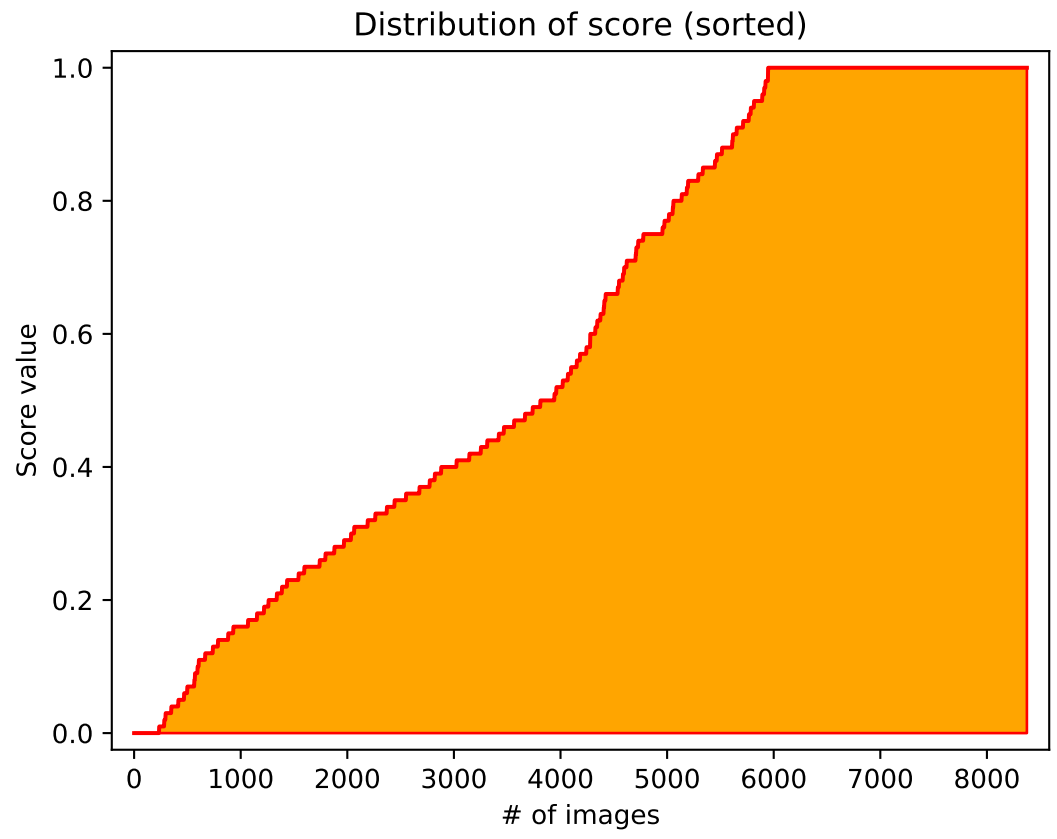

**Figure C.18.** Image scores from the initial dataset *5556x_markable_640x640* sorted by value. Note that we are using images generated from valid edges, which had score labels in the initial dataset.

# Appendix D
# CD Content

## D.1 CD Content

Following folders are placed on CD accompanying this thesis.

- **/Source code**
  Folder with python project, selected dataset and setup instructions.
- **/Documentation**
  Project documentation generated by Sphinx.
- **/Thesis TEXsource**
  Folder contains sources for this thesis. Can be compiled with `pdfcsplain`.
- **/Experiment results**
  Contains results of the experiments. We present selected evaluated GeoJSON files alongside with selected model files and saved histories used to plot graphs in this thesis.